\documentclass[Afour,sageh,times]{sagej}

\usepackage{moreverb,url}

\usepackage[toc,page]{appendix}
\usepackage[colorlinks,bookmarksopen,bookmarksnumbered,citecolor=red,urlcolor=red]{hyperref}
\usepackage{bm}
\usepackage{url}
\usepackage{graphicx} % for pdf, bitmapped graphics files
\usepackage[mathscr]{euscript}  
\usepackage{amsmath} % assumes amsmath package installed
\usepackage{amssymb}  % assumes amsmath package installed
\usepackage{multirow}
\usepackage[load=prefixed, binary-units=true, per-mode = symbol]{siunitx}
\usepackage[table]{xcolor}
\usepackage{adjustbox}
\usepackage{array}
\usepackage{booktabs}
\usepackage{framed}
\usepackage{dcolumn}
\usepackage{diagbox}
\usepackage{flushend}
\usepackage{pifont}% http://ctan.org/pkg/pifont
\newcommand{\cmark}{\text{\ding{51}}}%
\newcommand{\xmark}{\text{\ding{55}}}%

\newcolumntype{d}[1]{D{.}{.}{#1}}

\newcommand{\secref}[1]{Section~\ref{#1}}

\renewcommand{\eqref}[1]{Equation~(\ref{#1})}
\newcommand{\figref}[1]{Figure~\ref{#1}}
\newcommand{\tabref}[1]{Table~\ref{#1}}

\newcommand\BibTeX{{\rmfamily B\kern-.05em \textsc{i\kern-.025em b}\kern-.08em
T\kern-.1667em\lower.7ex\hbox{E}\kern-.125emX}}

\DeclareSIUnit{\million}{\text{M}}
\DeclareSIUnit{\billion}{\text{B}}

\hypersetup{draft}
\begin{document}

\runninghead{Radwan~\textit{et~al.}}

\title{Multimodal Interaction-aware Motion Prediction for Autonomous Street Crossing}

\author{Noha Radwan, Wolfram Burgard and Abhinav Valada}

\affiliation{Department of Computer Science, University of Freiburg, Germany}

\corrauth{Noha Radwan, Albert-Ludwigs-Universit\"at Freiburg,
Autonome Intelligente Systeme,
Georges-K\"ohler-Allee 080,
79110 Freiburg,
Germany.}

\email{radwann@cs.uni-freiburg.de}

\begin{abstract}
For mobile robots navigating on sidewalks, it is essential to be able to safely cross street intersections. Most existing approaches rely on the recognition of the traffic light signal to make an informed crossing decision. Although these approaches have been crucial enablers for urban navigation, the capabilities of robots employing such approaches are still limited to navigating only on streets that contain signalized intersections. 
In this paper, we address this challenge and propose a multimodal convolutional neural network framework to predict the safety of a street intersection for crossing. Our architecture consists of two subnetworks; an interaction-aware trajectory estimation stream IA-TCNN, that predicts the future states of all observed traffic participants in the scene, and a traffic light recognition stream AtteNet. Our IA-TCNN utilizes dilated causal convolutions to model the behavior of all the observable dynamic agents in the scene without explicitly assigning priorities to the interactions among them. While AtteNet utilizes Squeeze-Excitation blocks to learn a content-aware mechanism for selecting the relevant features from the data, thereby improving the noise robustness. Learned representations from the traffic light recognition stream are fused with the estimated trajectories from the motion prediction stream to learn the crossing decision. Incorporating the uncertainty information from both modules enables our architecture to learn a likelihood function that is robust to noise and mispredictions from either subnetworks. Simultaneously, by learning to estimate motion trajectories of the surrounding traffic participants and incorporating knowledge of the traffic light signal, our network learns a robust crossing procedure that is invariant to the type of street intersection. Furthermore, we extend our previously introduced Freiburg Street Crossing dataset with sequences captured at multiple intersections of varying types, demonstrating complex interactions among the traffic participants as well as various lighting and weather conditions. We perform comprehensive experimental evaluations on public datasets as well as our Freiburg Street Crossing dataset, which demonstrate that our network achieves state-of-the-art performance for each of the subtasks, as well as for the crossing safety prediction. Moreover, we deploy the proposed architectural framework on a robotic platform and conduct real-world experiments which demonstrate the suitability of the approach for real-time deployment and robustness to various environments.\end{abstract}


\maketitle

\section{Introduction}
\label{sec:intro}

Recent advances in robotics and machine learning have enabled the deployment of mobile robots for day-to-day tasks whether as domestic cleaners, navigational aids, autonomous driving vehicles or last-mile delivery agents. In most of these applications, as robots navigate closely around humans, it is essential that they follow the navigational conventions while also being robust to unexpected situations. The ability to autonomously navigate across street intersections is among the situations in which a robot can cause unintended outcomes for the surrounding traffic participants. 
%In most of these applications, as robots navigate and interact closely with humans, it is essential that they behave in a manner that is compliant to the behavioral norms in order to enable the ease of integration. 
%Over the last two decades, robotics research has seen a number of advancements pushing more towards the integration of robots into our daily lives whether as domestic cleaners, navigational aids, autonomous driving vehicles or last-mile delivery agents. Among the key requirements for mobile robots which share the space with humans, is their ability to identify possibly dangerous situations and seek a plan to avoid them, while behaving in a predictable manner for the surrounding humans. Handling street intersections is among these situations for which the behavior of the mobile robots can pose a threat or result in disastrous outcome for surrounding traffic participants. 

In order to decide if a street intersection is safe for crossing, humans are taught at an early age to follow a rigorous decision making process which is comprised of checking and waiting for the traffic light signal, followed by looking in both directions to ensure the safety of the intersection for crossing. Hence, solely relying on the traffic light information to make the crossing decision is suboptimal as not only is the traffic light recognition task challenging in itself, the signal alone does not ensure the intersection safety for crossing. For example, when a speeding vehicle such as an ambulance or firetruck approaches an intersection, it has the right of way as it does not necessarily follow the traffic regulations. Traffic participants such as pedestrians and vehicles are required to wait until the intersection becomes clear. This problem becomes even more challenging with the varying types of intersections and the associated rules on how to cross each variant. For instance, the standard convention at zebra crossings is that the pedestrian has the priority for crossing the intersection, whereas the oncoming traffic slows down and stops until they have crossed. While, at unmarked intersections such as a side street merging into a main road, there is neither a traffic light to regulate the crossing nor does the pedestrian have the right of way. Further complicating the problem, the topology of the road such as street width, presence of a middle island and direction of traffic play an important role in determining the crossing procedure. Hence, hard-coding a set of behavioral rules for a mobile robot to abide by at intersections is not only highly tedious, but also requires constant upkeep and tailoring to suit varying scenarios that change with each region or city.

In this work, we propose a convolutional neural network framework to address the problem of autonomous street crossing that considers the dynamicity of the scene as well as factors that influence the crossing decision such as the presence of a traffic light. Our network consists of two streams, an interaction-aware motion prediction stream to estimate the future states of all surrounding traffic participants and a traffic light recognition stream to predict the state of the traffic light. Our framework fuses feature maps from both network streams to learn the crossing decision in an end-to-end manner, rendering it tolerant to noise and mispredictions by either subnetwork, as well as making it inherently robust to the type of intersection and surrounding road topology.

%Furthermore, encoding a set of behavioral rules for a mobile robot to abide by at intersections is not only tedious, but also requires constant upkeep and tailoring to suit the varying environments. In this work, we propose a learning based approach to address the problem of autonomous street crossing by incorporating knowledge from a traffic light recognition system and a behavior prediction system to predict the safety of the intersection. By estimating the future  trajectories of surrounding traffic participants, the mobile robot inadvertently learns a socially compliant navigation behavior while gaining a more complete image of the scene. Incorporating this information in turn enables the robot to make a crossing decision that robust to the intersection type.
%is more robust to mispredictions from the traffic light recognition module.

\begin{figure*}
\footnotesize
\centering
\includegraphics[width=1\linewidth]{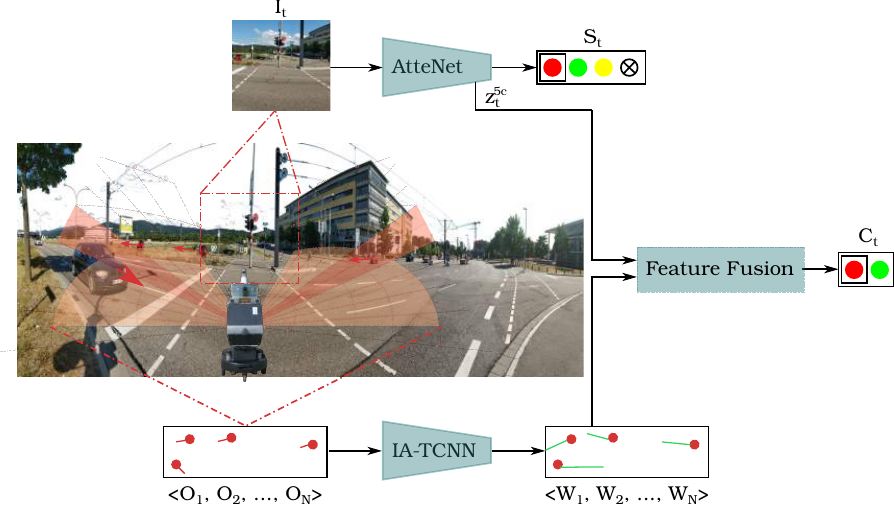}
\caption{Schematic representation of our proposed system for autonomous street crossing. Our approach is comprised of two main modules; a traffic light recognition network and an interaction-aware motion prediction network. Feature map outputs of both modules are utilized to predict the safety of the intersection for crossing.}
\label{fig:coverfig}
\end{figure*}

Predicting and modeling the behavior of agents whether pedestrians or vehicles is an extremely challenging problem that requires understanding the navigation conventions as well as the complex interactions among the various agents. For humans, identifying and following these conventions during navigation is a skill learned over several years that often needs readjustment depending on the environment. Hence, formalizing a set of behavioral rules for a mobile robot to uphold is both complex and taxing, requiring constant maintenance for each new environment. Recently, learning based motion prediction approaches~\citep{kim2017probabilistic, alahi2016social} have shown considerable robustness in modeling interactions among agents in real-world scenarios. However, as the density of the scene increases, their run-time and representational capabilities decrease, as they rely on modeling each agent separately by considering only their local neighborhood. Furthermore, current approaches which employ recurrent networks for modeling the behavior of agents define a local neighborhood of surrounding agents whose actions are likely to affect the motion of the current ``query" agent. Deciding which agents to include in the local neighborhood requires handcrafted definitions and domain-level knowledge which is undesirable~\citep{burgard2020perspectives} for achieving a robust framework that can be applied in different settings.

In order to address these problems, in this paper we propose the novel IA-TCNN architecture for interaction-aware motion prediction to jointly estimate the future trajectories of all the observed agents in the scene from 2D trajectory data. We utilize a data driven method that leverages the 2D tracks from LiDAR data provided through an obstacle detection/tracking module to represent the behavior of the different agents in the scene. Through leveraging the inherent motion interdependencies between the different agents and by employing causal convolutions, we enable our approach to simultaneously predict the behavior of all observed agents in the scene and learn these interactions without manually specifying a set of behavioral rules~\citep{helbing1995social, pellegrini2009you}. While, sequence modeling problems such as trajectory estimation have been mostly tackled using recurrent neural networks, recent studies have shown that temporal convolutional neural networks are able to more effectively model sequence-to-sequence tasks~\citep{bai2018empirical, valada2017deep}. To the best of our knowledge, this work is the first to employ causal convolutions in networks that perform behavior and motion prediction.

% However, as the complexity of the scene increases, the run-time and representational capabilities of such approaches substantially decreases, since they rely on modeling each agent separately by considering only their local neighborhood. To address the aforementioned issues, we propose a novel scalable approach for interaction-aware motion prediction in populated environments. We frame the problem of trajectory estimation as a sequence-to-sequence modeling task. By utilizing a data driven method to represent the behavior of the different agents, we enable our approach to leverage the inherent interdependencies in their motion, thereby learning interactions without manually specifying a set of behavioral rules~\citep{helbing1995social, pellegrini2009you}. While sequence modeling problems have been mostly tackled using recurrent neural network approaches, recent studies show that using convolutional architectures can indeed outperform recurrent networks~\citep{bai2018empirical}. We propose the novel IA-TCNN network to address the task of interaction-aware motion prediction by employing causal convolutions which facilitate both accurate modeling of the sequential behavior of the various agents and online deployment in resource constrained systems. 
 
We propose the novel AtteNet architecture for traffic light recognition that is robust to varying weather and lighting conditions. Our architecture incorporates \mbox{\textit{Squeeze-Excitation}} blocks~\citep{hu2017squeeze}, thereby enabling it to learn a robust feature recalibration method that explicitly models the complex interdependencies between the channels of the various feature maps. This allows the network to actively suppress irrelevant features in the scene and highlight the most relevant features, which in turn enables it to learn representations that are robust to noisy data. Note that we do not strive to achieve state-of-the-art performance for traffic light recognition. Our aim is however to incorporate the information regarding the traffic light signal into the street crossing predictor to learn a model that acts in accordance with the navigational norms.

%Furthermore, we propose a convolutional neural network architecture, which we dub AtteNet, for the task of traffic light recognition. By incorporating \textit{Squeeze-Excitation} blocks~\citep{hu2017squeeze} into our network architecture, we improve its representational capabilities and enable it to learn robust features that are invariant to noise in the data. While developing an approach for accurate traffic light recognition is not the main focus of this work, we provide the aforementioned network for the sake of completeness, however, it can be replaced by any other module for traffic light recognition. 

\figref{fig:coverfig} depicts the proposed architecture for intersection safety prediction along with the constituting subnetworks. The input to our network is an RGB image of the scene and 2D trajectory information of all observed dynamic agents in terms of spatial coordinates encoded relative to the robot for an interval of time. Our network simultaneously predicts the traffic light signal, the future states of all traffic participants over a prediction window and the safety of the intersection for crossing during this interval. As our method does not rely on structural knowledge of the environment or any form of communication with the surrounding traffic participants, it can be applied independently of the intersection type. Our model can be readily used to continuously track the states of surrounding agents as well the traffic light status and change the crossing behavior online. Due to the modularity of our network, the inference is in fact easily interpretable by visualizing the states of surrounding agents and the traffic light status.

We benchmark our IA-TCNN architecture on several publicly available datasets, namely ETH~\citep{pellegrini2009you}, UCY~\citep{lerner2007crowds} and L-CAS~\citep{yan2017online}, in addition to our own Freiburg Street Crossing dataset~\citep{radwan17iros}. For the traffic light recognition task, we benchmark on Nexar~\citep{nexar} and Bosch~\citep{BehrendtNovak2017ICRA} datasets as well as the Freiburg Street Crossing dataset. While for the autonomous crossing prediction, we perform extensive experimental evaluations on the extended Freiburg Street Crossing dataset. Additionally, in order to evaluate the performance of our approach for autonomous crossing prediction in real-world scenarios, we deploy our framework on a robotic platform and conduct real-world experiments at various street intersections in Freiburg. {A video showing real-world experiments can be found at \color{red}\url{https://youtu.be/I70fsqW3VOk}}. The results demonstrate that our architecture achieves state-of-the-art performance on each of the tasks, while being robust to enable generalization to new and unseen environments without the need for retraining or pre/post-processing.

% Feature maps from the traffic light recognition network are fused with the estimated trajectories for the surrounding agents and utilized to predict the safety of the intersection. As we do not require any prior structural knowledge of the environment, our proposed approach can be robustly applied in varying environments independent of the type of intersection with no preprocessing. We perform extensive experimental evaluations on each of the subproblems using several publicly available datasets and demonstrate the performance gains using the proposed methods.

In summary, the primary contributions of this work are as follows:
\begin{itemize}
\item A novel multimodal convolutional neural network architecture for intersection crossing safety prediction that jointly predicts the state of the traffic light and the future trajectories of surrounding traffic participants, thereby learning a crossing decision that is intersection agnostic.% Our network then utilizes information from both subnetworks to learn a crossing decision that is invariant to intersection types as well as underlying road topologies.
\item The novel IA-TCNN architecture for interaction-aware motion prediction to model the complex behavior and interactions among all observed agents in a scene while maintaining a fast inference time and being efficiently deployable in robots with limited resources.
%\item The novel AtteNet architecture for traffic light recognition that learns robust representations from the scene data to enable accurate recognition. %that utilizes \textit{Squeeze-Excitation} blocks to learn robust representations by leveraging global information to adaptively select relevant features from the input data.
\item We extend the previously introduced Freiburg Street Crossing dataset~\citep{radwan17iros} consisting of images, LIDAR as well as RADAR data by eight sequences captured at various intersections along with annotations for the traffic light state, trajectory annotations for the tracked dynamic objects and the corresponding crossing decision, and make the dataset publicly available.
\item We present extensive qualitative and quantitative analysis on each of the proposed modules on various publicly available benchmarks coupled with real-world experiments to demonstrate their utility in challenging real-world scenarios.
%over 25000 RGB images from different intersections along with annotations for the state of the traffic light and the corresponding crossing decision\todo{add extra sequences and dataset link}.
%\item We present extensive qualitative and quantitative analysis as well as ablation studies on each of the proposed modules to investigate their aptness in various challenging scenarios.
\end{itemize}

%By incorporating the uncertainty information from each prediction, our model inherently accounts for

\section{Related Works}
\label{sec:relatedworks}
Over the last decade, there has been significant work in the areas of motion prediction, traffic light recognition and intersection handling. In the following, we review some of the techniques developed thus far for addressing each of these tasks.\\
\textbf{Motion Prediction} approaches can be divided into two categories; methods modeling interactions among pedestrians and approaches modeling the behavior of vehicles. Among the first methods to model pedestrian interactions is the \textit{Social Forces (SF)} method of~\cite{helbing1995social} in which they apply a potential field based approach with attractive and repulsive forces to model the interactions among various pedestrians in the surrounding environment. An improvement of the \textit{SF} method was subsequently proposed by~\cite{yamaguchi2011you}, in which the authors employ a data-driven approach to estimate the hidden variables affecting the behavior of the agents such as group affinity and destinations. \cite{trautman2010unfreezing} propose a solution for the freezing robot problem in crowded navigation by utilizing concepts for human crowd navigation and collision avoidance. Using Gaussian processes to estimate the non-Markov nature of agents in the wild, they are able to predict navigation trajectories for the robot that are safer and shorter than trajectories taken by the compared pedestrians. Subsequently,~\cite{trautman2015robot} present a detailed navigation study in urban crowded environments investigating the effect of different cooperation strategies on the overall navigation performance utilizing insights from how humans navigate to deter the freezing robot problem. \cite{lerner2007crowds} use an example-based reactive approach to model pedestrian behavior by creating a database of local spatio-temporal scenarios. During an interaction, the autonomous agent samples its trajectory incrementally by considering similar spatio-temporal scenarios from the database. Subsequently,~\cite{pellegrini2009you} introduced the \textit{Linear Trajectory Avoidance (LTA)} method which uses similar concepts from crowd simulation to model the behavior of pedestrians in crowded environments using linear extrapolation over short intervals. \cite{van2008reciprocal} propose an alternative approach for multi-agent navigation, in which they extend on the velocity obstacle concept by assuming that surrounding agents follow a similar ``collision avoidance" policy. \cite{kuderer2012feature} employ a maximum entropy reinforcement learning approach to model human navigation behavior. In order to approximate the feature expectations, the proposed method employs Dirac delta functions at the modes of the distributions. However, while this approach was able to accurately model the behavior of pedestrians, suboptimal behavior often emerged due to the large amount of data required to capture the stochasticity of the human behavior. In order to address this problem and enable accurate modeling of the pedestrian behavior, \cite{kretzschmar2014learning} propose computing feature expectations using Hamiltonian Markov chain Monte Carlo sampling. \cite{kretzschmar2016socially} further expand on their proposed approach by utilizing a joint mixture distribution to capture both the discrete and continuous aspects of the design problem. They build upon the reinforcement learning approach with Hamiltonian Markov chain Monte Carlo sampling to learn a socially compliant navigation behavior. \cite{pfeiffer2016predicting} present a data-driven approach for interaction-aware motion prediction, wherein the authors employ a maximum entropy distribution over the navigation behavior observed from demonstrations. In order to facilitate deployment in real-world environment, the authors further propose a receding horizon motion planner that does not require knowledge of the destination for the surrounding pedestrians. The authors further provide evidence that learning interaction-aware motion trajectories from human demonstrations is insufficient due to the peaked interest of pedestrians in the robot which results in different behavior than the trained policy.

While the aforementioned approaches are able to capture the pedestrian behavior in specific situations, the need for defining hand-crafted features make them undesirable for deployment in dynamic environments. Inspired by the success of deep learning based approaches~\citep{valada2018incorporating} in the various areas of computer vision and robotics,~\cite{alahi2016social} propose an approach dubbed \textit{Social LSTM}. Using a Long-Short Term Memory (LSTM) network architecture and a \textit{Social Pooling} layer that leverages spatial information of nearby pedestrians thereby implicitly modeling interactions among them. Similarly,~\cite{sun20173dof} use a sequence-to-sequence LSTM encoder-decoder architecture to predict the pedestrian position and angle of direction. The authors show that incorporating the angular information in addition to the temporal information leads to a significant improvement in the accuracy of the prediction. \cite{vemula2017social} propose an alternative \textit{Social Attention} method to predict future trajectories based on capturing the relative importance of pedestrians regardless of their proximity. The authors formulate the problem as a spatio-temporal graph with nodes representing the pedestrians and edges capturing the dynamics of the interactions between two pedestrians such as orientation and distance. Concurrently,~\cite{pfeiffer2017data} propose an LSTM based data driven model for motion prediction by incorporating the obstacle map of the environment and encoding the surrounding pedestrians in polar angular space, thereby enabling fast inference times in crowded environments. \cite{chen2017socially} propose a deep reinforcement learning framework for socially aware motion planning. Unlike the aforementioned methods, the proposed approach relies on the fact that it is easier to specify which behavior is undesirable as opposed to defining socially compliant navigation. The authors develop a symmetrical neural network architecture and demonstrate in simulation generalization to densely crowded environments. \cite{gupta2018social} propose the use of recurrent based Generative Adversarial Network (GAN) to generate and predict socially acceptable paths. Their proposed \textit{SGAN} approach is comprised of an LSTM-based encoder-decoder generator to predict the future trajectories, followed by an LSTM-based discriminator to predict whether each generated trajectory follows the social norms. Similarly, \cite{sadeghian2018sophie} present a framework for predicting trajectories based on GAN dubbed \textit{SoPhie}. By utilizing an RGB image from the scene and the trajectory information of the pedestrians, the method computes both the physical and social context vectors by focusing on only the relevant information for each observed pedestrian. The computed vectors are then utilized by an LSTM-based GAN module to generate physically and socially acceptable trajectories. More recently, approaches that leverage the scene structure for predicting the future trajectories are proposed (\cite{manh2018scene}, \cite{varshneya2017human} and \cite{xue2018ss}). While utilizing structural knowledge of the scene~\citep{valada2016convoluted} enables such methods to achieve highly accurate motion estimates, it simultaneously limits the transferability of these approaches to environments that are known in advance due to either the need of a preprocessing mapping stage or training data from the environment. In the context of autonomous street crossing prediction, generalization capabilities to unseen intersections and new environments is a crucial requirement for our system. We therefore, do not compare our proposed approach to such methods that are dependent on the scene structure.

Over the years, several methods have been proposed for trajectory estimation of vehicles~\citep{lefevre2014survey, park2018sequence}. \cite{lefevre2011exploiting} propose a Bayesian network to infer the driver's intention by utilizing the digital map of the road network. \cite{kim2017probabilistic} propose a trajectory prediction method that employs a recurrent approach to predict the future coordinates of all surrounding vehicles using an occupancy grid map representation with probability values to reflect the uncertainty of the predictions. Similarly, \cite{bauman18predicting} propose an encoder-decoder architecture to predict the ego-motion of the vehicle using previous path information. In order to minimize the potential collision risk, \cite{park2018sequence} propose an encoder-decoder LSTM architecture accompanied with beam search to produce the most likely $\mathit{K}$ trajectories. 

Despite the varying application areas of the motion prediction task, there is a growing consensus that recurrent units in combination with trajectory information of the most relevant pedestrians/vehicles can provide accurate predictions. While this is true, it comes at the cost of the representational and run-time capabilities of these methods. As the majority of the aforementioned approaches model each pedestrian/vehicle separately by predicting only their local neighborhood, suboptimal behavior often occurs in complex densely populated environments. In this work, we propose a novel scalable neural network architecture to address the problem of learning trajectories in populated environments. Instead of the widely employed recurrent units such as LSTMs, our proposed network utilizes causal convolutions to model the sequential behavior of the various agents in the scene. Furthermore, by jointly learning the trajectories for all agents in the scene, our network is able to better leverage the interdependencies in the motion without the need for explicitly defining the relative importance of each agent. Finally, our approach is not restricted to modeling the behavior of either pedestrians or vehicles, but is rather able to learn and infer the complex interactions among the various types of agents in the scene.\\
\textbf{Traffic Light Recognition} is one of the vital tasks for autonomous agents operating in urban environments whether pedestrian assistant robots or autonomous vehicles. Although traffic lights are designed to be relatively easily perceived by humans, they are not always easily identified in camera images due to their small size, presence of other sources of similar lights e.g. brake lights, billboards and other traffic lights in different directions, and partial occlusions caused by different objects in the scene~\citep{jensen2016vision}. Furthermore, due to the highly dynamic nature of the environment, traffic light recognition approaches need to have fast inference times to enable safe deployment. In order to accurately recognize traffic lights in varying illumination conditions, \cite{john2014traffic} employ a Convolutional Neural Network (CNN) based approach to extract features from the image. Accompanied by a GPS sensor to identify the regions of interest within the image, the approach produces a saliency map containing the traffic light location to enable recognition in low lighting conditions. \cite{BehrendtNovak2017ICRA} propose a system for detecting, tracking and recognizing traffic lights for autonomous vehicles. Their approach utilizes the YOLO architecture~\citep{redmon2016you} to detect the location of the traffic lights within the image. The traffic lights are then tracked using the ego-motion information and stereo imagery to triangulate their relative position. Finally the state of the light is identified using a small neural network trained on the extracted regions. 

Similarly, in order to enable accurate traffic light recognition in complex scenes, \cite{li2018traffic} utilize prior information from the image regarding the position and size of the traffic light in order to reduce the computational complexity of locating it within the image. Additionally, they propose an aggregate channel feature method accompanied with inter-frame information analysis to facilitate accurate and consistent recognition across the different frames. With the goal of improving the run-time capabilities and reducing the computational resources, \cite{liu2017real} propose a traffic light recognition system operating in an online manner on smartphones. Using ellipsoid geometry in the HSV colorspace, their approach is able to extract region proposals which are in turn passed through a kernel function to recognize the phase and type of the traffic light. 

In contrast to the aforementioned methods for traffic light recognition, we do not perform any preprocessing or utilize any structural prior from the scene, rather our proposed network is able to attend to areas in the image containing the traffic light, thereby increasing ease of deployment and robustness to new environments.
\\
\textbf{Intersection Safety Prediction}: Among the early works on enabling autonomous street crossing for pedestrian assistant robots are the works of \cite{baker2003vision, baker2005automated} in which the authors propose a system to detect and track vehicles using cameras mounted on both sides of the robot. Using image differencing and edge extraction techniques, the method is able to identify and track vehicles in a two lane street. Subsequently, \cite{bauer2009autonomous} proposed an autonomous city explorer robot to navigate in urban environments. In their approach, the robot is able to handle street crossings by identifying and recognizing the state of the traffic light. In order to identify the safety of intersections for autonomous vehicles, \cite{de2013autonomous} propose a negotiation approach by solving local optimization problems for each of the vehicles approaching the intersection. Similarly, \cite{medina2015automation} propose a decentralized \textit{Cooperative Intersection Control (CIC)} system to enable safe navigation of a T-intersection for a platoon of vehicles. An alternate approach to cooperative intersection crossing is proposed in~\cite{azimi2014stip}, in which the authors propose a vehicle-to-vehicle intersection protocol guided by a GPS model, where each vehicle periodically broadcasts its pose and intent to nearby vehicles and the crossing priority is then decided by the vehicles among themselves. 

Inspired by learning from demonstration approaches, \cite{diaz2017veer} propose an approach to aid visually impaired users to remain within the crosswalk bounds while crossing a road. Their proposed method processes images from the scene to extract the relative destination of the user and in turn produces an audio signal as a beacon for the user to follow to reach the goal. More recently, \cite{habibi2018context, jaipuria2018transf} present techniques for pedestrian intent prediction at intersections by utilizing the contextual information of the scene and \textit{Augmented Semi Non-negative Sparse Coding (ASNSC)} for learning the motion primitives to enable more accurate predictions of the trajectories at street crossings. \cite{fang2018is} develop an approach for predicting the crossing intention of pedestrians. Their proposed method first detects and tracks pedestrians approaching the sidewalks and then utilize this information to estimate the pose of the pedestrians by fitting skeletal features which are in turn utilized by a Random Forest classifier to predict the crossing intent. 

In our previous work~\citep{radwan17iros}, we proposed a classification approach to predict the safety of an intersection for crossing by training a Random Forest classifier on tracked detections from both radar and LiDAR scanners which enables fast and reliable detections of oncoming traffic. While this method has the advantage of being independent to the intersection type, it lacks the ability to generalize to new unseen scenarios as it learns a discriminative model of the problem. This in turn can lead to suboptimal behavior when learning from imperfect data or when encountering an unseen situation. To address this issue, in this work we utilize information from both the interaction-aware motion prediction and traffic light recognition approaches to predict the safety of the intersection for crossing. By leveraging the predicted trajectories of surrounding vehicles and pedestrians in addition to the state of the traffic light if present, our proposed approach is able to accurately estimate the safety of the intersection for crossing. Furthermore, as we do not rely on any prior knowledge of the environment or any form of communication technique with the surrounding traffic participants, our proposed approach can be easily deployed in various environments without any additional preprocessing steps.

\section{Technical Approach}
\label{sec:approach}
In this section, we detail our proposed system for predicting the safety of the intersection for crossing by jointly learning to predict the future motion of the observed traffic participants and simultaneously recognizing the traffic light state. Our framework fuses the predicted future states as well as the uncertainities of the traffic participants from the motion prediction stream with feature maps from the traffic light recognition stream in order to predict the safety of the intersection for crossing. Note that the proposed networks for interaction-aware motion prediction and traffic light recognition can be deployed separately for their respective tasks. Furthermore, it is worth noting that in this work our goal is to investigate the impact of utilizing the information from the traffic light signal in combination with the predicted trajectories for surrounding dynamic agents to learn a safe street crossing classifier. We propose a traffic light recognition architecture for the sake of completeness and to demonstrate its effect on the overall system. However, it can be easily replaced with other traffic light detection/recognition modules.
% by learning to predict interaction-aware motion, recognize the traffic light state and fuse the output of both to predict the intersection safety. While the approach presented in this work focuses on the use of the aforementioned methods for learning the intersection safety, each of the methods can be deployed independently. We formulate the problem of predicting the intersection safety in the context of behavior prediction by predicting the trajectories of the surrounding dynamic objects and thus enabling our approach to learn socially compliant navigation behavior with minimum supervision. Furthermore, we propose a traffic light recognition network that utilizes the channel interdependencies within an image, thereby improving the representational capabilities and achieving accurate recognition results. In order to predict the intersection safety, we utilize the predictions of the interaction-aware motion prediction module and the traffic light recognition module to learn a probability distribution over the crossing decision.

Our proposed architecture depicted in~\figref{fig:coverfig} consists of two convolutional neural network streams; an interaction-aware motion prediction stream IA-TCNN and a traffic light recognition stream AtteNet. The learned representations from both streams are concatenated channel-wise and passed to the road crossing module which in turn produces a probability distribution over the crossing decision. Given an RGB image at the current timestep and the 2D trajectory information for each dynamic object over a window of time with respect to the position of the robot, the output of our model is the traffic light state, the predicted trajectory for each object over the prediction interval and the crossing decision. In the following sections, we will first detail each of the networks, followed by the fusion procedure for predicting the safety of the intersection for crossing.
\subsection{Interaction-aware Motion Prediction}
\label{sec:iamp}

Given the observed trajectory information for the observable dynamic agents in the vicinity of the robot either from a tracker or an object detection module, our motion prediction network predicts for each agent the future trajectory over a prediction interval.

In order to predict interaction-aware trajectories for each dynamic agent without explicitly specifying the importance of each agent for the behavior of the surrounding agents, we create a feature vector for each observation interval with size $N {\times} T_\text{obs} {\times} F$, where $N$ is the number of dynamic agents observed within the interval, $T_\text{obs}$ is the observation interval and $F$ is the set of features obtained from the tracker/detection module for each object. We order the dynamic agents in this feature vector with respect to their detection time and apply padding for objects that are visible later within the interval. Under this representation, the input feature vector for our network has the following format:
\begin{eqnarray}
\label{eq:featureVec}
\mathbb{O} = 
\begin{pmatrix}
    \mathcal{O}_1^{t_1} & \mathcal{O}_1^{t_2} & \mathcal{O}_1^{t_3} & \cdots & \mathcal{O}_1^{t_\text{obs}}\\[6pt]
   	\mathcal{O}_2^{t_1} & \mathcal{O}_2^{t_2} & \mathcal{O}_2^{t_3} & \cdots & \mathcal{O}_2^{t_\text{obs}}\\[6pt]
   	\vdots		& \vdots		  & \vdots 		& \cdots & \vdots\\[6pt]
   	\mathcal{O}_N^{t_1} & \mathcal{O}_N^{t_2} & \mathcal{O}_N^{t_3} & \cdots	 & \mathcal{O}_N^{t_\text{obs}}
  \end{pmatrix}
\end{eqnarray}

We represent each trajectory point for each agent $\mathcal{O}_i$ by the spatial coordinates of the agent $(x_i^t, y_i^t)$, the velocity $v_i^t$ and the yaw angle $q_i^t = (qw_i^t, qz_i^t)$ in normalized quaternion representation. As the trajectory lengths are expected to vary among agents, we pad the beginning and end of each trajectory such that they are aligned within the observation interval. Furthermore, in order to cope with the varying number of dynamic agents observed within a single interval, we additionally pad the input features to a fixed value $N$ which corresponds to the maximum number of agents observed within an interval. We elaborate further on how this padding is handled during training and inference at the end of this section. The set of features representing each agent are encoded relative to the robot, thereby facilitating the transferability of the method to different environments. Furthermore, we disregard the size information for each agent, thus each observed dynamic agent is represented as a point with the aforementioned features. While employing this representation discards the size information for each agent, this information is not essential for predicting the future trajectories. We further assume that it can be easily recovered from the object detection module from which the features are obtained. The trajectory of agent $i$ during the observation interval \mbox{$T_{\mathrm{obs}} = \left\lbrace 1, \ldots, t_{\mathrm{obs}} \right\rbrace$} is represented as: 
\begin{align} 
\label{eq:traj} 
\mathcal{O}_i = \left\lbrace\left(x_i^t, y_i^t, v_i^t, qw_i^t, qz_i^t\right) \in \mathbb{R}^5 \mid t \in T_{\mathrm{obs}}\right\rbrace.
\end{align} 
Our network predicts a feature vector with the same order as~\eqref{eq:featureVec} over the prediction interval \mbox{$T_{\mathrm{pred}} = \left\lbrace t_{\mathrm{obs}} + 1, \ldots, t_{\mathrm{pred}} \right\rbrace$.}

In order to represent this problem as a sequence-to-sequence modeling task, the predicted output at timestep $ t \in T_{\mathrm{pred}}$ can only depend on inputs from $t' \in T_{\mathrm{obs}}$. In other words, predictions cannot depend on future states of traffic participants. Moreover, we predict the future trajectories for an interval greater than or equal to the observation interval, as estimating the trajectories for an interval shorter than the observation interval is comparatively trivial. In this work, we strive to accurately predict the future states of dynamic agents for an interval longer than the observation interval.
%we need to satisfy two key requirements:
%\begin{itemize}
%\item The prediction interval should be greater than or equal to the observation interval in order to make the problem reasonably challenging. We demonstrate the efficiency of predicting the trajectories for an interval shorter than the observation interval in the ablation study in~\secref{subsec:ablationMP}. 
%\item The predicted output at timestep $ t \in T_{\mathrm{pred}}$ can only depend on inputs from $t\prime \in T_{\mathrm{obs}}$. In other words, predictions cannot depend on future states of traffic participants.
%\end{itemize}

\begin{figure}
\footnotesize 
\centering 
\setlength{\tabcolsep}{0.3cm} 
\begin{tabular}{cc}
{\includegraphics[width=0.3\linewidth]{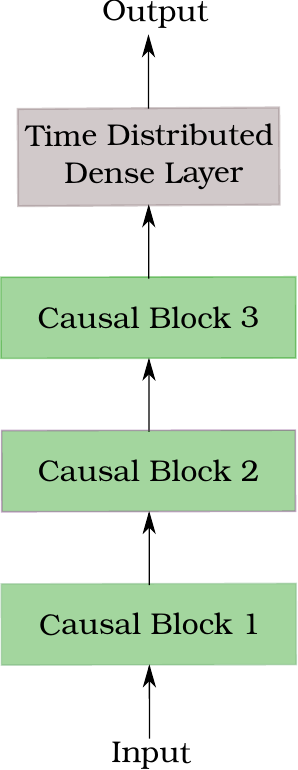}} &
{\includegraphics[width=0.3\linewidth]{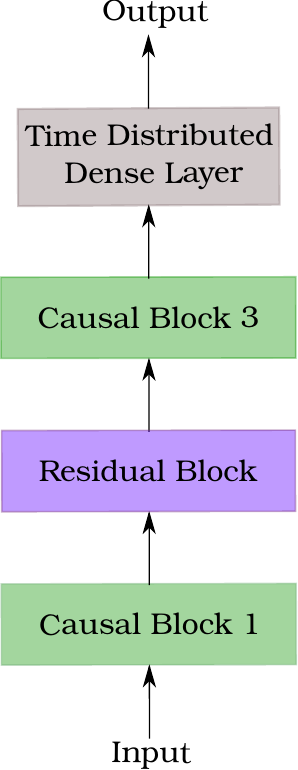}} \\ 
\\
\multicolumn{1}{c}{(a) IA-TCNN} & \multicolumn{1}{c}{(b) IA-DResTCNN}\\ 
\multicolumn{2}{c}{\includegraphics[width=0.5\columnwidth, angle=270]{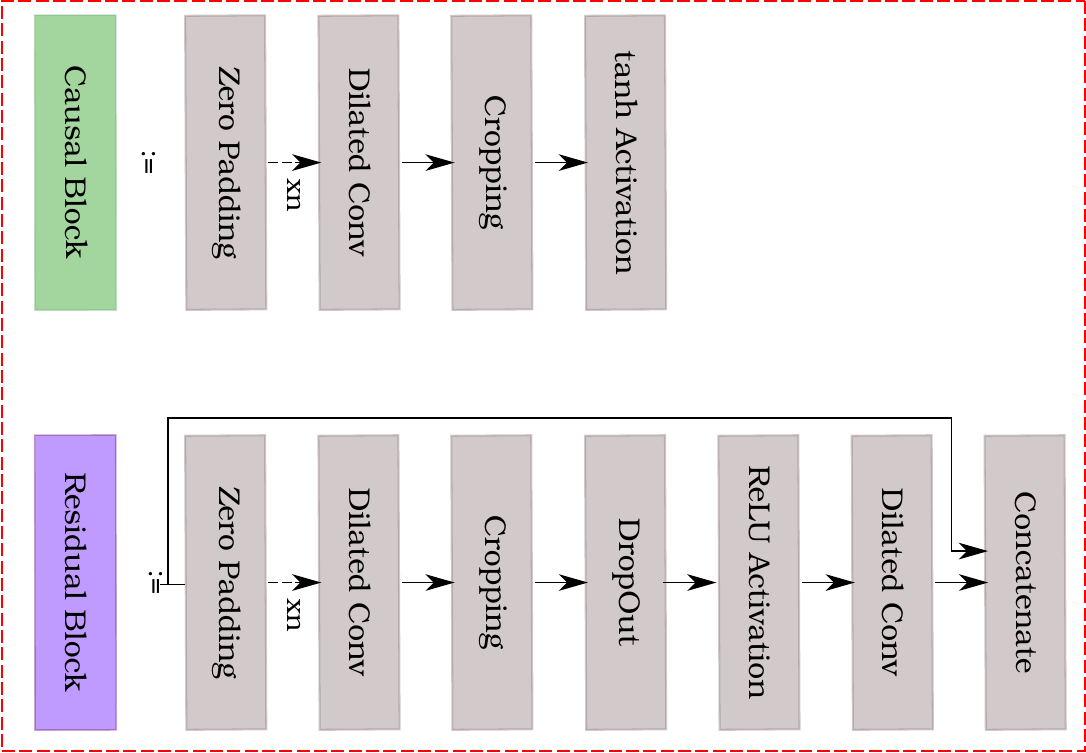}}\\
\end{tabular} 
\caption{Illustration of the proposed network architecture for interaction-aware motion prediction. We propose two variants of our architecure (a)IA-TCNN and (b) IA-DResTCNN. The legend enclosed in red dashed lines shows the constituting layers for each block.} 
\label{fig:iatcnn}
\end{figure}

We propose the \textit{Interaction-aware Temporal Convolutional Neural Network (IA-TCNN)} architecture depicted in~\figref{fig:iatcnn}(a) which fulfills the above criteria. Our network consists of three causal blocks; where each block contains zero-padding followed by $n$ dilated causal convolutions, cropping and a $\mathsf{tanh}$ activation function. In each block, we employ zero padding and cropping layers to fulfill the requirement of predicting a trajectory with length greater than or equal to the observed trajectory. We utilize causal convolutions where the output at each timestep is convolved with elements from earlier timesteps, thereby preventing information leak across different layers. 

Although the amount of previous information utilized by causal convolutions is linear to the network depth, increasing the depth or using extremely large filter sizes increases the inference time as well as the training complexity. We overcome this problem by employing dilated causal convolutions to increase the size of the receptive field without increasing the depth of the network. We use a constant kernel size of $30$ for each of the convolutional layers with filter sizes of $\left[128, 128, 128\right]$ respectively and increase the dilation rate by $1$ for each following convolution. Similar to current deep learning approaches for motion prediction, we model the predicted spatial coordinates of each pedestrian using a bivariate Gaussian distribution in order to obtain a measure of confidence over the output of the network (see \cite{alahi2016social, sun20173dof, vemula2017social}). The output of the last block is passed to a time distributed fully connected layer of size $9$ to produce temporal predictions for each timestep of the prediction interval, where for each pedestrian the network predicts the mean $\mu_i^t = \left(\mu_x, \mu_y, \mu_v\right)_i^t$, standard deviation $\sigma_i^t = \left(\sigma_x, \sigma_y, \sigma_v\right)_i^t$, correlation coefficient $\rho_i^t$ and quaternion components $\left(qw_i^t, qz_i^t\right)$.

We propose two variants of our method depicted in~\figref{fig:iatcnn} to further investigate the suitability of the proposed architecture for the sequence modeling task. \mbox{IA-LinConv} closely resembles the IA-TCNN architecture with the exception of setting the dilation rate $r=1$ and the number of dilated convolutions $n=1$, thus obtaining a single standard convolution per causal block. This variant is proposed to investigate the effect of adding a dilation factor on improving the representational learning ability of the network. In the second variant IA-DResTCNN, we replace the middle causal block with a residual causal block and the $\mathsf{tanh}$ activation function with a ReLU. By introducing residual connections in the network, we investigate if the current depth, filter size and dilation factor affect the stability of the architecture. Employing residual connections within the temporal convolutional network was proposed in~\cite{rene2017temporal} for action segmentation and detection with encouraging results demonstrating that using residual connections results in a larger receptive field without drastically increasing the number of parameters. We introduce this architectural variant with the goal of investigating if such a hypothesis is valid for the task of motion prediction.

We train our model by minimizing the weighted combination of the negative log likelihood loss of the groundtruth position $(x_i^t, y_i^t, v_i^t)$ under the predicted Gaussian distribution parameters $(\hat{\mu}_i^t, \hat{\sigma}_i^t, \hat{\rho}_i^t)$ and the $\mathcal{L}_2$ loss of the orientation in normalized quaternion representation $\hat{q}_i^t$ as follows:
\begin{align}
\label{eq:mploss}
\mathcal{L}_{\gamma} &= \left\| q_i^t - \hat{q}_i^t \right\|_2 \nonumber \\
\mathcal{L}_{p} &= -\log\left(\mathbb{P}\left(x_i^t, y_i^t, v_i^t\mid \hat{\mu}_i^t, \hat{\sigma}_i^t, \hat{\rho}_i^t\right)\right)\\
%\mathcal{L}_{mp} &= \sum_i^N \sum_t^{t_{\mathrm{pred}}} \Big\left( \mathcal{L}_{pos} \exp(-\hat{s}_{pos}) \Big\right. \nonumber\\ &+ \Big\left. %\mathcal{L}_{\gamma} \exp(-\hat{s}_{\gamma}) + \hat{s}_{pos} + \hat{s}_{\gamma} \Big\right)
\mathcal{L}_{MP} &= \sum_i^N \sum_t^{t_{\mathrm{pred}}} \mathcal{L}_{p} \exp(-\hat{s}_{p}) + \mathcal{L}_{\gamma} \exp(-\hat{s}_{\gamma}) + \hat{s}_{p} + \hat{s}_{\gamma} \nonumber,
\end{align}
where $N$ is the number of dynamic agents, $\hat{s}_{p}$, $\hat{s}_{\gamma}$ are learnable weighting variables for balancing the translational and rotational components of the predicted pose.\\
Since in real world data, the trajectories of different dynamic agents have varying lengths due to the limited sensor range and in order to fully leverage all the information available during training, we train our proposed IA-TCNN with dynamic sequence lengths by using binary activation masks predicted by the network to signify the end of a trajectory. During training, we use an input activation mask of size $N{\times}T_\text{obs}$ with the purpose of encoding the valid parts of the input. As our input feature vector has dynamic shape for the first two dimensions (number of observed agents and the time for each an agent is observed), the input activation mask encodes the valid parts for both dimensions and ``masks out" the padded sections with respect to the first two dimensions of the input. During inference, the network predicts a feature vector of fixed size $N{\times}T_\text{pred}{\times}F$ and an output activation mask of size $N{\times}T_\text{pred}$. Similar to the input activation mask, the output mask encodes the valid positions within the output feature vector along the first two dimensions of the output. Training the network with dynamic feature length (in terms of number of agents and observation/prediction time) enables the network to learn a robust representation that better aligns with real-world data; for instance when a pedestrian or vehicle exits the field of view of the sensor. The predicted trajectory is then first multiplied by the activation mask before computing the prediction error. Moreover, by utilizing the information from all agents during the observation interval, we eliminate the need for creating handcrafted definitions which attempt to explicitly model how the behavior of a dynamic agent is affected by the surrounding agents. Furthermore, it expedites the information flow throughout the various layers of the network, hence facilitating fast trajectory estimation for all the dynamic agents in the scene.

\subsection{Traffic Light Recognition}
\label{sec:tlr}
In this section, we describe the architecture of our traffic light recognition subnetwork, which given an input RGB image $I_t$ predicts the state of the traffic light $s_t \in S = \left\lbrace \textit{Red}, \textit{Green}, \textit{Yellow}, \textit{Off} \right\rbrace$. We build upon the ResNet-50 architecture~\citep{he2016identity} with pre-activation residual units which allow unimpeded information flow throughout the network thus enabling the construction of more complex models that are easily trainable. A description of the difference between standard residual units and pre-activation residual units can be found at~\secref{app:resunits}.Our proposed network \textit{AtteNet} consists of five bottleneck residual blocks with multiple pre-activation residual units. We replace the traditional ReLU activation function with ELUs which increases the robustness of the network to noise in the training data while allowing for shorter convergence time. Note that unlike the IA-DResTCNN architecture for interaction-aware motion prediction, we utilize the bottleneck residual units as the building block of our network due to their ability to aid in training deeper architectures without significantly increasing the number of parameters.

\begin{figure}
\footnotesize
\centering
\includegraphics[width=1.0\linewidth]{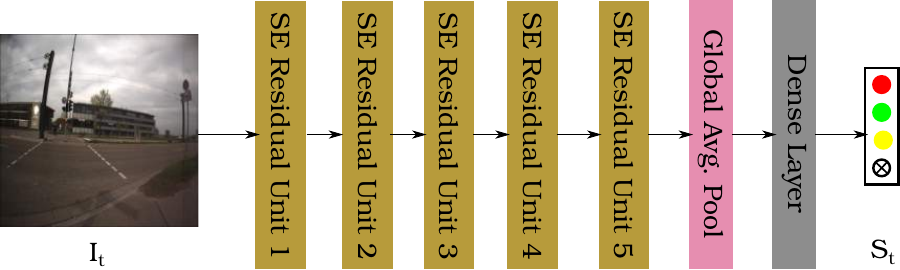}
\caption{Schematic depiction of our proposed AtteNet architecture for traffic light recognition. Given an RGB image $I_t$, our network predicts the state of the traffic light $s_t$ by aggregating the interdependencies between the different channels across the layers.}
\label{fig:attenet}
\end{figure}

In order to improve the representational learning abilities of our network, we introduce \textit{Squeeze-Excitation (SE)}~\secref{app:seblock} blocks into our network~\citep{hu2017squeeze}. Using SE blocks enables the network to perform feature recalibration, which in turn allows the network to utilize the global information in the images to selectively emphasize and suppress features depending on their usefulness for the task at hand. In order to further reduce the number of parameters of our model, we replace the fully connected layers in the SE blocks with $1\times1$ convolutional layers. We add a global average pooling layer after the fifth residual block, followed by a fully connected layer of size $4$ which produces the prediction of the network. Our final architecture is shown in~\figref{fig:attenet}. During training, we minimize the $\mathsf{softmax}$ cross entropy loss between the labels and the predicted logits.

\subsection{Learning To Cross The Road}
\label{sec:crossing}
In order to learn a crossing strategy that is robust to the type of intersection, we propose fusing the output predictions from the trajectory estimation subnetwork and the traffic light recognition subnetwork. Incorporating the traffic light recognition information is crucial at signalized intersections as the robot is expected to act within the behavioral norms obeying the intersection crossing rules such as crossing only when the light is green. At the same time, in certain situations, one cannot rely solely on the traffic light information to cross such as when an ambulance or police car is speeding towards an intersection. In such cases, despite the green pedestrian traffic light, the robot is expected to wait at the sidewalk until the intersection becomes safe for crossing. Similarly at unsignalized intersections, the robot is expected to identify safe crossing intervals from unsafe intervals. In these situations, utilizing the information from the trajectory estimation module is crucial to ensure safe crossing prediction. To achieve this goal, we perform element-wise concatenation of the feature maps from the traffic light recognition stream and the motion prediction stream. More specifically, the predicted Gaussian distribution parameters from IA-TCNN are first passed to a fully connected layer of dimension $D$, the output of which is reshaped to $H\times W\times C$ which corresponds in shape to the output of layer \textit{Res-5c} of AtteNet. Since we apply padding to the input and output of the IA-TCNN network to maintain constant feature size which corresponds to the maximum number of observable dynamic agents, fusing the features from both networks can be done in a rather straightforward manner. The output tensor from the reshaping is then concatenated with the output of layer \textit{Res-5c} of AtteNet and fed to a fully connected layer with $512$ units. This is then followed by another fully connected layer with $\mathsf{softmax}$ activation and $2$ output units signalizing the intersection safety state $c_t \in C = \left\lbrace \textit{Cross}, \textit{Don't Cross} \right\rbrace$. By utilizing the Gaussian distribution parameters to model the trajectories, we enable our model to incorporate the confidence information regarding the likelihood of the predictions, which in turn improves the robustness of our method to the prediction accuracy. We train the model by minimizing the $\mathsf{softmax}$ cross entropy loss function. In~\secref{subsec:crossEv}, we evaluate the impact of incorporating information from each of the streams on the accuracy of the learned crossing decision.

\section{Experimental Evaluation}
\label{sec:evaluation}

In order to evaluate our proposed system for predicting the safety of the intersection for crossing, we first evaluate each of the constituting subtasks followed by providing detailed results on the performance of the combined model. We evaluate our approach on multiple publicly available datasets as well as deploy it on our robotic platform shown in~\figref{fig:obelix} and evaluate the performance in a real-world environment in~\secref{sec:genExps}. We further provide comprehensive details of our evaluation protocol to facilitate comparison and benchmarking. We first present the datasets used for evaluating our approach, followed by the training schedule and extensive qualitative and quantitative analysis of the results. {Furthermore, a video showing real-world experiments can be found at \color{red}\url{https://youtu.be/I70fsqW3VOk}}.
%In this section, we perform an extensive evaluation of the performance of our proposed IA-TCNN for the task of motion prediction on both indoor and outdoor datasets, followed by an analysis of the proposed AtteNet for the task of traffic light recognition, and finally, we demonstrate the efficacy of combining both approaches for learning the safety of the intersection for road crossing.

\subsection{Datasets \& Augmentation}
\label{sec:data}
In the following, we discuss in detail each of the datasets used for evaluation as well as any preprocessing or augmentation procedure applied. As our architecture is comprised of multiple sub-networks and due to the unavailability of a large public labeled dataset capturing both motion prediction, traffic light recognition and intersection safety prediction, we split our evaluation procedure into three parts and subsequently employ public datasets for each of the aforementioned tasks.

\subsubsection{Motion Prediction Datasets}\mbox{}\\

\begin{figure}
\footnotesize 
\centering 
\setlength{\tabcolsep}{0.2cm} 
\begin{tabular}{p{3.5cm} p{3.5cm}}
{\includegraphics[width=3.5cm, height=3.5cm]{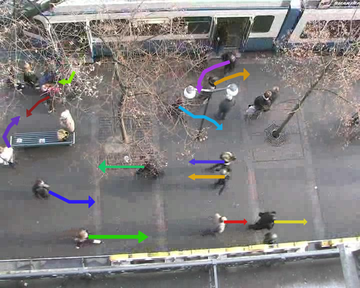}} &
{\includegraphics[width=3.5cm, height=3.5cm]{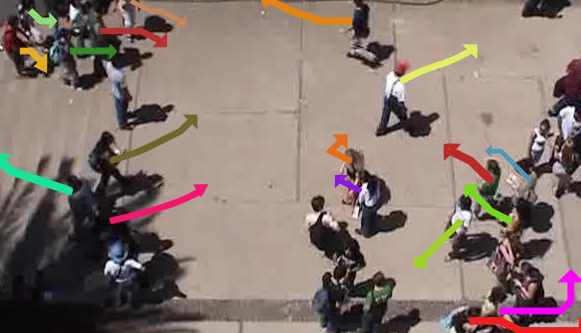}}\\
\\ 
\multicolumn{1}{c}{(a) ETH-Hotel} & \multicolumn{1}{c}{(b) UCY-Uni} \\
\\
{\includegraphics[width=3.5cm, height=3.5cm]{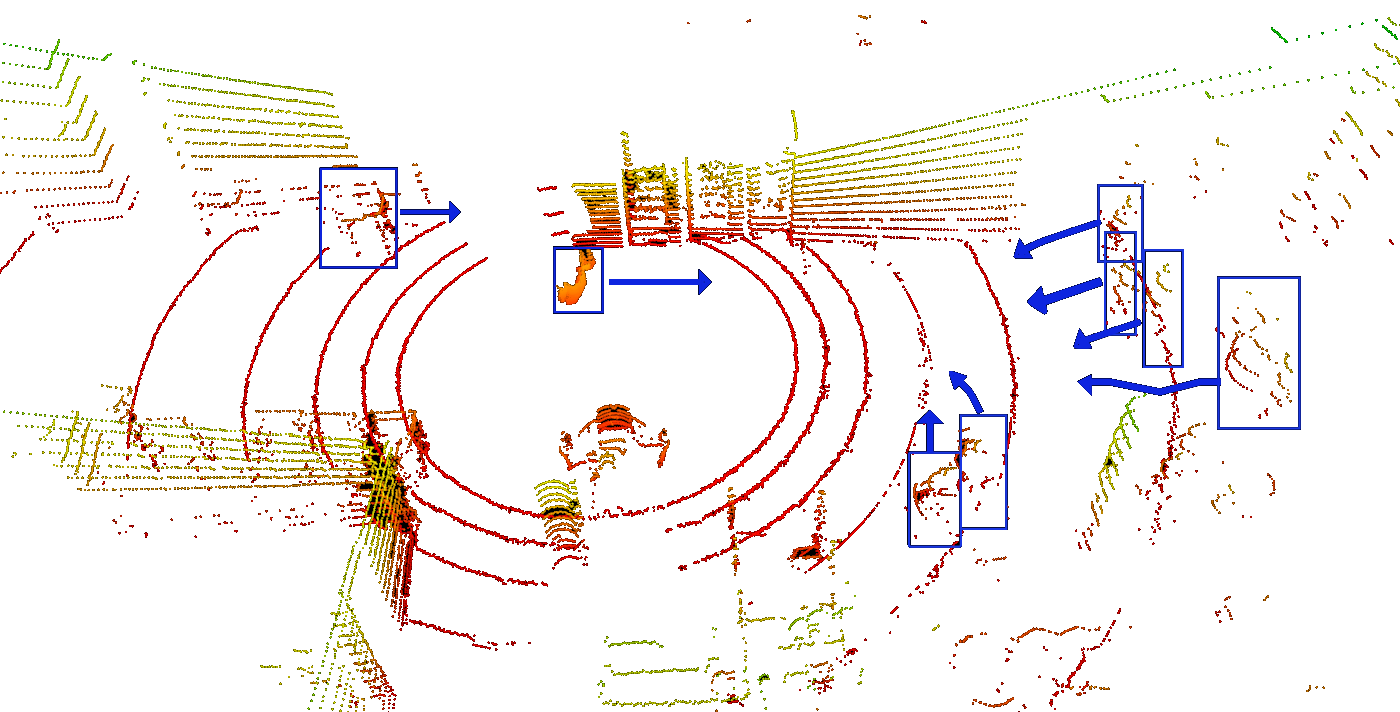}} &
{\includegraphics[width=3.5cm, height=3.5cm]{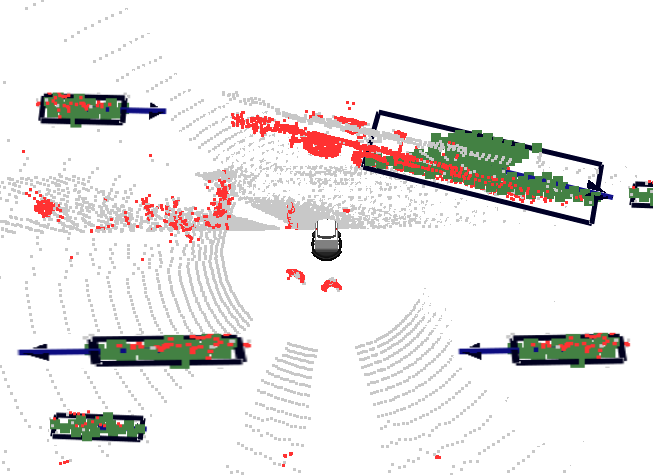}} \\
\\
\multicolumn{1}{c}{(c) L-CAS} & \multicolumn{1}{c}{(d) Freiburg Street Crossing (FSC)}
\end{tabular} 
\caption{Sample trajectories from the various datasets employed for benchmarking the interaction-aware motion prediction subnetwork. The benchmarking datasets include both camera captured sequences (ETH, UCY datasets) and LiDAR and radar captured sequences (L-CAS and Freiburg Street Crossing datasets). Overall, the datasets cover a wide range of motions among the various participants such as group behavior, trolley pushing and crowd navigation.} 
\label{fig:mpDatasets}
\end{figure}

In order to investigate the accuracy of our proposed architecture for the task of motion prediction, we rely on three benchmarking datasets that are commonly employed for the aforementioned problem. The selected datasets vary with respect to the environment in which they were captured (indoors versus outdoors), the capturing mode (camera versus LiDAR) and difficulty (crowded environment versus empty environment). \figref{fig:mpDatasets} provides an overview of the different datasets for motion prediction. An elaborate description of the different datasets can be found in~\secref{app:mopred_data}. Evaluating on such a large variety of environments enables us to gain a better understanding of the advantages and limitations of our proposed approach. Below, we provide a description for each of the datasets along with any pre-processing that was carried out prior  to training.

\subsubsection{Traffic Light Recognition Datasets}\mbox{}\\

\begin{figure}
\footnotesize 
\centering 
\setlength{\tabcolsep}{0.2cm} 
\begin{tabular}{p{3.5cm} p{3.5cm}}
{\includegraphics[width=3.5cm, height=3.5cm]{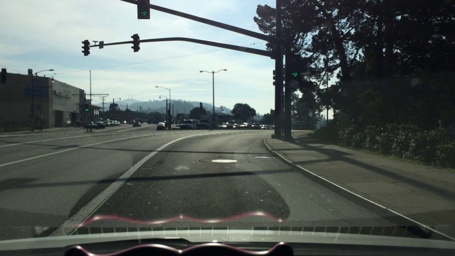}} &
{\includegraphics[width=3.5cm, height=3.5cm]{figures/nexar_red_red}} \\
\multicolumn{1}{c}{(a) Nexar} & \multicolumn{1}{c}{(b) Nexar} \\
{\includegraphics[width=3.5cm, height=3.5cm]{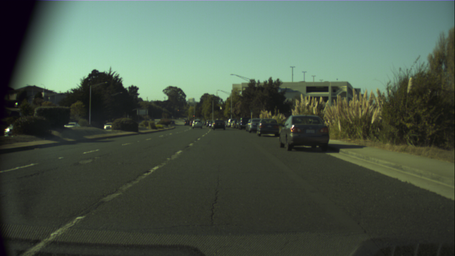}} &
{\includegraphics[width=3.5cm, height=3.5cm]{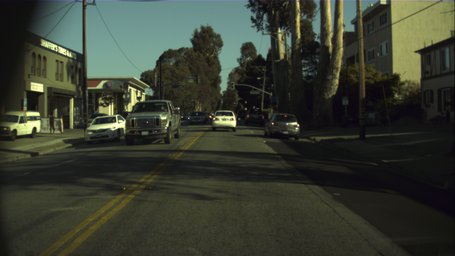}}\\
\multicolumn{1}{c}{(c) Bosch} & \multicolumn{1}{c}{(d) Bosch} \\
{\includegraphics[width=3.5cm, height=3.5cm]{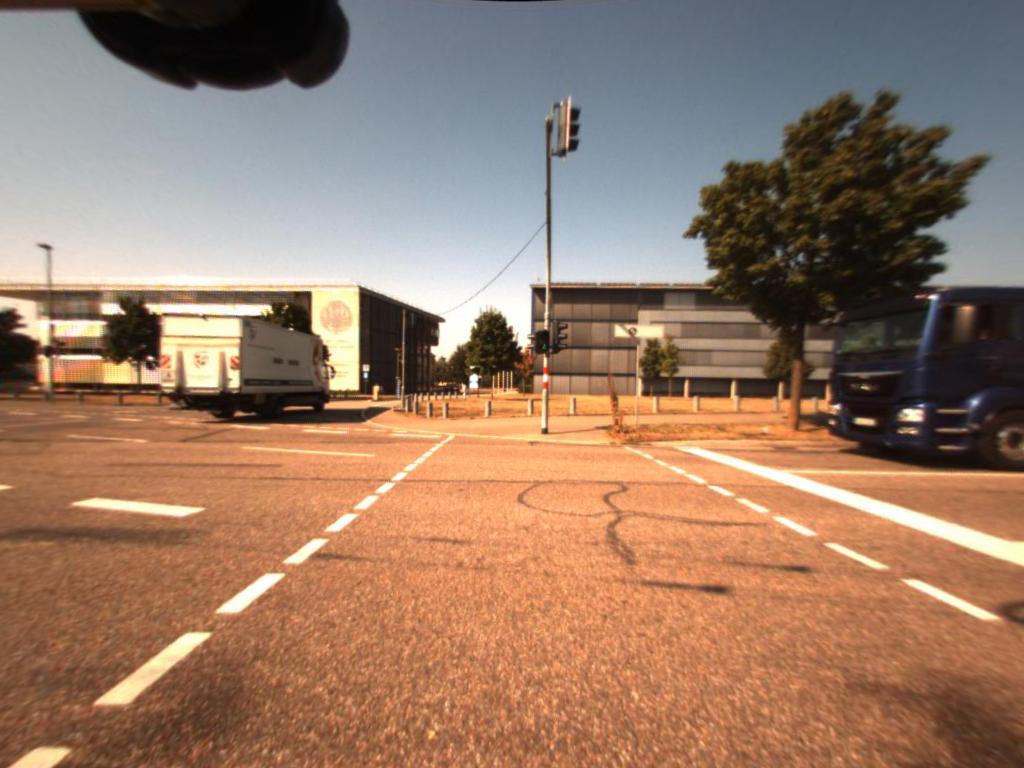}} &
{\includegraphics[width=3.5cm, height=3.5cm]{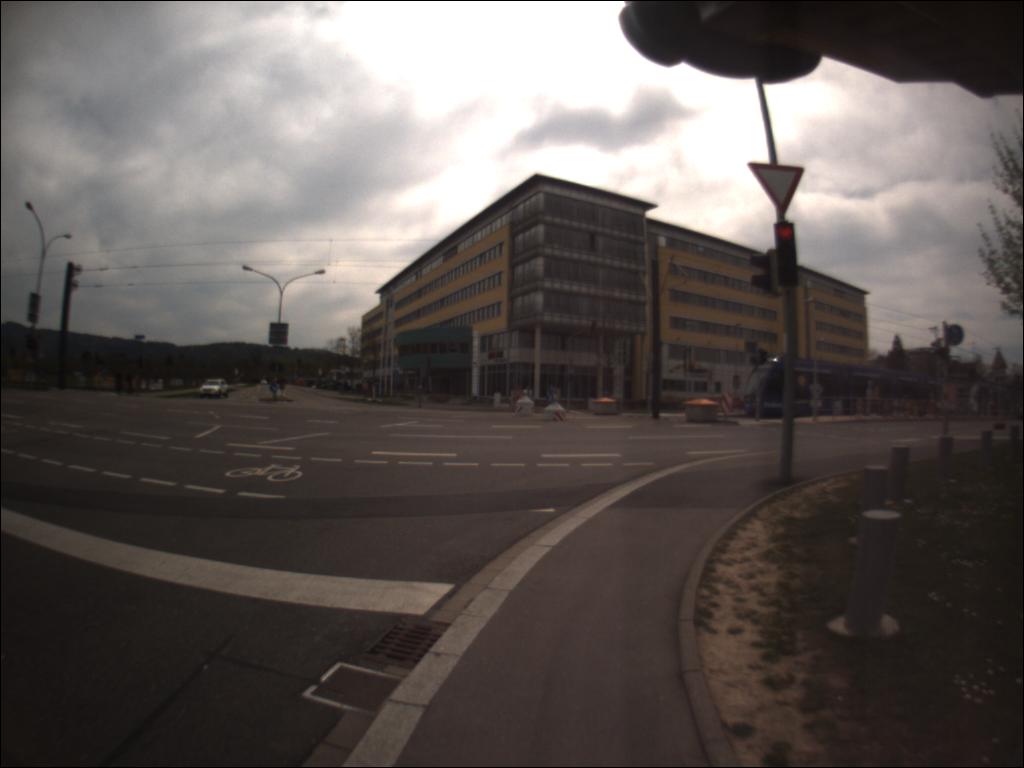}} \\
\multicolumn{1}{c}{(e) FSC} & \multicolumn{1}{c}{(f) FSC}
\end{tabular} 
\caption{Example images from the traffic light benchmark datasets which we evaluate on in this work, namely the Bosch Traffic Light dataset, the Nexar Challenge dataset and the Freiburg Street Crossing (FSC) dataset. The datasets cover a wide range of challenges for the task of traffic light recognition including occlusions, varying lighting and weather conditions, presence of multiple traffic lights in the image and motion blur.} 
\label{fig:tlrDatasets}
\end{figure}

We investigate the performance accuracy of our AtteNet architecture for the task of traffic light recognition by evaluating the model on two publicly available datasets. Despite targeting pedestrian traffic lights in our approach, to the best of our knowledge, there is a lack of publicly available pedestrian traffic light datasets. Nonetheless, the two tasks share a number of similarities which render benchmarking the performance on traffic light datasets a reliable estimate of the overall performance of the model on pedestrian traffic lights. Furthermore, our newly proposed Freiburg Street Crossing dataset contains several instances of pedestrian traffic lights, enabling us to evaluate the performance of our AtteNet on real-world data. We benchmark the performance of our model by evaluating on the Nexar Traffic Lights Challenge dataset and the Bosch Small Traffic Lights dataset depicted in~\figref{fig:tlrDatasets}. The datasets were selected due to the challenging nature of the data as images are captured in varying lighting conditions, contain multiple traffic lights and occlusions to the traffic light. Furthermore, the datasets contain a large number of sources of distraction such as brake lights of other cars and glass buildings which reflect the traffic light signal. Through evaluating our proposed AtteNet on such challenging datasets, we aim to gain an understanding of the power of the network for the task at hand, as well its generalizability to various environments and lighting conditions. In the following, we describe each of the datasets along with data augmentation procedures that were carried out. An extensive description of the different datasets can be found in~\secref{app:tlr_data}.

\subsubsection{Intersection Safety Prediction Dataset}\mbox{}\\

\begin{figure}
\footnotesize
\centering
\includegraphics[width=0.7\linewidth]{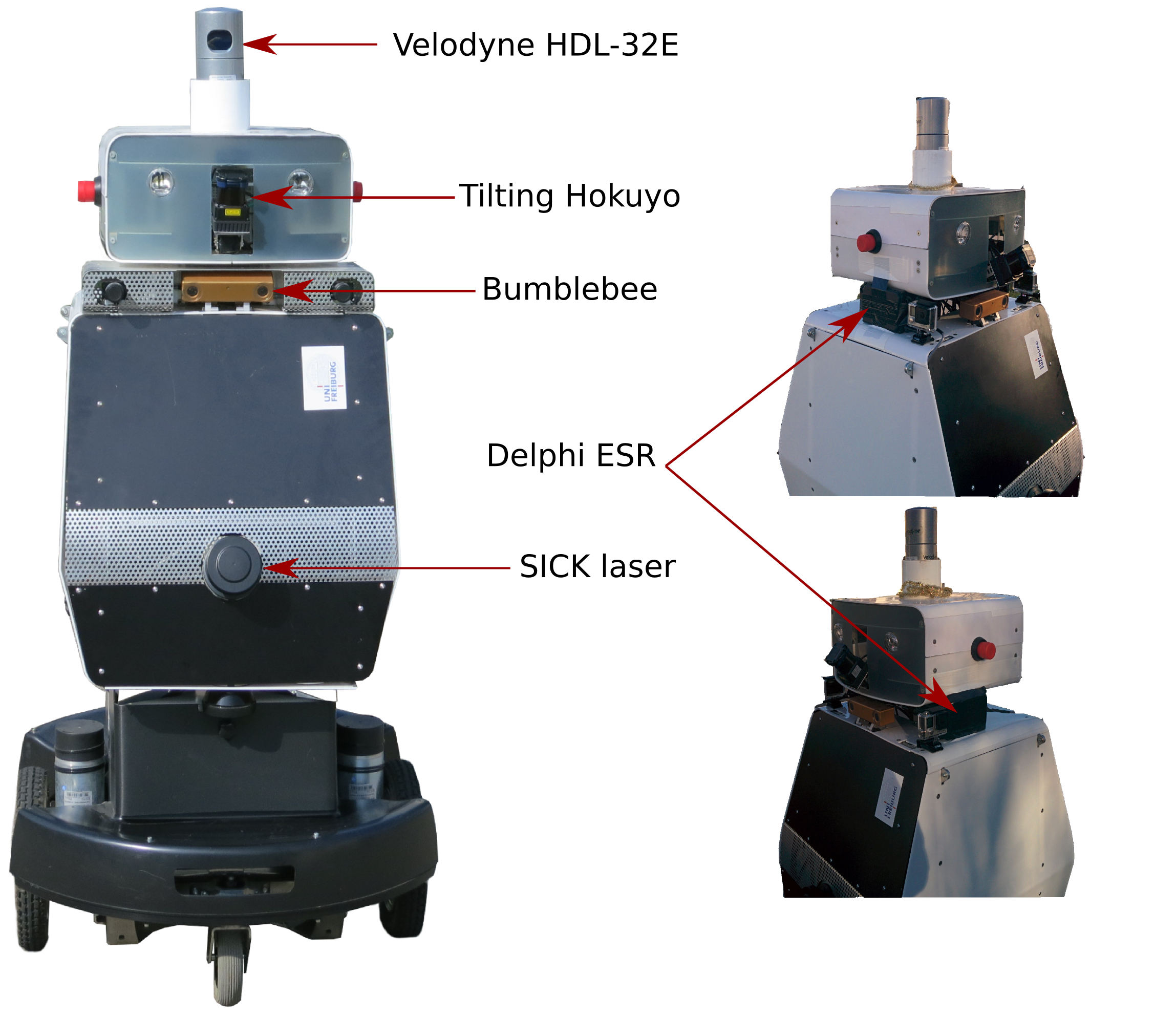}
\caption{Our robotic platform (Obelix) that was used for capturing the Freiburg Street Crossing dataset and conducting real-world experiments.}
\label{fig:obelix}
\end{figure}

\begin{figure}
\footnotesize 
\centering 
\setlength{\tabcolsep}{0.2cm} 
\begin{tabular}{p{3cm} p{3cm}}
{\includegraphics[width=3cm, height=3cm]{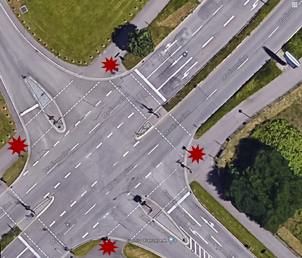}} &
{\includegraphics[width=3cm, height=3cm]{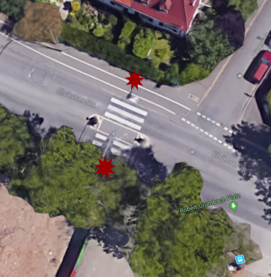}}\\
\\
{\includegraphics[width=3cm, height=3cm]{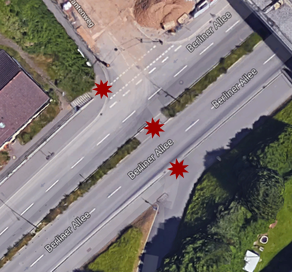}} &
{\includegraphics[width=3cm, height=3cm]{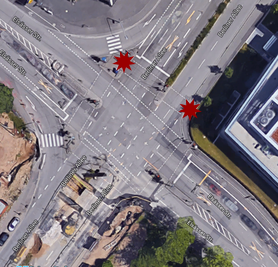}}\\
\\
{\includegraphics[width=3cm, height=3cm]{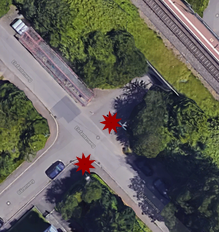}} &
{\includegraphics[width=3cm, height=3cm]{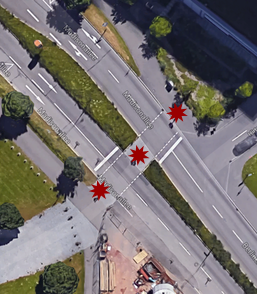}} 
\end{tabular} 
\caption{Birds-eye-view images from some of the road intersections included in the FSC dataset. The dataset covers a wide range of intersections including signalized, zebra crossings and crossings with middle islands, as well as different road curvature and streets merging. The red stars mark the position of the robot during data capturing. Note that for several intersections, data was gathered from different sides of the road (marked by multiple red stars in a single image), in order to capture traffic behavior from multiple angles.}
\label{fig:FSCIntersecs}
\end{figure}

In order to evaluate the overall performance of our model for the joint tasks of motion prediction, traffic light recognition and intersection safety prediction, we extend our previously proposed dataset~\cite{radwan17iros} with additional sequences and labels for each of the aforementioned tasks. The \textbf{Freiburg Street Crossing} (FSC) dataset consists of tracked detections of cars, cyclists and pedestrians captured at different intersections in Freiburg, Germany using a 3D LiDAR scanner and Delphi Electronically Scanning Radars (ESRs) mounted on our robotic platform shown in~\figref{fig:obelix}~\citep{radwan17iros}. Note that both the data capturing procedure and all experiments on this dataset were conducted using this robotic platform. Furthermore, we follow the same data collection and labeling procedure as in~\citep{radwan17iros}. The data was captured over the course of two weeks and it is divided into 10 different sequences containing approximately over 2000 tracked objects\footnote{The extended dataset is publicly available at:\\ {\color{red}\url{http://aisdatasets.cs.uni-freiburg.de/streetcrossing}}}. Each object is identified by a unique track ID, spatial coordinates, velocity and orientation angle. We obtain this information through the radar and LiDAR trackers~\cite{kummerle2015autonomous}. Additionally, we augment the dataset with hand-labeled annotation information in the form of intervals indicating the safety of the intersection for crossing. In this work, we further augmented the dataset with 8 extra sequences captured at different intersections. \figref{fig:FSCIntersecs} shows birds-eye-view images of some of the intersections captured in this dataset. For each sequence we provide the detected tracks from the radar and LiDAR trackers (\figref{fig:mpDatasets}(d)), along with the RGB camera images (\figref{fig:tlrDatasets}(e,f)) and the intersection safety for crossing. Furthermore, we provide annotations of the camera images regarding the state of the traffic light $S = \left\lbrace \textit{Red, Green, Off} \right\rbrace$.

Several factors make benchmarking on this dataset extremely challenging for motion prediction and traffic light recognition tasks including large number of traffic participants, varying motion dependencies among different dynamic objects (\figref{fig:mpDatasets}(d)), motion blur in the images, presence of reflecting surfaces and varying lighting conditions as shown in \figref{fig:tlrDatasets}(e,f). The dataset covers a wide range of road topologies and intersection types as depicted in~\figref{fig:FSCIntersecs}, which makes benchmarking on this dataset extremely challenging. To the best of our knowledge, this is the first dataset with multitask labels for motion prediction, traffic light recognition and intersection safety prediction. During training, we apply a leave-one-out procedure for the motion prediction task by randomly selecting trajectories from all sequences except the testing sequence. However, for the traffic light recognition task, we divide the data into a $4:1$ split, and apply random brightness and contrast modulations as an augmentation procedure for the training images.

\subsection{Training Schedule}
In the following, we describe the training procedure used for each of the motion prediction, traffic light recognition and intersection safety prediction tasks. In order to train our \mbox{IA-TCNN} model such that it is robust to the varying number of pedestrians observable in each interval, we introduce a variable to represent the maximum number of distinct trajectories observed within an interval and initially set it to the maximum observed in all the datasets. During training and testing, we use an activation mask to encode the positions of valid trajectories and discard all remaining information. We train our model for $100$ epochs with a mini-batch size of $12$. We employ the Adam solver~\citep{kingma2014adam} for optimization, with a learning rate of $\num{5e-4}$ and apply gradient clipping. Details regarding the sequence length used for training and the observation and prediction lengths used for testing are covered in~\secref{subsec:ablationMP}.

We train our AtteNet model for traffic light recognition on random crops of size $224\times224$ and test on the center crop which we found adds more regularization to the network and helps learning a more generalized model. We use the SGD solver with momentum to optimize our AtteNet model, using an initial learning rate of $\num{4e-3}$ and a polynomial weight decay of $\num{2e-4}$. We train our approach for $100$ epochs using a mini-batch size of $32$ and dropout probability of $0.2$. Note that both IA-TCNN and AtteNet are trained from scratch on their respective tasks for each of the aforementioned datasets. In order to learn the final model for predicting the intersection safety for crossing, we initially bootstrap the training of both IA-TCNN and AtteNet using transfer learning from the respective optimization procedures. We combine each of the task-specific loss functions using learnable weighting parameters and use a single optimizer to train all subnetworks concurrently. Training all tasks jointly aims at finding optimal weights that satisfy the constraints of each task as well as their interdependencies. Moreover, employing learnable weighting parameters ensures the proper balancing between the distinct tasks. We employ the Adam optimizer with an initial learning rate of $\num{5e-5}$. The final model is trained for $100$ epochs with a mini-batch size of $10$. All experiments are conducted using the Tensorflow library~\citep{tensorflow2015-whitepaper} on a single Nvidia Titan X GPU.

\subsection{Evaluation of Motion Prediction}
\label{sec:mopred}

\subsubsection{Comparison with the State-of-the-art}
\label{sec:sotaMP}\mbox{}\\
We benchmark the performance of our approach against several state-of-the-art methods for motion prediction including Social-LSTM~\citep{alahi2016social}, Social-Attention~\citep{vemula2017social}, Pose-LSTM~\citep{sun20173dof}, SGAN~\citep{gupta2018social} and SoPhie~\citep{sadeghian2018sophie}. Furthermore, we compare against the Social Forces model~\citep{helbing1995social} and a basic LSTM implementation as baselines. Note that for each of the methods, we report the numbers directly from the corresponding manuscripts, with the exception of the Social Forces model where we report the numbers from~\cite{alahi2016social} as the original manuscript does not report evaluations using the metrics employed by the aforementioned methods. Furthermore, we use our own implementation for the LSTM baseline. We evaluate the accuracy of our motion prediction model by reporting the following metrics:
\begin{itemize}
\item \textit{Average Displacement Error}: mean squared error over all predicted and groundtruth points in the trajectory.
\item \textit{Final Displacement Error}: distance between the predicted and groundtruth poses at the end of the prediction interval.
\end{itemize}

\begin{table*}
\footnotesize 
\centering
\caption{Average Displacement Accuracy of IA-TCNN on the task of motion prediction in comparison to existing methods on the L-CAS dataset.}
\label{tab:ada_lcas}
\begin{tabular}{p{1.2cm}p{1.5cm}p{1.5cm}p{1.5cm}p{1.5cm}|p{1.5cm}}
\hline\noalign{\smallskip}
Dataset & Social-LSTM & Pose-LSTM & IA-LinConv & IA-DResTCNN & IA-TCNN (Ours)\\
\noalign{\smallskip}\hline\hline\noalign{\smallskip}
L-CAS & $1.19\meter,\, \textsc{nan}$& $0.95\meter,\, 35.0\degree$ & $0.34\meter,\, 23.8\degree$ & $0.46\meter,\, 33.1\degree$ & $\mathbf{0.11\meter,\, 21.7\degree}$\\
\noalign{\smallskip}\hline\noalign{\smallskip}
\end{tabular}
\end{table*}

\begin{table*}
\footnotesize 
\centering
\caption{Average Displacement Accuracy of IA-TCNN on the task of motion prediction in comparison to existing methods on the ETH and UCY datasets.}
\label{tab:ada}
\begin{tabular}{p{1.4cm}p{1.cm}p{1.cm}p{1.cm}p{1.cm}p{1.cm}p{1.cm}p{1.2cm}p{1.4cm}p{1.4cm}|p{1.2cm}}
\hline\noalign{\smallskip}
Dataset & Social Forces & Basic LSTM & Social-LSTM & Social-Attention & SGAN & SoPhie & IA-LinConv & IA-DResTCNN & IA-TCNN (fixed length) & IA-TCNN (Ours)\\
\noalign{\smallskip}\hline\hline\noalign{\smallskip}
ETH-Univ & $0.41\meter$& $0.39\meter$ & $0.50\meter$ & $0.39\meter$ & $0.81\meter$ & $0.70\meter$ & $0.27\meter$ & $0.43\meter$ & $0.23\meter$ & $\mathbf{0.15\meter}$\\
ETH-Hotel & $0.25\meter$ & $0.32\meter$ & $\mathbf{0.11\meter}$ & $0.29\meter$ & $0.72\meter$ & $0.76\meter$ & $0.28\meter$ & $0.36\meter$ & $0.24\meter$ & $0.16\meter$ \\
Zara01 & $0.40\meter$ & $0.18\meter$ & $0.22\meter$ & $0.20\meter$ & $0.34\meter$ & $0.30\meter$ & $0.34\meter$ & $0.45\meter$ & $0.16\meter$ & $\mathbf{0.14\meter}$ \\
Zara02 & $0.40\meter$ & $0.28\meter$ & $0.25\meter$ & $0.30\meter$ & $0.42\meter$ & $0.38\meter$ & $0.38\meter$ & $0.37\meter$ & $0.17\meter$ & $\mathbf{0.19\meter}$\\
UCY-Uni & $0.48\meter$ & $0.30\meter$ & $\mathbf{0.27\meter}$ & $0.33\meter$ & $0.60\meter$ & $0.54\meter$ & $0.41\meter$ & $0.36\meter$ & $0.31\meter$ & $0.29\meter$\\
\noalign{\smallskip}\hline\noalign{\smallskip}
Average & $0.39\meter$ & $0.29\meter$ & $0.27\meter$ & $0.30\meter$ & $0.58\meter$ & $0.54\meter$ & $0.34\meter$ & $0.39\meter$ & $0.22\meter$ & $\mathbf{0.19\meter}$\\
\noalign{\smallskip}\hline\noalign{\smallskip}
\end{tabular}
\end{table*}

On the L-CAS, ETH and UCY datasets, we follow the standard evaluation procedure~\citep{sun20173dof, alahi2016social} of training using a sequence length of 20 frames and using observation and prediction lengths of 8 frames ($3.2\second$) and 12 frames ($4.8\second$) respectively during testing. Note that the observation length is implemented as a sliding window therefore the robot does not have to wait for a certain amount of time in order to be able to make a prediction. Only in the beginning of the system initialization, the buffer of the window needs to be accumulated and subsequently the prediction can be made as fast as new location information of the surrounding agents are available. The caveat of this initialization time is that the robot has to be started in a safe area for the duration of $5\second$ before it begins to navigate. We opted to use the same sequence length as existing methods in order to facilitate a fair comparison and avoid any performance bias induced by increasing or decreasing the observation interval. We follow this setting for all the datasets. Additionally, we also present results with varying observation and prediction lengths in \tabref{tab:varyingObsPredsUcy}. \tabref{tab:ada_lcas} shows the average displacement accuracy of our approach on the L-CAS dataset. As demonstrated by the results, both our proposed variants, IA-LinConv and IA-DResTCNN are able to outperform the standard recurrent-based approaches by $64.2\%$ and $32.0\%$ in the translational and rotational components respectively. This in turn corroborates the advantage of utilizing a causal convolutional architecture over the standard recurrent methods. Moreover, by utilizing our proposed IA-TCNN, we achieve an average displacement error of $0.11\meter$ and $21.7\degree$ further improving upon the results by $67.6\%$ and $8.8\%$ in translation and rotation respectively. The improvement over the results achieved by IA-LinConv is attributed to employing dilated convolutions which increase the size of the receptive field, thereby increasing the content captured in each layer. However, we observe that adding a residual block to our network to improve the feature discriminability, as in IA-DResTCNN, does not help in improving the prediction accuracy despite it being helpful for other sequence modeling tasks such as language modeling~\citep{bai2018empirical}. We speculate that this model tends to accumulate more temporal information than the remaining methods through skip connections which has an adverse effect of adding noise and eventually degrading the quality of the predicted motion trajectories.

%Concurrently, incorporating too much information through increasing the depth of the network, as in IA-DResTCNN, can have adverse effects on the prediction accuracy.
In~\tabref{tab:ada}, we present the average displacement accuracy of our proposed methods in comparison to state-of-the-art approaches on different sequences from the ETH and UCY crowd set datasets. Due to the complexity of the pedestrian interactions demonstrated in this dataset, employing the \mbox{IA-LinConv} model does not yield significant improvement over recurrent-based approaches due to the small receptive field at each layer. By employing dilated convolutions in our \mbox{IA-TCNN} architecture, the network is able to better capture the interactions across the various pedestrians, thereby achieving an improvement of $29.6\%$ in comparison to the previous state-of-the-art. Similar to IA-LinConv, the accuracy of IA-DResTCNN is comparable to that of recurrent-based approaches. However, the convergence time of the network is $5\times$ more than IA-LinConv and IA-TCNN. We additionally compare the performance of our IA-TCNN architecture with a fixed trajectory length variant (IA-TCNN fixed length). The fixed trajectory length variant was trained using only trajectories of pedestrians that have a minimum length equal to the sequence length (20 frames). We compare this variant with our dynamic length trained architecture to investigate if using binary activation masks has an impact over the performance of our method. The results show that as hypothesized using binary activation masks during training enables our network to learn that not all pedestrians must be observed for the same amount of time which leads to an overall improvement in the accuracy of the trained model.

We report the final displacement accuracy of our proposed IA-TCNN on the various sequences of the ETH and UCY datasets in~\tabref{tab:fda}. Despite the sparse amount of sequences available for each dataset and the complexity of the pedestrian interactions demonstrated, our method is able to achieve the lowest final displacement error on all sequences of the ETH and UCY datasets with an average improvement of $55.7\%$ in comparison to previous methods. It is worth noting that while other approaches incorporate information about nearby pedestrians or the surrounding environment to predict the trajectories, our proposed method is able to accurately infer the trajectories, surpassing the performance of state-of-the-art methods, without leveraging information about the structure of the scene or performing any pre/postprocessing on the trajectory data.

\begin{table*}
\footnotesize 
\centering
\caption{Final Displacement Accuracy of IA-TCNN on the task of motion prediction in comparison to existing methods on the ETH and UCY datasets.}
\label{tab:fda}
\begin{tabular}{p{1.4cm}p{1.2cm}p{1.2cm}p{1.2cm}p{1.2cm}p{1.2cm}p{1.2cm}p{1.2cm}p{1.4cm}|p{1.2cm}}
\hline\noalign{\smallskip}
Dataset & Social Forces & Basic LSTM & Social-LSTM & Social-Attention & SGAN & SoPhie & IA-LinConv & IA-DResTCNN & IA-TCNN (Ours)\\
\noalign{\smallskip}\hline\hline\noalign{\smallskip}
ETH-Univ & $0.59\meter$& $1.06\meter$ & $1.07\meter$ & $3.74\meter$ & $1.52\meter$ & $1.43\meter$ & $0.27\meter$ & $0.60\meter$ & $\mathbf{0.21\meter}$\\
ETH-Hotel & $0.37\meter$ & $0.33\meter$ & $0.23\meter$ & $2.64\meter$ & $1.61\meter$ & $1.67\meter$ & $0.32\meter$ & $0.52\meter$ & $\mathbf{0.18\meter}$\\
Zara01 & $0.60\meter$ & $0.93\meter$ & $0.48\meter$ & $0.52\meter$ & $0.69\meter$ & $0.63\meter$ & $0.54\meter$ & $1.08\meter$ & $\mathbf{0.27\meter}$\\
Zara02 & $0.68\meter$ & $1.09\meter$ & $0.50\meter$ & $2.13\meter$ & $0.84\meter$ & $0.78\meter$ & $0.47\meter$ & $0.88\meter$ & $\mathbf{0.25\meter}$\\
UCY-Uni & $0.78\meter$ & $1.25\meter$ & $0.77\meter$ & $3.92\meter$ & $1.26\meter$ & $1.24\meter$ & $0.66\meter$ & $1.03\meter$ & $\mathbf{0.46\meter}$\\
\noalign{\smallskip}\hline\noalign{\smallskip}
Average & $0.60\meter$ & $0.93\meter$ & $0.61\meter$ & $2.59\meter$ & $1.18\meter$ & $1.15\meter$ & $0.45\meter$ & $0.82\meter$ & $\mathbf{0.27\meter}$\\
\noalign{\smallskip}\hline\noalign{\smallskip}
\end{tabular}
\end{table*}

We benchmark on the Freiburg Street Crossing (FSC) dataset largely due to the variety of motion trajectories and complex interactions. Furthermore, unlike the ETH, UCY and L-CAS datasets, the FSC dataset includes trajectories and interactions among various types of dynamic objects such as cyclists, vehicles and pedestrians which in turn both increases the difficulty of the prediction task, as well as renders the data more representative of real-world scenarios. On the FSC dataset, we train using a sequence length of $10\second$ and use observation and prediction lengths of $5\second$. As the radar sensor has a larger field-of-view than the LiDAR, and in order to observe the traffic participants in both sensors, we experimentally identified that a time window of $10\second$ is appropriate for correlating objects in both sensors on this dataset. We report the average displacement accuracy of our proposed method in~\tabref{tab:ada_fsc}. The results show an improvement in employing IA-LinConv and \mbox{IA-DResTCNN} over the LSTM baseline, specifically in terms of rotation and velocity estimation. We attribute this to the increased complexity of the interactions demonstrated in this dataset, in addition to the presence of multiple types of dynamic objects which exhibit different interaction and motion behavior. Furthermore, we observe that by employing the IA-DResTCNN architecture, the rotational accuracy of the pose is further improved in comparison to the IA-LinConv architecture. We attribute this improvement partially to the bigger receptive field at each layer due to the dilation factor employed. The best performance is achieved by leveraging the proposed IA-TCNN architecture which is able to balance the motion specific pose components for each dynamic object independent of their type, yielding an average displacement error of $0.20\meter,\, 6.71\degree$ and $0.84\si{\meter\per\second}$.

\begin{table*}
\footnotesize 
\centering
\caption{Average Displacement Accuracy of our motion prediction network IA-TCNN on the Freiburg Street Crossing (FSC) dataset.}
\label{tab:ada_fsc}
\begin{tabular}{p{1.4cm}p{3.5cm}p{3.5cm}p{3.5cm}|p{3.5cm}}
\hline\noalign{\smallskip}
Dataset & Basic LSTM & IA-LinConv & IA-DResTCNN & IA-TCNN (Ours)\\
\noalign{\smallskip}\hline\hline\noalign{\smallskip}
Seq.-1 & $0.37\meter,\, 10.34\degree,\, 0.62\si{\meter\per\second}$ & $0.22\meter,\, 8.66\degree,\, 0.38\si{\meter\per\second}$ & $0.49\meter,\, \mathbf{6.25\degree},\, 0.48\si{\meter\per\second}$ & $\mathbf{0.21\meter},\, 6.95\degree,\, 0.43\si{\meter\per\second}$\\
Seq.-2 & $0.28\meter,\, 13.15\degree,\, 0.74\si{\meter\per\second}$ & $0.48\meter,\, 9.69\degree,\, 0.55\si{\meter\per\second}$ & $0.64\meter,\, 8.57\degree,\, 0.68\si{\meter\per\second}$ &  $\mathbf{0.28\meter},\, \mathbf{8.51\degree,\, 0.41\si{\meter\per\second}}$\\
Seq.-3 & $0.37\meter,\, 20.68\degree,\, 0.86\si{\meter\per\second}$ & $\mathbf{0.46\meter},\, 14.24\degree,\, \mathbf{0.80\si{\meter\per\second}}$ & $0.85\meter,\,
\mathbf{12.46\degree},\, 1.01\si{\meter\per\second}$ & $0.48\meter,\, 13.43\degree,\, 0.91\si{\meter\per\second}$\\
Seq.-4 & $0.27\meter,\, 16.54\degree,\, 0.85\si{\meter\per\second}$ & $0.26\meter,\, 6.56\degree,\, 0.63\si{\meter\per\second}$ & $0.46\meter,\, 5.87\degree,\, 0.73\si{\meter\per\second}$ & $\mathbf{0.26\meter,\, 4.52\degree,\, 0.60\si{\meter\per\second}}$\\
Seq.-5 & $0.32\meter,\, 8.17\degree,\, 1.26\si{\meter\per\second}$ & $0.21\meter,\, 7.40\degree,\, \mathbf{0.40\si{\meter\per\second}}$ & $0.42\meter,\, 7.20\degree,\, 0.48\si{\meter\per\second}$ & $\mathbf{0.21\meter,\, 6.83\degree},\, 0.45\si{\meter\per\second}$\\
Seq.-6 & $\mathbf{0.26\meter},\, 20.53\degree,\, \mathbf{1.98\si{\meter\per\second}}$ & $0.38\meter,\, 16.37\degree,\, 2.26\si{\meter\per\second}$ & $0.65\meter,\, \mathbf{8.23\degree},\, 2.27\si{\meter\per\second}$ & $0.31\meter,\, 16.72\degree,\, 2.08\si{\meter\per\second}$\\
Seq.-7 & $0.24\meter,\, 3.29\degree,\, 0.69\si{\meter\per\second}$ & $0.09\meter,\, 2.00\degree,\, 0.28\si{\meter\per\second}$ & $0.11\meter,\, 1.58\degree,\, 0.29\si{\meter\per\second}$ & $\mathbf{0.05\meter,\, 1.43\degree,\, 0.23\si{\meter\per\second}}$\\
Seq.-8 & $0.35\meter,\, 6.56\degree,\, 1.23\si{\meter\per\second}$ & $\mathbf{0.16\meter},\, 5.90\degree,\, 1.08\si{\meter\per\second}$ & $0.40\meter,\, \mathbf{2.92\degree},\, 1.34\si{\meter\per\second}$ & $0.21\meter,\, 3.60\degree,\, \mathbf{1.06\si{\meter\per\second}}$\\
Seq.-9 & $0.31\meter,\, 7.65\degree,\, 1.10\si{\meter\per\second}$ & $\mathbf{0.16\meter},\, 8.57\degree,\, 1.06\si{\meter\per\second}$ & $0.38\meter,\, \mathbf{2.70\degree},\, 1.35\si{\meter\per\second}$ & $0.18\meter,\, 3.19\degree,\, \mathbf{1.00\si{\meter\per\second}}$\\
Seq.-10 & $0.37\meter,\,	6.55\degree,\, 0.90\si{\meter\per\second}$ & $0.25\meter,\, 2.80\degree,\, 0.98\si{\meter\per\second}$ & $0.28\meter,\, 2.50\degree,\, 1.00\si{\meter\per\second}$ & $\mathbf{0.14\meter},\, \mathbf{2.42\degree,\, 0.78\si{\meter\per\second}}$\\
Seq.-11 & $0.24\meter,\, 2.72\degree, 0.72\si{\meter\per\second}$ & $0.24\meter,\, 2.03\degree,\, \mathbf{0.57\si{\meter\per\second}}$ & $0.28\meter,\, \mathbf{1.15\degree},\, 0.61\si{\meter\per\second}$ & $\mathbf{0.17\meter},\, 1.80\degree,\, 0.64\si{\meter\per\second}$\\
Seq.-12 & $0.28\meter,\, 4.96\degree,\, \mathbf{1.14\si{\meter\per\second}}$ & $0.21\meter,\, 2.93\degree,\, 1.38\si{\meter\per\second}$ & $0.35\meter,\, \mathbf{1.94\degree},\, 1.38\si{\meter\per\second}$ & $\mathbf{0.19\meter},\, 2.94\degree,\, 1.21\si{\meter\per\second}$\\
Seq.-13 & $0.30\meter,\, 8.51\degree,\, \mathbf{0.79\si{\meter\per\second}}$ & $0.32\meter,\, 7.75\degree,\, 1.85\si{\meter\per\second}$ & $0.60\meter,\, \mathbf{5.90\degree},\, 2.14\si{\meter\per\second}$ & $\mathbf{0.28\meter},\, 6.11\degree,\, 1.64\si{\meter\per\second}$\\
Seq.-14 & $0.32\meter,\, 6.00\degree,\, 1.13\si{\meter\per\second}$ & $0.25\meter,\, 4.92\degree,\, 0.69\si{\meter\per\second}$ & $0.35\meter,\, \mathbf{4.01\degree},\, 0.72\si{\meter\per\second}$ & $\mathbf{0.17\meter},\, 4.04\degree,\, \mathbf{0.53\si{\meter\per\second}}$\\
Seq.-15 & $0.39\meter,\, 3.31\degree,\, 0.79\si{\meter\per\second}$ & $0.19\meter,\, 2.34\degree,\, 0.60\si{\meter\per\second}$ & $0.22\meter,\, 2.09\degree,\, 0.64\si{\meter\per\second}$ & $\mathbf{0.14\meter,\, 1.64\degree,\, 0.48\si{\meter\per\second}}$ \\
Seq.-16 & $0.34\meter,\, 14.69\degree,\, 1.51\si{\meter\per\second}$ & $0.13\meter,\, 15.84\degree,\, 1.32\si{\meter\per\second}$ & $0.34\meter,\, 16.99\degree,\, 1.30\si{\meter\per\second}$ & $\mathbf{0.10\meter,\, 10.33\degree,\, 1.20\si{\meter\per\second}}$\\
Seq.-17 & $0.36\meter,\, 33.42\degree,\, 0.83\si{\meter\per\second}$ & $0.17\meter,\, 23.06\degree,\, \mathbf{0.73\si{\meter\per\second}}$ & $0.44\meter,\, 26.86\degree,\, 0.86\si{\meter\per\second}$ & $\mathbf{0.14\meter,\, 13.69\degree},\, 0.83\si{\meter\per\second}$\\
Seq.-18 & $0.32\meter,\, 23.77\degree,\, 0.94\si{\meter\per\second}$ & $0.12\meter,\, 16.15\degree,\, 0.92\si{\meter\per\second}$ & $0.32\meter,\, 17.44\degree,\, 0.91\si{\meter\per\second}$ & $\mathbf{0.12\meter,\, 12.62\degree,\, 0.75\si{\meter\per\second}}$\\
\noalign{\smallskip}\hline\noalign{\smallskip}
Average & $0.32\meter,\, 11.71\degree,\, 0.95\si{\meter\per\second}$ & $0.24\meter,\,  8.73\degree,\, 0.92\si{\meter\per\second}$ & $0.42\meter,\, 7.48\degree,\, 1.01\si{\meter\per\second}$ & $\mathbf{0.20\meter,\, 6.71\degree,\, 0.84\si{\meter\per\second}}$\\
\noalign{\smallskip}\hline\noalign{\smallskip}
\end{tabular}
\end{table*}

\subsubsection{Ablation Study \& Qualitative Evaluation}
\label{subsec:ablationMP}\mbox{}\\
In this section, we perform detailed studies on the influences of various components of our proposed architecture. Note that for all experiments, we use the same data-sepcific frame rate as in~\secref{sec:sotaMP} to facilitate ease of comparison. \tabref{tab:varyingObsPredsUcy} shows the effect of varying the observation and prediction lengths on the average displacement accuracy of IA-TCNN on the Uni sequence of the UCY crowd set dataset. For short observation lengths \mbox{($2-4$)~frames}, the error in the predicted trajectory linearly increases with the increase in the prediction length, with the lowest error achieved using a prediction interval smaller than or equal to the observation interval. This accounts for the increased difficulty of making accurate predictions given short trajectory information as future interactions cannot be reliably predicted. Concurrently, by increasing the observation length, the prediction accuracy gradually increases with small improvements between \mbox{($6-8$)} observation frames. This can be attributed to the reduction in the amount of significant information over time due to the short interaction times between the pedestrians and the low likelihood of abrupt changes in the behavior of one or more pedestrians. %The results further validate the constraint on using a prediction length greater than or equal to the observation interval in order to make the prediction problem more challenging.  

\begin{figure*}
\footnotesize 
\centering 
\setlength{\tabcolsep}{0.2cm} 
\begin{tabular}{p{0.1cm} p{3cm} p{3cm} p{3cm} p{3cm}}
\rotatebox[origin=c]{90}{Observation} & \raisebox{-0.5\height}{\includegraphics[width=1\linewidth]{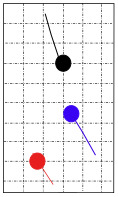}} &
\raisebox{-0.5\height}{\includegraphics[width=1\linewidth]{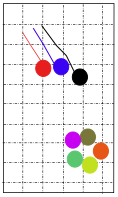}} & \raisebox{-0.5\height}{\includegraphics[width=1\linewidth]{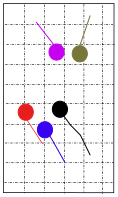}} & \raisebox{-0.5\height}{\includegraphics[width=1\linewidth]{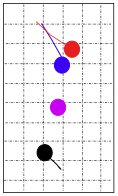}}\\
\\
\rotatebox[origin=c]{90}{Prediction} & \raisebox{-0.5\height}{\includegraphics[width=1\linewidth]{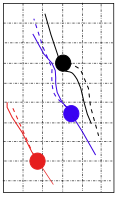}} &
\raisebox{-0.5\height}{\includegraphics[width=1\linewidth]{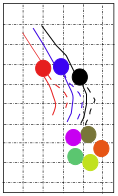}} & \raisebox{-0.5\height}{\includegraphics[width=1\linewidth]{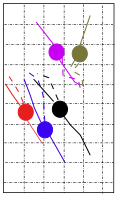}} & \raisebox{-0.5\height}{\includegraphics[width=1\linewidth]{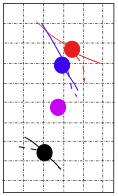}}\\
\\
& \multicolumn{1}{c}{(a)} & \multicolumn{1}{c}{(b)} & \multicolumn{1}{c}{(c)} & \multicolumn{1}{c}{(d)}
\end{tabular} 
\caption{Qualitative analysis of our interaction-aware motion prediction IA-TCNN network on four example sequences from the UCY-UNI dataset. The top figures show the observed trajectories for all the pedestrians, along with the current pedestrian position marked by the colored circle. The bottom figures show the groundtruth (solid lines) and predicted (dashed lines) trajectories for each corresponding figure. Scenarios (a) and (c) show collision avoidance, while (b) and (d) show group and social interactions. Our approach is able to accurately model the pedestrian interactions and group behavior in each of the different scenarios presented.} 
\label{fig:pathPredUCYUni}
\end{figure*}

In~\tabref{tab:orientVeloMP}, we evaluate the effect of incorporating orientation and velocity information on the accuracy of the predicted motion trajectories on the FSC dataset. We compare the performance of the following variants:
\begin{itemize}
\item LSTM: Basic LSTM architecture predicting the future position of each agent.
\item MP-M1: Predicting only the position of each agent.
\item MP-M2: Predicting the position and orientation for each agent.
\item MP-M3: Predicting the position, orientation and velocity of each agent, corresponding to our proposed IA-TCNN architecture.
\end{itemize}

\begin{table*}
\footnotesize 
\centering
\caption{Effect of predicting the orientation and velocity information on the average displacement error for the task of motion prediction on the Freiburg Street Crossing (FSC) dataset. We report the average performance on all of the sequences.}
\label{tab:orientVeloMP}
\begin{tabular}{p{2.5cm}p{2.7cm}p{2.2cm}p{3.0cm}}
\hline\noalign{\smallskip}
Model & Orientation & Velocity & Avg. Disp. Error\\
\noalign{\smallskip}\hline\hline\noalign{\smallskip}
LSTM & \xmark & \xmark & $0.40\meter$ \\
MP-M1 & \xmark & \xmark & $0.25\meter$ \\
MP-M2 & \cmark & \xmark & $0.22\meter,\, 12.09\degree$ \\
MP-M3 (IA-TCNN) & \cmark & \cmark & $\mathbf{0.20\meter,\, 6.71\degree,\, 0.84\si{\meter\per\second}}$\\
\noalign{\smallskip}\hline\noalign{\smallskip}
\end{tabular}
\end{table*}

Employing our proposed architecture with causal convolutions results in an improvement of $40.0\%$ in the average displacement error over a standard LSTM-based approach. Incorporating the orientation information of each agent, further improves the results by $12.0\%$ over the MP-M1 model. This improvement corroborates the utility of predicting the orientation information of each agent in the overall accuracy of the predicted poses. This occurs as a direct consequence of the fact that most real-world agents rarely turn on the spot, but rather slowly change their orientation along the path. Hence incorporating the orientation information enables the model to predict more accurate trajectories in such situations. Finally, incorporating the velocity information further improves the average displacement error by $9.01\%$ over the MP-M2 model. Despite the fact that velocity can be approximated through the difference in the predicted positions over the time interval, we believe that directly predicting the velocity provides a more accurate information source specifically in highly dynamic scenarios such as street intersections and busy roads where the agents can quickly vary their speed. In such situations, the velocity obtained from computing the difference in positions over the time interval can be often too inaccurate for reliable estimation. However, directly predicting the velocity enables the network to benefit from this information and hence enables robust trajectory estimates in various environments.

We further investigate the effect of the various architectural choices leading to the development of our IA-TCNN architecture for predicting interaction-aware motion trajectories in~\tabref{tab:networkArchMP}. More precisely, we investigate the average displacement accuracies of the following models on the FSC dataset:
\begin{itemize}
\item MP-M31: Basic LSTM-based architecture for motion prediction with fixed equal weights assigned to the translation and orientation components of the pose along with the speed.
\item MP-M32: Causal convolutional architecture with fixed equal weights for the pose and speed components.
\item MP-M3: Causal convolutional architecture with adaptive learnable weights for the various pose and speed components. This architecture corresponds to the proposed IA-TCNN model.
\end{itemize}

The results in \tabref{tab:networkArchMP} show that utilizing causal convolutions for temporal feature aggregation results in an improvement of $25.0\%,\, 8.37\%,\, 4.21\%$  in the translation, orientation and speed respectively over employing LSTMs. This improvement validates our hypothesis that employing causal convolutions over basic recurrent blocks such as LSTMs enables the network to better leverage the temporal information, thereby improving the quality of the predicted poses. Utilizing adaptive weighting parameters to learn the optimal weight for the translation and orientation components of the pose results in a further improvement of $16.67\%,\, 37.47\%,\,7.69\%$ in the average displacement accuracy in terms of translation, orientation and speed respectively. By employing adaptive weights that vary during the optimization procedure, the most favorable set of weights is found that enables the optimal balancing between the various components and preventing a single term from dominating the loss without benefiting the remaining components.

\begin{table*}
\footnotesize 
\centering
\caption{Effect of the various architectural choices on the average displacement error for the task of motion prediction on the Freiburg Street Crossing (FSC) dataset. We report the average performance on all of the sequences.}
\label{tab:networkArchMP}
\begin{tabular}{p{2.5cm}p{2.7cm}p{2.2cm}p{3.0cm}}
\hline\noalign{\smallskip}
Model & Temporal Block & Loss Weights & Avg. Disp. Error\\
\noalign{\smallskip}\hline\hline\noalign{\smallskip}
MP-M31 & LSTM & fixed equal & $0.32\meter,\, 11.71\degree,\, 0.95\si{\meter\per\second}$ \\
MP-M32 & Causal Convolution & fixed equal & $0.24\meter,\, 10.73\degree,\, 0.91\si{\meter\per\second}$ \\
MP-M3 (IA-TCNN) & Causal Convolution & adaptive & $\mathbf{0.20\meter,\, 6.71\degree,\, 0.84\si{\meter\per\second}}$\\
\noalign{\smallskip}\hline\noalign{\smallskip}
\end{tabular}
\end{table*}

\begin{table*}
\footnotesize 
\centering
\caption{Effect of the varying the observation and prediction lengths in frames on the average displacement error for our proposed IA-TCNN method on the task of motion prediction on the UCY-Uni dataset.}
\label{tab:varyingObsPredsUcy}
\begin{tabular}{c|ccccccccccc}
\hline\noalign{\smallskip}
\backslashbox{Obs. Length}{Pred. Length}& $2$ & $3$ & $4$  & $5$ & $6$ & $7$ & $8$ & $9$ & $10$  & $11$ & $12$\\
\noalign{\smallskip}\hline\hline\noalign{\smallskip}
$2$ & $\mathbf{0.42\meter}$ & $0.48\meter$ & $0.50\meter$ & $0.53\meter$ & $0.56\meter$ & $0.61\meter$ & $0.63\meter$ & $0.62\meter$ & $0.62\meter$ & $0.61\meter$ & $0.60\meter$ \\
$3$ & $\mathbf{0.31\meter}$ & $0.36\meter$ & $0.41\meter$ & $0.45\meter$ & $0.49\meter$ & $0.52\meter$ & $0.52\meter$ & $0.51\meter$ & $0.51\meter$ & $0.51\meter$ & $0.51\meter$ \\
$4$ & $\mathbf{0.23\meter}$ & $0.30\meter$ & $0.35\meter$ & $0.39\meter$ & $0.42\meter$ & $0.43\meter$ & $0.43\meter$ & $0.43\meter$ & $0.43\meter$ & $0.43\meter$ & $0.44\meter$ \\
$5$ & $\mathbf{0.20\meter}$ & $0.26\meter$ & $0.31\meter$ & $0.36\meter$ & $0.36\meter$ & $0.37\meter$ & $0.37\meter$ & $0.37\meter$ & $0.38\meter$ & $0.38\meter$ & $0.38\meter$ \\
$6$ & $\mathbf{0.18\meter}$ & $0.23\meter$ & $0.28\meter$ & $0.29\meter$ & $0.30\meter$ & $0.32\meter$ & $0.32\meter$ & $0.33\meter$ & $0.33\meter$ & $0.33\meter$ & $0.34\meter$ \\
$7$ & $\mathbf{0.15\meter}$ & $0.20\meter$ & $0.22\meter$ & $0.24\meter$ & $0.26\meter$ & $0.27\meter$ & $0.28\meter$ & $0.29\meter$ & $0.29\meter$ & $0.3	0\meter$ & $0.31\meter$ \\
$8$ & $\mathbf{0.14\meter}$ & $0.17\meter$ & $0.19\meter$ & $0.21\meter$ & $0.23\meter$ & $0.24\meter$ & $0.25\meter$ & $0.26\meter$ & $0.26\meter$ & $0.27\meter$ & $0.29\meter$\\
\noalign{\smallskip}\hline\noalign{\smallskip}
\end{tabular}
\end{table*}

In order to evaluate the performance of our proposed model in various types of interactions, we visualize four example scenes from the Uni sequence of the UCY dataset in~\figref{fig:pathPredUCYUni}. The top row shows the observation sequence for each pedestrian represented by a solid line with the current position of the pedestrian depicted by a circle. In the bottom row, we plot the groundtruth trajectory (solid line) and the predicted trajectory of the network (dashed line). \figref{fig:pathPredUCYUni}(a) presents a scenario with collision avoidance for two individuals. In this scenario, our IA-TCNN method is able to predict the temporary change in direction for both individuals to avoid collision. Our proposed model is also able to represent group behavior as shown in~\figref{fig:pathPredUCYUni}(b), where it predicts a common change of direction for all members of the group. \figref{fig:pathPredUCYUni}(c) shows a more complex scene with collision avoidance and overtaking maneuvers. The pedestrians depicted in red, blue and black display an example of the overtaking maneuver, where the red colored pedestrian is walking with a slightly lower velocity. Our model predicts that the blue colored pedestrian will adjust their trajectory to the right while increasing their velocity in order to overtake the red colored pedestrian. In order to avoid potential collision with the blue colored pedestrian, the model predicts a trajectory for the black colored pedestrian that is slightly deviated to the right. As for the purple and olive colored individuals, the model incorrectly predicts a trajectory where the olive colored pedestrian attempts to overtake the purple colored pedestrian. Whereas a more socially acceptable behavior, as shown by the groundtruth trajectory in this example would be to halt and wait for the purple colored pedestrian to pass.

\figref{fig:pathPredUCYUni}(d) shows another complex scenario with one static pedestrian in the middle, and a crossing maneuver between the red and blue colored pedestrians. In this example, our model predicts a trajectory where the red colored pedestrian follows the blue one. Note that while our approach incorporates the rotational information of the various dynamic objects into the prediction, on the ETH and UCY datasets, we do not utilize the information about the heading of each individual as this information is only available for one of the datasets, which further hinders it from being combined with the others. Nonetheless, we believe in such scenarios that information about the heading of each individual can significantly reduce the error in the predictions as shown by the results in~\tabref{tab:ada_lcas}, since sudden changes in direction are uncommon in the behavior of pedestrians.

\begin{table}
\footnotesize 
\centering
\caption{Comparison of the performance of our interaction-aware motion prediction network with the state-of-the-art on UCY-UNI dataset.}
\label{tab:runtime}
\begin{tabular}{p{2.1cm}p{1.2cm}p{1.4cm}p{1.1cm}p{0.5cm}}
\hline\noalign{\smallskip}
Method & Avg. Disp. Error ($\meter$) & Final Disp. Error ($\meter$)& Run-time ($\second$)& Size ($\mega\byte$)\\
\noalign{\smallskip}\hline\hline\noalign{\smallskip}
Basic LSTM & $0.30$ & $1.25$ & $0.29$ & $\mathbf{6.1}$ \\
Social-LSTM & $\mathbf{0.27}$ & $0.77$ & $1.78$ & $95.8$ \\
SGAN & $0.60$ & $1.26$ & $\mathbf{0.04}$ & $N/A$ \\
IA-TCNN (Ours) & $0.29$ & $\mathbf{0.46}$ & $0.06$ & $7.0$\\
\noalign{\smallskip}\hline\noalign{\smallskip}
\end{tabular}
\end{table}

We further compare the run time and model size of our approach with various recurrent based approaches in~\tabref{tab:runtime}. While the lowest average displacement error is achieved by the Social-LSTM approach~\citep{alahi2016social}, both the run-time and model size render it infeasible to be deployed in real-world scenarios. Similarly, while the SGAN method~\citep{gupta2018social} achieves fast run-time, it has the lowest average and final displacement accuracies among all methods. The results show that using our proposed IA-TCNN, we improve upon the final displacement accuracy by $40.3\%$ while achieving analogous average displacement error in comparison with the best performing model with a competitive run time of $0.06\second$ on a single Nvidia Titan X GPU. Moreover, our model requires only $7.0\mega\byte$ of storage space, thereby making it efficiently deployable in resource limited systems, while achieving accurate predictions in an online manner. It is worth noting that none of the aforementioned datasets contain significant amount of clutter or static obstacles in the scene. Furthermore, as our model does not incorporate the structural information of the scene in any manner, its performance is expected to vary depending on the amount of clutter in the environment. However, as highly cluttered scenes with multiple static obstacles were rarely encountered in the investigated scenarios, we do not tackle this in our current work. Our goal was rather to develop a dynamic interaction-aware motion prediction network that is capable of online performance in resource limited systems. For future work, we aim to additionally incorporate the structural information of the scene in order to achieve more reliable trajectory estimates.

\subsection{Evaluation of Traffic Light Recognition}
\subsubsection{Comparison with the State-of-the-art}\mbox{}\\
In this section, we evaluate the performance of our AtteNet on the task of traffic light recognition by benchmarking against several network architectures tailored for the aforementioned task namely SqueezeNet~\citep{iandola2016squeezenet}, DenseNet~\citep{huang2016densely} and ResNet~\citep{he2016deep}. We compare against the SqueezeNet architecture due to its relatively small size and high representational ability which enables it to be efficiently deployed in an online manner. This was demonstrated in the Nexar challenge where the first place winner used the SqueezeNet architecture achieving a recognition accuracy of $94.95\%$. Concurrently, we benchmark against DenseNet and ResNet architectures due to their top performance in various classification and regression tasks. We employ the ResNet-50 architecture with five residual blocks as a baseline. Similarly, we utilize the \mbox{DenseNet-121} architecture with four dense blocks and a growth-rate of 16. We quantify the performance of each architecture by reporting the prediction accuracy, precision and recall rates. Furthermore, we visualize the precision-recall plots for each traffic light state on each of the compared datasets and compare the performance of the benchmarked architecture against our proposed architecture with the goal of obtaining more insight into the performance of the individual approaches for each of the traffic light states. \tabref{tab:TLRecog} shows the classification accuracy of AtteNet on all three datasets; Nexar, Bosch and FSC. AtteNet outperforms all the baselines on each of the datasets by an average of $2.63\%$ which in turn validates the suitability of our proposed architecture for the task of traffic light recognition. Furthermore, by employing AtteNet, we are able to outperform the state-of-the-art on the Nexar challenge dataset.

\begin{figure*}
\footnotesize 
\centering 
\setlength{\tabcolsep}{0.1cm} 
\begin{tabular}{p{5cm} p{5cm} p{5cm}}
{\includegraphics[width=1\linewidth]{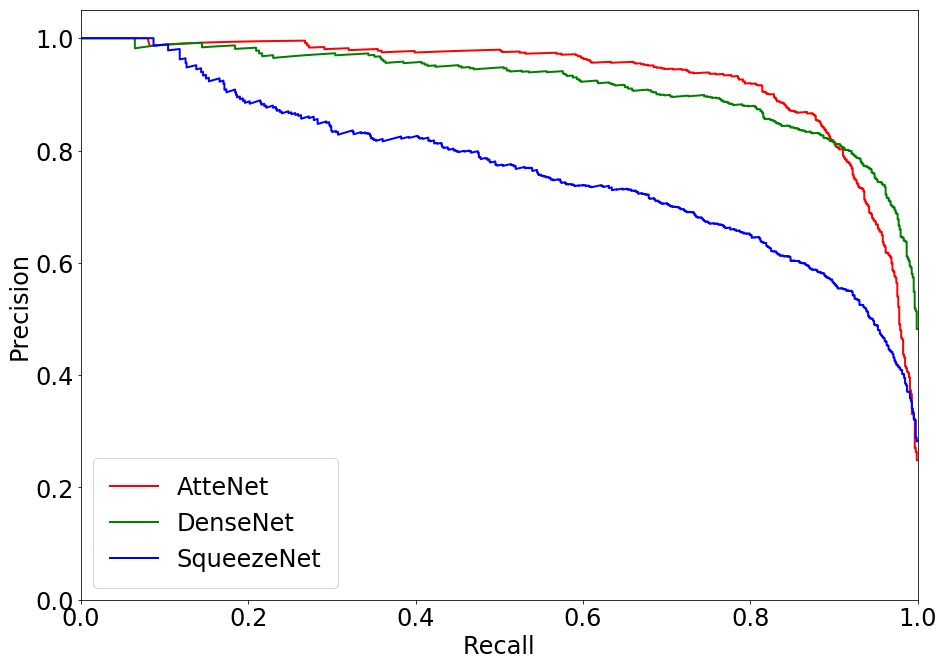}} &
{\includegraphics[width=1\linewidth]{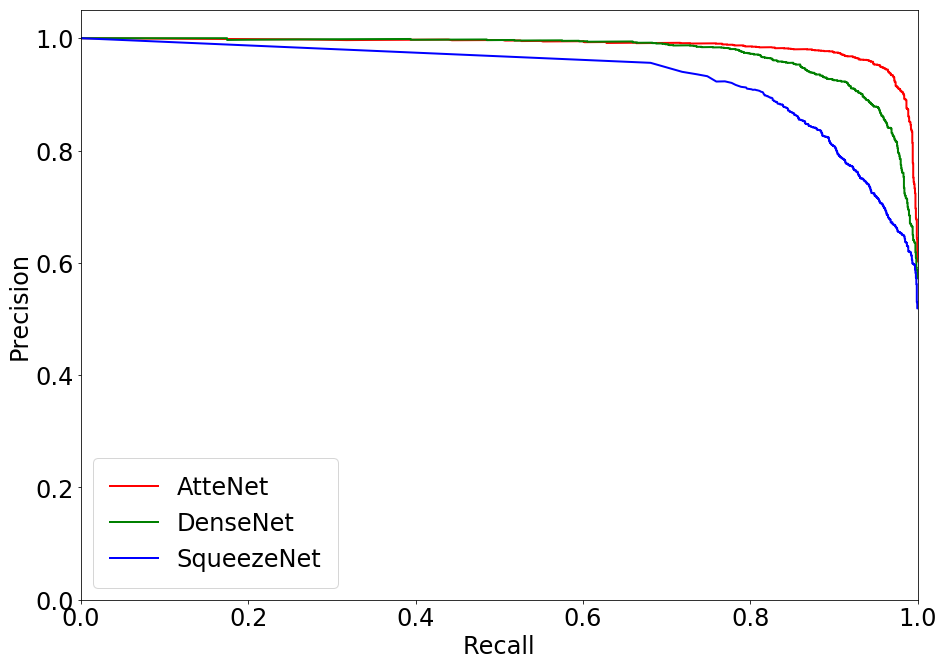}} &
{\includegraphics[width=1\linewidth]{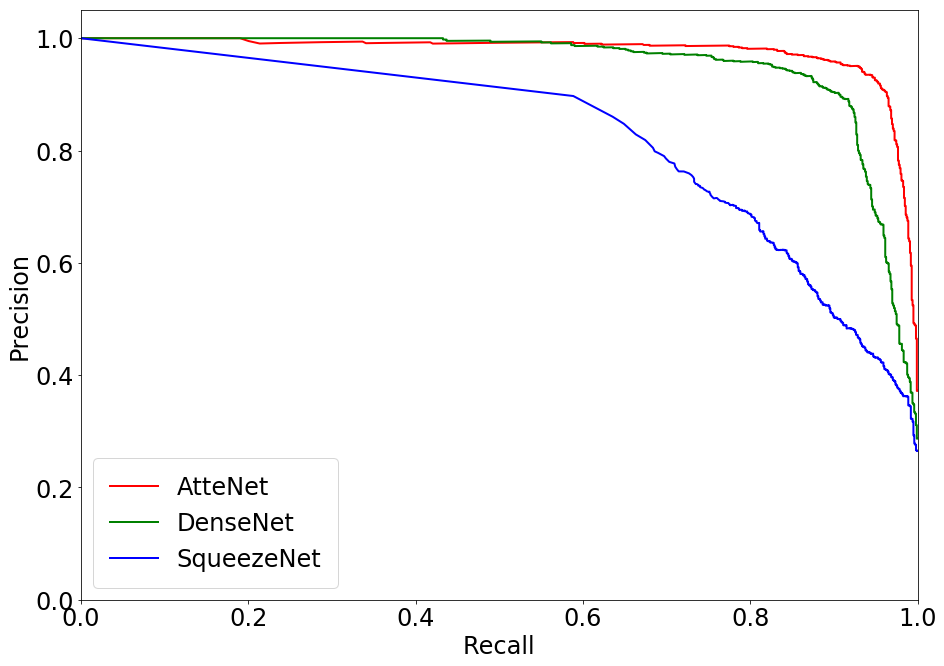}}
\\
\multicolumn{1}{c}{(a) Off} & \multicolumn{1}{c}{(b) Red} & \multicolumn{1}{c}{(c) Green} \\
\end{tabular} 
\caption{Precision-Recall plots comparing the performance of the various methods on the Nexar dataset. We visualize the curves for the three traffic light states: (a) Off, (b) Red and (c) Green. Our proposed AtteNet outperforms the compared approaches for all traffic light states.}
\label{fig:pr_nexar}
\end{figure*}

Analyzing the precision and recall rates for each class on the Nexar dataset in~\tabref{tab:nexarpr} shows that our proposed AtteNet is capable of accurately identifying the various traffic light signals with the highest recall despite the challenging lighting conditions demonstrated in the dataset. \figref{fig:pr_nexar} depicts the precision-recall curves for the compared approaches on the Nexar dataset. The results show that the AtteNet outperforms each of the compared methods on all traffic light states. This is further corroborated in~\figref{fig:tsne}(a) which plots the 3D t-Distributed Stochastic Neighbor Embedding \mbox{(t-SNE)}~\citep{maaten2008visualizing} of the features learned by our proposed AtteNet on the Nexar dataset in which data points belonging to the same traffic light category are distinctively clustered together. We discuss more about these plots in the ablation study presented in the following section.

\begin{figure*}
\footnotesize 
\centering 
\setlength{\tabcolsep}{0.1cm} 
\begin{tabular}{p{5cm} p{5cm}}
{\includegraphics[width=1\linewidth]{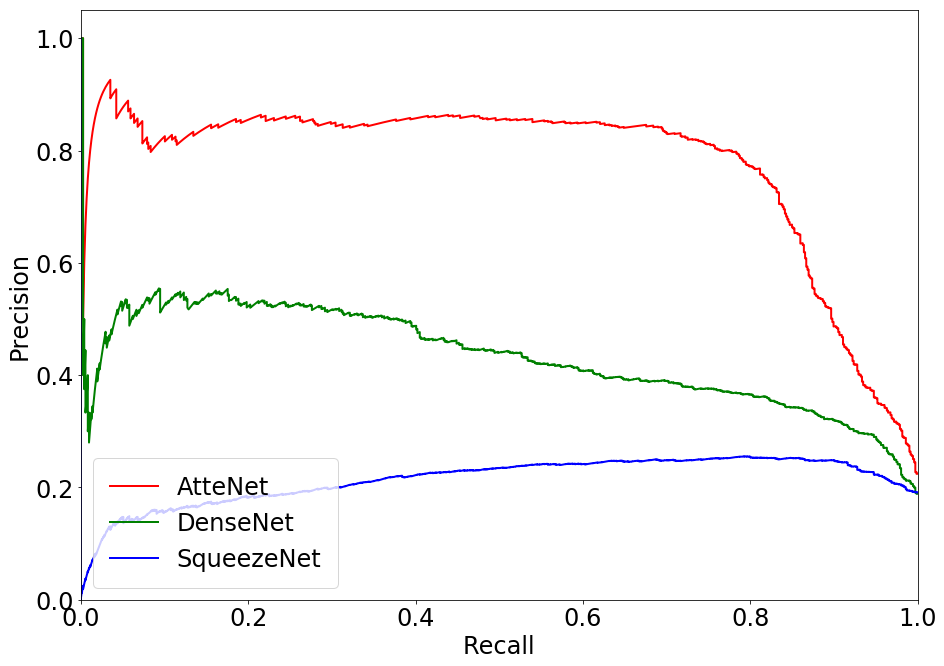}} &
{\includegraphics[width=1\linewidth]{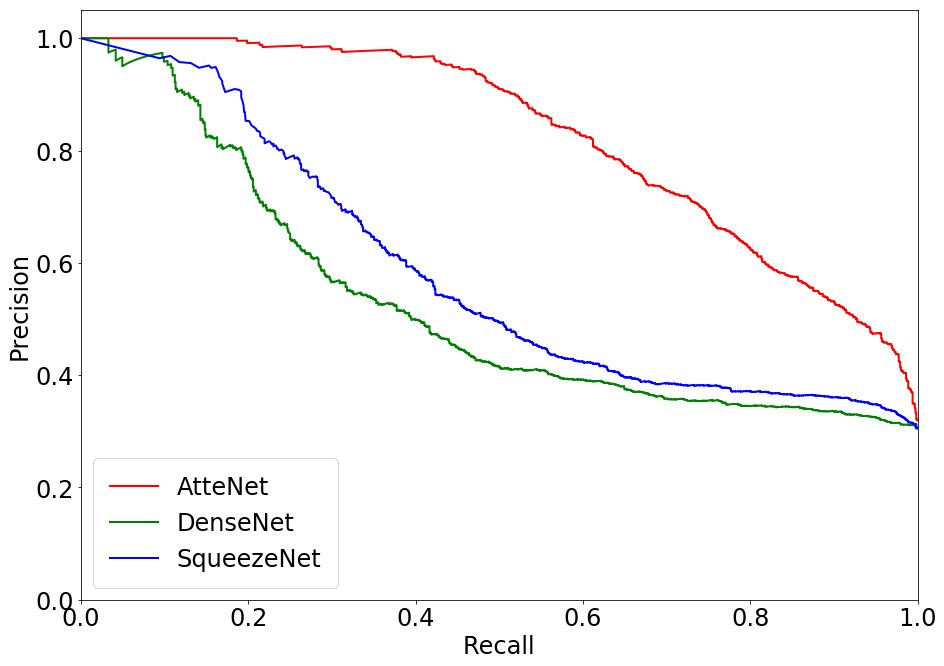}} \\
\multicolumn{1}{c}{(a) Off} & \multicolumn{1}{c}{(b) Red} \\
\\
{\includegraphics[width=1\linewidth]{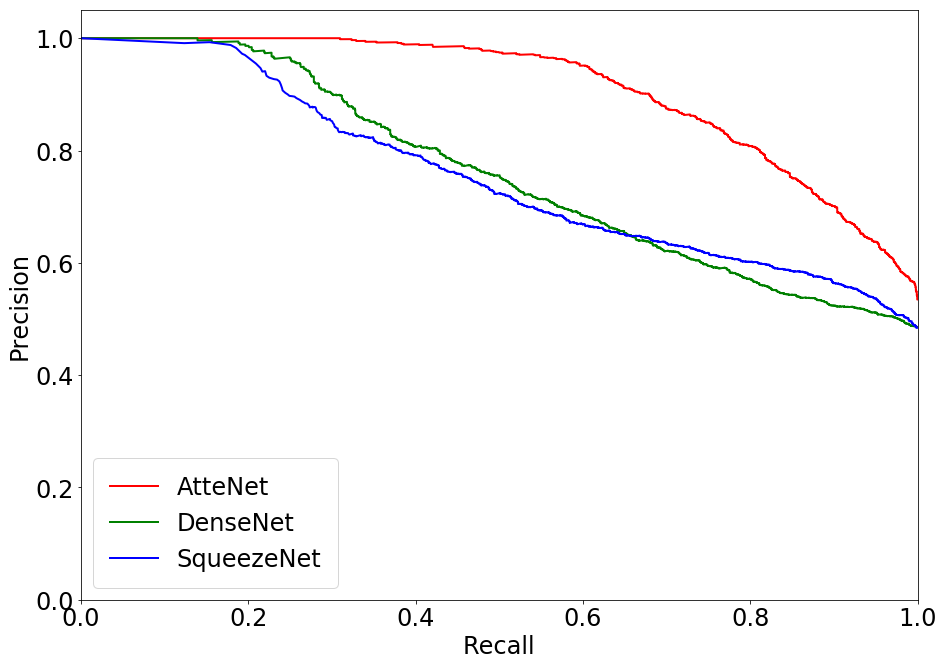}} &
{\includegraphics[width=1\linewidth]{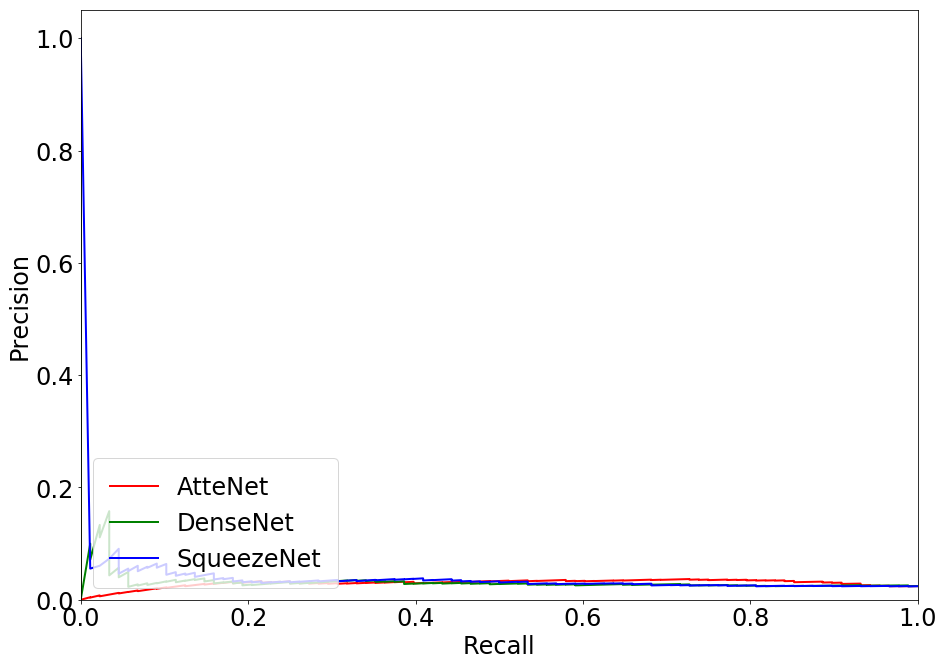}}
\\
 \multicolumn{1}{c}{(c) Green} & \multicolumn{1}{c}{(d) Yellow} \\
\end{tabular} 
\caption{Comparison of the Precision-Recall curves of the various approaches on the Bosch Small Traffic Lights dataset. For each approach, we plot the performance for all traffic light states (Off, Red, Green and Yellow). Our proposed network architecture significantly outperforms the other methods on the three major classes (Off, Red and Green).}
\label{fig:pr_bosch}
\end{figure*}

\begin{figure*}
\footnotesize 
\centering 
\setlength{\tabcolsep}{0.1cm} 
\begin{tabular}{p{5cm} p{5cm} p{5cm}}
{\includegraphics[width=1\linewidth]{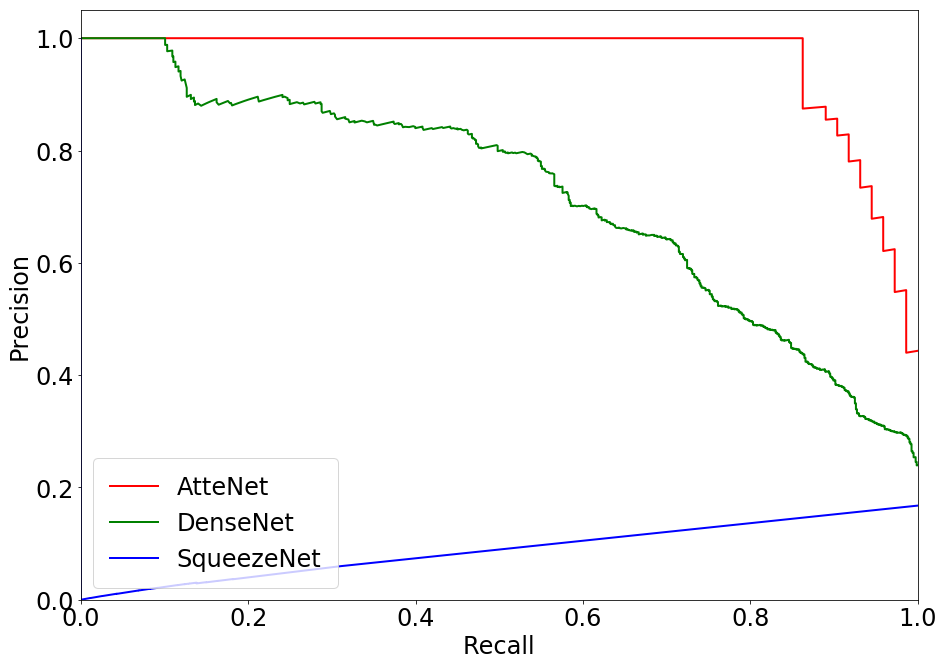}} &
{\includegraphics[width=1\linewidth]{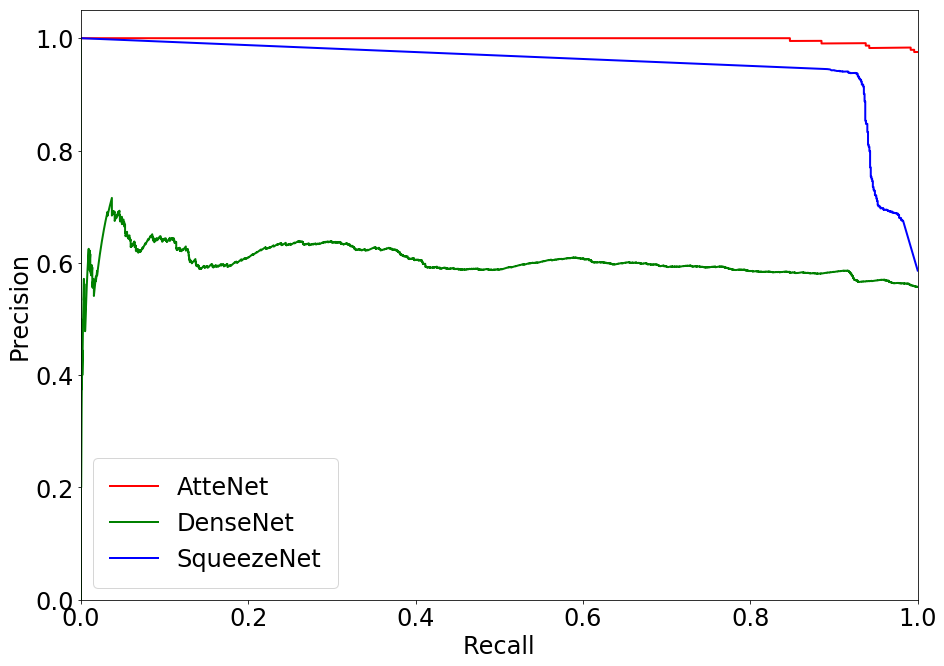}} &
{\includegraphics[width=1\linewidth]{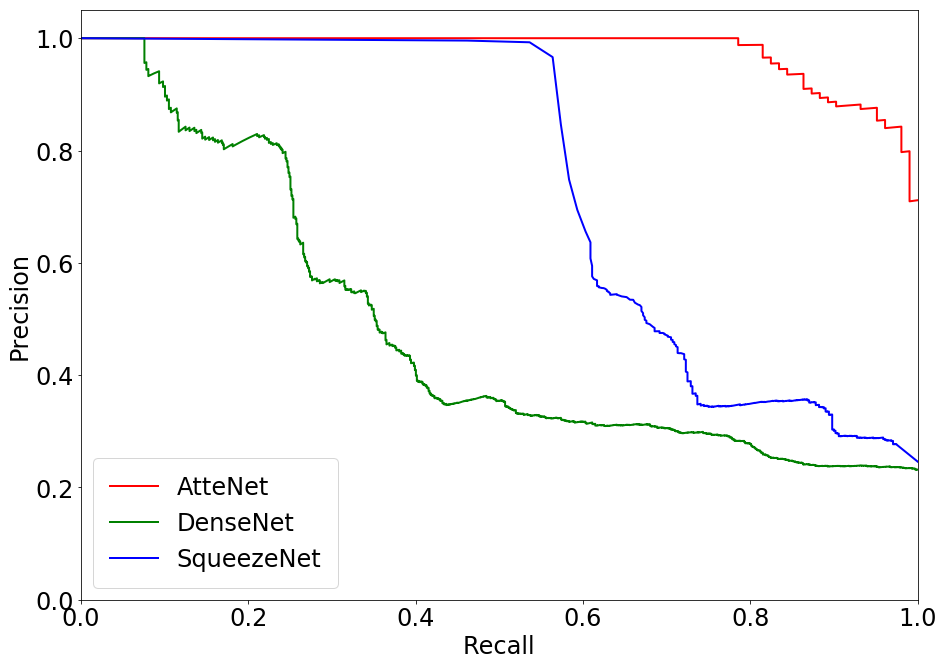}}
\\
\multicolumn{1}{c}{(a) Off} & \multicolumn{1}{c}{(b) Red} & \multicolumn{1}{c}{(c) Green} \\
\end{tabular} 
\caption{Depiction of the Precision-Recall curves on the Freiburg Street Crossing (FSC) dataset for the task of traffic light recognition. We plot the performance of each approach for the three traffic light states: (a) Off, (b) Red and (c) Green. Our proposed AtteNet significantly outperforms the compared variants on all three classes.}
\label{fig:pr_fsc}
\end{figure*}

\tabref{tab:boschpr} shows the precision and recall rates of our proposed AtteNet in comparison to the baseline approaches on the Bosch dataset. Unlike the Nexar dataset, the Bosch traffic lights dataset contains four categories for the traffic light signal by including a label for the yellow state. This in turn increases the difficulty of the task at hand as there only exists few labeled examples for the aforementioned class creating an imbalance in the distribution of the distinct classes. Nonetheless, our proposed approach is able to achieve comparable precision to the baseline variants and the highest recall rate. Furthermore, we plot the precision-recall curves on the Bosch dataset in~\figref{fig:pr_bosch}. The results further corroborate the suitability of AtteNet for traffic light recognition, as it significantly outperforms the compared methods on the three major classes (Off, Red and Green). However, for the "Yellow" traffic light state, all methods perform sub-optimally as seen in~\figref{fig:pr_bosch}(d), which we attribute to the scarcity of the examples containing the "Yellow" traffic light state, thereby increasing the difficulty of learning a good decision function for this class. In~\tabref{tab:fscpr}, we present the precision and recall rates on the FSC dataset. Our proposed AtteNet architecture outperforms the baselines in terms of precision on each of the individual classes, while achieving high recall rates. This further corroborates the suitability of our proposed method for recognizing traffic lights in various conditions as shown in~\figref{fig:tsne}(c) depicting the distribution of the learned features by our model in comparison to the baseline. In~\figref{fig:pr_fsc}, we plot the precision-recall curves for the compared approaches on the FSC dataset. Despite the low performance of SqueezeNet on the "Off" class as shown in~\figref{fig:pr_fsc}(a), it significantly outperforms DenseNet on the remaining two classes. We hypothesize this might be due to the ratio between the traffic light size and the image size which renders it difficult for the network to accurately localize the traffic light thereby leading to spurious recognition. Our proposed AtteNet architecture on the other hand, significantly outperforms the compared approaches for each of the traffic light states further corroborating its suitability for the task at hand.

\begin{table}
\footnotesize 
\centering
\caption{Comparison of classification accuracy of AtteNet with existing CNN traffic light recognition models.}
\label{tab:TLRecog}
\begin{tabular}{p{1.cm}p{1.2cm}p{1.2cm}p{1.2cm}|p{1.2cm}}
\hline\noalign{\smallskip}
Dataset & SqueezeNet & DenseNet & ResNet & AtteNet (Ours) \\
\noalign{\smallskip}\hline\hline\noalign{\smallskip}
Nexar & $94.7\%$ & $91.5\%$ & $88.9\%$ & $\mathbf{95.3\%}$ \\
Bosch & $62.9\%$ & $79.1\%$ & $80.9\%$ & $\mathbf{82.9\%}$ \\
FSC & $76.1\%$ & $79.7\%$ & $86.5\%$ & $\mathbf{91.8\%}$ \\
\noalign{\smallskip}\hline\noalign{\smallskip}
%Average & $N/A$ & $0.44\meter,\, 10.4\degree$ & $0.47\meter,\, 9.81\degree$ & $0.31\meter,\, 9.85\degree$ & $0.25\meter,\, \text{N/A}$ \\
\end{tabular}
\end{table}

\begin{table*}
\footnotesize 
\centering
\caption{Comparison of precision and recall of AtteNet for traffic light recognition on the Nexar dataset.}
\label{tab:nexarpr}
\begin{tabular}{ccccccc}
\hline\noalign{\smallskip}
Model & \multicolumn{3}{c}{Precision} & \multicolumn{3}{c}{Recall} \\
\cmidrule(lr){2-4} \cmidrule(lr){5-7}
 & {No Light} & {Red} & {Green} & {No Light} & {Red} & {Green} \\
\midrule
\midrule
SqueezeNet & $91.8\%$ &  $\mathbf{96.2\%}$ & $94.4\%$ & $90.0\%$ & $96.0\%$ & $95.8\%$ \\
DenseNet & $ 83.2\%$ &  $ 93.2\%$ & $ 91.5\%$ & $ 88.7\%$ & $ 94.7\%$ & $ 87.9\%$ \\
ResNet & $73.4\%$ &  $92.6\%$ & $92.6\%$ & $90.0\%$ & $88.6\%$ & $84.2\%$ \\
AtteNet & $\mathbf{93.8\%}$ &  $95.7\%$ & $\mathbf{95.6\%}$ & $\mathbf{90.3\%}$ & $\mathbf{97.3\%}$ & $\mathbf{95.8\%}$ \\
\hline\noalign{\smallskip}
\end{tabular}
\end{table*}

\begin{table*}
\footnotesize 
\centering
\caption{Comparison of precision and recall of AtteNet for traffic light recognition on the Bosch dataset.}
\label{tab:boschpr}
\begin{tabular}{ccccccccc}
\hline\noalign{\smallskip}
Model & \multicolumn{4}{c}{Precision} & \multicolumn{4}{c}{Recall} \\
\cmidrule(lr){2-5} \cmidrule(lr){6-9}
 & {No Light} & {Red} & {Green} & {Yellow} & {No Light} & {Red} & {Green} & {Yellow} \\
\midrule
\midrule
SqueezeNet & $ 85.0\%$ &  $ 77.4\%$ & $ 84.0\%$ & $ \mathbf{56.0\%}$ & $ 77.6\%$ & $ 80.5\%$ & $87.6\%$ & $15.9\%$ \\
DenseNet & $ 76.1\%$ &  $ 79.6\%$ & $ 80.2\%$ & $ 33.3\%$ & $ \mathbf{79.4\%}$ & $ 69.1\%$ & $88.9\%$ & $4.5\%$ \\
ResNet & $ 80.3\%$ &  $ 76.6\%$ & $ \mathbf{87.5\%}$ & $ 26.6\%$ & $ 77.0\%$ & $ 82.2\%$ & $86.9\%$ & $13.6\%$ \\
AtteNet & $ \mathbf{85.2\%}$ &  $ \mathbf{79.4\%}$ & $ 84.3\%$ & $ 55.2\%$ & $ 77.6\%$ & $ \mathbf{82.6\%}$ & $\mathbf{89.2\%}$ & $\mathbf{16.4}\%$ \\
\hline\noalign{\smallskip}
\end{tabular}
\end{table*}

\begin{table*}
\footnotesize 
\centering
\caption{Comparison of precision and recall of AtteNet for traffic light recognition on the Freiburg Street Crossing (FSC) dataset.}
\label{tab:fscpr}
\begin{tabular}{ccccccc}
\hline\noalign{\smallskip}
Model & \multicolumn{3}{c}{Precision} & \multicolumn{3}{c}{Recall} \\
\cmidrule(lr){2-4} \cmidrule(lr){5-7}
 & {No Light} & {Red} & {Green} & {No Light} & {Red} & {Green} \\
\midrule
\midrule
SqueezeNet & $0.0\%$ &  $98.5\%$ & $97.5\%$ & $0.0\%$ & $96.3\%$ & $\mathbf{97.0\%}$ \\
DenseNet & $ 55.2\%$ &  $ 98.5\%$ & $ 91.1\%$ & $ 99.2\%$ & $ 69.5\%$ & $ 84.3\%$ \\\
ResNet & $71.3\%$ &  $86.1\%$ & $86.3\%$ & $93.2\%$ & $84.6\%$ & $65.3\%$ \\
AtteNet & $\mathbf{75.4\%}$ &  $\mathbf{98.6\%}$ & $\mathbf{99.7\%}$ & $\mathbf{99.3\%}$ & $\mathbf{96.8\%}$ & $71.6\%$ \\
\hline\noalign{\smallskip}
\end{tabular}
\end{table*}

\subsubsection{Ablation Study \& Qualitative Analysis}\mbox{}\\
In this section, we investigate the various architectural decisions made while designing AtteNet as well as present qualitative analysis of the obtained results on the benchmarking datasets. In order to understand the design choices made in AtteNet, we compare the improvements gained employing each of the following variants:
\begin{itemize}
\item ResNet: ResNet-50 base architecture
\item AtteNet-M1: ResNet-50 base architecture with pre-activation residual units
\item AtteNet-M2: ResNet-50 with pre-activation residual units and ELUs
\item AtteNet-M3: ResNet-50 with squeeze-excitation blocks, pre-activation residual units and ELUs
\item AtteNet-M4: ResNet-50 with $1\times1$ convolution squeeze-excitation blocks, pre-activation residual units and ELUs.
\end{itemize}
\tabref{tab:networkChoices} reports the accuracy, precision and recall rates of the aforementioned variants on the Nexar dataset. We observe that the most notable improvement is achieved by replacing the traditional identity residual units with the pre-activation residual units, increasing the accuracy by $3.8\%$. This shows that utilizing the pre-activation residual units enables the network to better regularize the information flow which in turn leads to better representational learning. Replacing the traditional ReLU activation function for ELUs yields an additional $1.6\%$ increase in the recognition accuracy which validates the importance of applying activation functions that are robust to noisy and unbalanced data, as is the case in the current dataset. This can be primarily attributed to the ability of ELUs to help in reducing the bias shift in the neurons which also yields a faster convergence. By incorporating squeeze-excitation blocks and replacing the fully connected layers with $1\times1$ convolutional layers, we are able to further improve on the recognition accuracy of the model. This corroborates the significance of learning different weighting factors for the various channels of the feature maps to enable the network to learn the interdependencies between the channels which in turn improves its recognition capabilities as shown by the improved precision values. 

\begin{table*}
\footnotesize 
\centering
\caption{Comparative Analysis of AtteNet on the Nexar dataset for the task of traffic light recognition.}
\label{tab:networkChoices}
\begin{tabular}{cccccccc}
\hline\noalign{\smallskip}
Model & Accuracy & \multicolumn{3}{c}{Precision} & \multicolumn{3}{c}{Recall} \\
\cmidrule(lr){3-5} \cmidrule(lr){6-8}
 & & {No Light} & {Red} & {Green} & {No Light} & {Red} & {Green} \\
\midrule
\midrule
ResNet & $88.9\%$ & $73.4\%$ &  $92.6\%$ & $92.6\%$ & $90.0\%$ & $88.6\%$ & $84.2\%$ \\
AtteNet-M1 & $92.7\%$ & $86.3\%$ &  $95.8\%$ & $92.7\%$ & $89.3\%$ & $94.3\%$ & $92.7\%$ \\
AtteNet-M2 & $94.3\%$ & $92.0\%$ &  $94.9\%$ & $95.2\%$ & $89.7\%$ & $96.1\%$ & $94.9\%$ \\
AtteNet-M3 & $94.7\%$ & $90.7\%$ &  $\mathbf{97.5\%}$ & $94.2\%$ & $\mathbf{93.0\%}$ & $94.9\%$ & $\mathbf{96.7\%}$ \\
AtteNet-M4 & $\mathbf{95.3\%}$ & $\mathbf{93.8\%}$ &  $95.7\%$ & $\mathbf{95.6\%}$ & $90.3\%$ & $\mathbf{97.3\%}$ & $95.8\%$ \\
\hline\noalign{\smallskip}
\end{tabular}
\end{table*}

\begin{figure*}
\footnotesize 
\centering 
\setlength{\tabcolsep}{0.1cm} 
\begin{tabular}{p{5cm} p{5cm} p{5cm}}
{\includegraphics[width=1\linewidth]{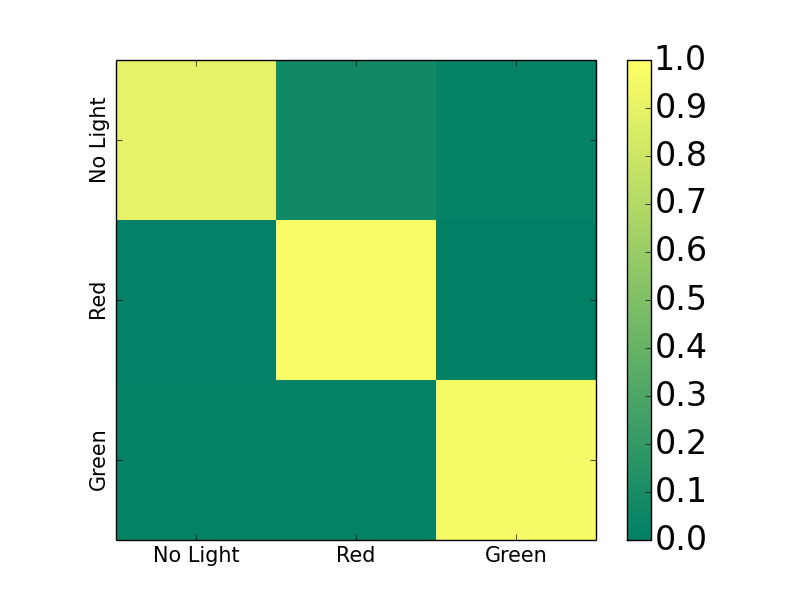}} &
{\includegraphics[width=1\linewidth]{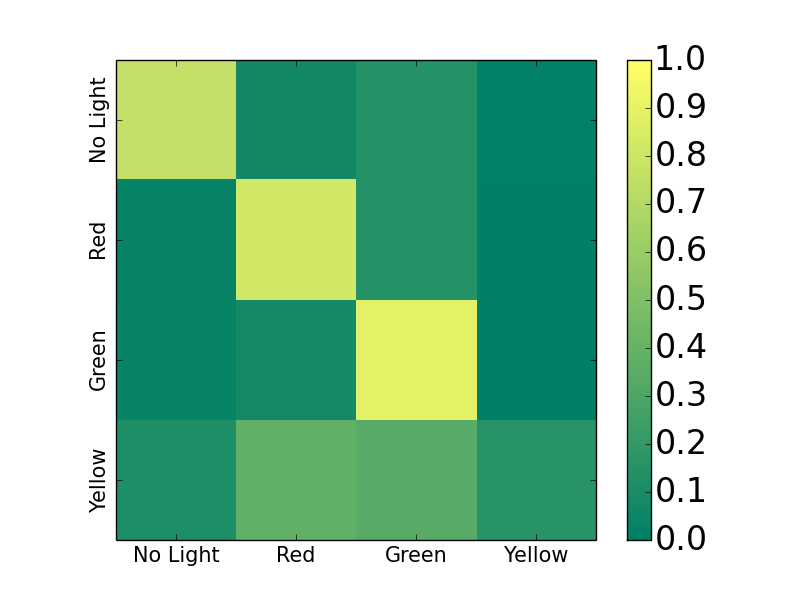}} &
{\includegraphics[width=1\linewidth]{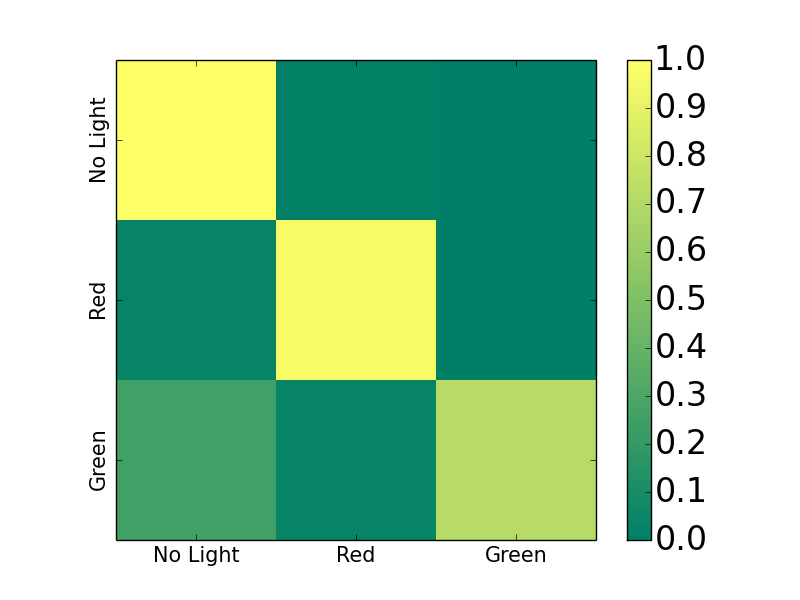}}
\\
\multicolumn{1}{c}{(a) Nexar} & \multicolumn{1}{c}{(b) Bosch} & \multicolumn{1}{c}{(c) FSC} \\
\end{tabular} 
\caption{Confusion matrix of our proposed AtteNet that was trained and evaluated on the different datasets for traffic light recognition. Using our proposed architecture, we are able to accurately disambiguate the distinct traffic light states despite the small size of the light in the image and the various illumination conditions. The groundtruth labels are depicted on the x-axis, while the predictions on the y-axis.}
\label{fig:confMats}
\end{figure*}

Furthermore, we show the confusion matrix of AtteNet trained and evaluated on the different datasets in~\figref{fig:confMats}. On the Nexar dataset, our introduced architecture is able to accurately disambiguate the distinct classes as shown by the diagonal pattern of the confusion matrix. On the bosch dataset shown in~\figref{fig:confMats}(b), AtteNet is able to distinguish with high accuracy between three of the four classes with the yellow traffic light often misclassified as red or green. We believe this occurs as a result of the large imbalance in the distribution of the training data wherein the yellow traffic light occurs $6-10\times$ less compared to the remaining classes. A potential remedy for this problem is to employ class balancing techniques, apply more augmentations to images belonging to this class, or by adding more images of the yellow class to the training set. \figref{fig:confMats}(c) shows the confusion matrix of AtteNet on the FSC dataset. The results indicate that our proposed AtteNet is able to accurately distinguish the various classes further demonstrating the appropriateness of the architecture for the given task.

\begin{figure*}
\footnotesize 
\centering 
\setlength{\tabcolsep}{0.1cm} 
\begin{tabular}{p{0.1cm} p{5cm} p{5cm} p{5cm}}
\rotatebox[origin=c]{90}{AtteNet} & \raisebox{-0.5\height}{\includegraphics[width=1\linewidth]{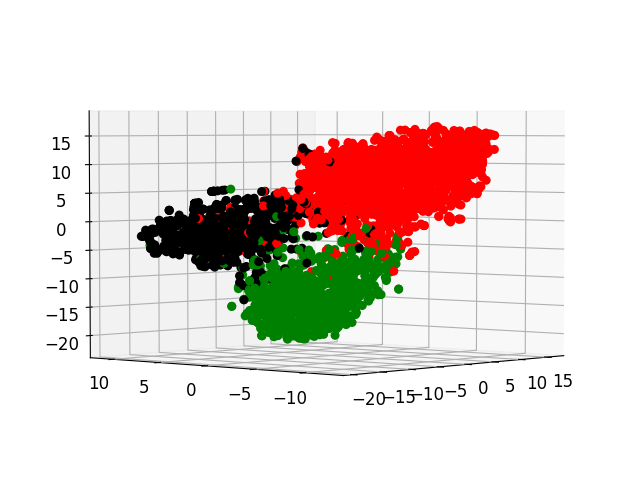}} &
\raisebox{-0.5\height}{\includegraphics[width=1\linewidth]{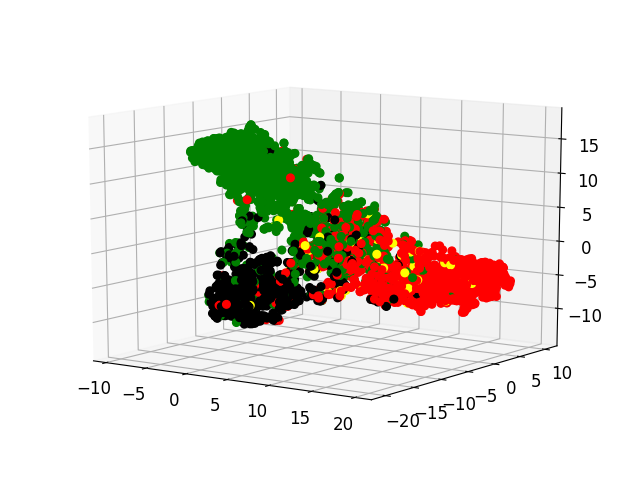}} &
\raisebox{-0.5\height}{\includegraphics[width=1\linewidth]{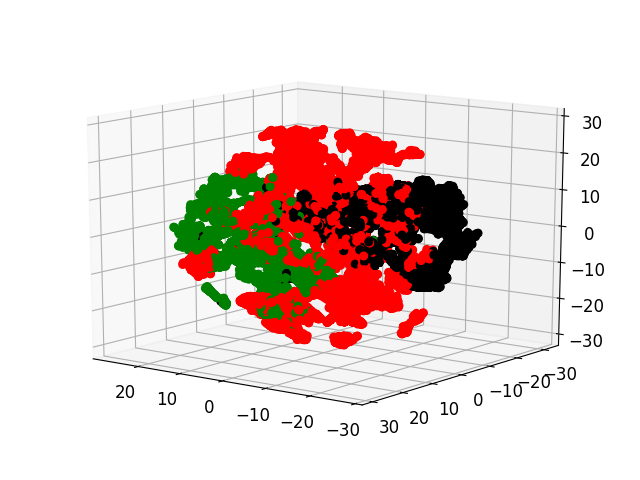}}
\\
\rotatebox[origin=c]{90}{DenseNet} & \raisebox{-0.5\height}{\includegraphics[width=1\linewidth]{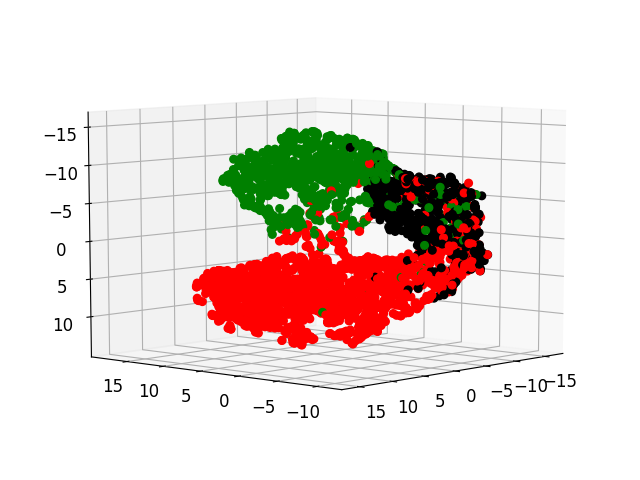}} &
\raisebox{-0.5\height}{\includegraphics[width=1\linewidth]{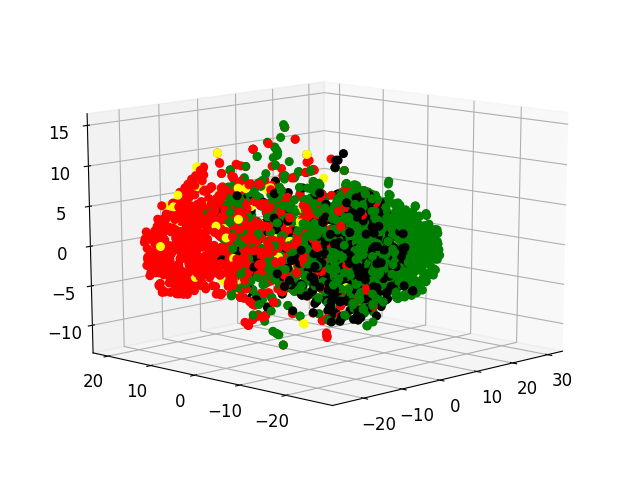}} &
\raisebox{-0.5\height}{\includegraphics[width=1\linewidth]{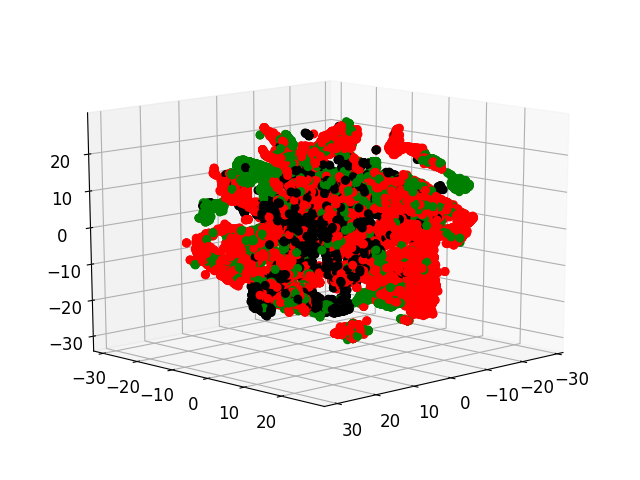}}
\\
& \multicolumn{1}{c}{(a) Nexar} & \multicolumn{1}{c}{(b) Bosch} & \multicolumn{1}{c}{(c) FSC} \\
\end{tabular} 
\caption{3D t-Distributed Stochastic Neighbor Embedding (t-SNE) of features from the penultimate layer of our proposed AtteNet in comparison to DenseNet as a baseline trained on the various datasets for the task of traffic light recognition. The color of the points corresponds to the respective traffic light status, where black denotes off. Features learned by AtteNet can better capture the distribution of the different classes in comparison to DenseNet where the clusters are not well separated.} 
\label{fig:tsne}
\end{figure*}

In order to gain a better understanding of the representations learned by the network, we employ the t-Distributed Stochastic Neighbor embedding (t-SNE)~\citep{maaten2008visualizing} on the learned features of the network. Through obtaining the set of principal components of the data, t-SNE is able to transform the data to a lower dimensional space, thereby revealing cluster and subcluster structures. In~\figref{fig:tsne}, we display the down-projected features obtained after applying t-SNE on the features from the penultimate layer of AtteNet and DenseNet on the various datasets. Unlike DenseNet, the features learned in AtteNet show clear cluster patterns separating the different classes, whereas in DenseNet there is no clear distinction between the features learned for the various classes especially in the Bosch and FSC dataset shown in~\figref{fig:tsne}(b-c). Examining the t-SNE results of AtteNet on the Bosch dataset shows three distinct clusters for the off, red and green classes, with the yellow class falling in between the red and green cluster. Nonetheless, the representations learned by AtteNet are able to better capture the distinct classes in the dataset in comparison to DenseNet where all four classes are merged together in one cluster.
%
% which further validates the low precision and recall results obtained for this class as shown in~\figref{fig:confMats} and~\tabref{tab:boschpr}. Due to the small number of data points for this particular class, the network is unable to learn features that discriminate this class from the remainder classes resulting in the representation shown in the figure.

%\begin{table}
%\footnotesize 
%\centering
%\caption{Comparison of the model size and runtime of AtteNet with baseline approaches on the Freiburg Street Crossing dataset.}
%\label{tab:sizetime}
%\begin{tabular}{p{1.5cm}p{1.3cm}p{1.3cm}p{1.3cm}}
%\hline\noalign{\smallskip}
%Model & Accuracy & Runtime & Model Size\\
%\noalign{\smallskip}\hline\hline\noalign{\smallskip}
%SqueezeNet & $76.1\%$ & $\mathbf{5.79\milli\second}$ & $\mathbf{2.9\mega\byte}$\\
%DenseNet & $79.7\%$ & $46.6\milli\second$ & $27.0\mega\byte$\\
%ResNet & $86.5\%$ & $24.8\milli\second$ & $90\mega\byte$ \\
%AtteNet (Ours) & $\mathbf{91.7\%}$ & $30.3\milli\second$ & $100\mega\byte$ \\
%%Average & $N/A$ & $0.44\meter,\, 10.4\degree$ & $0.47\meter,\, 9.81\degree$ & $0.31\meter,\, 9.85\degree$ & $0.25\meter,\, \text{N/A}$ \\
%%\noalign{\smallskip}\hline\noalign{\smallskip}
%\end{tabular}
%\end{table}

\begin{figure*}
\footnotesize 
\centering 
\setlength{\tabcolsep}{0.2cm} 
\begin{tabular}{p{0.1cm}p{4cm} p{4cm} p{4cm}}
\rotatebox[origin=c]{90}{Input Image} & \raisebox{-0.5\height}{\includegraphics[width=\linewidth]{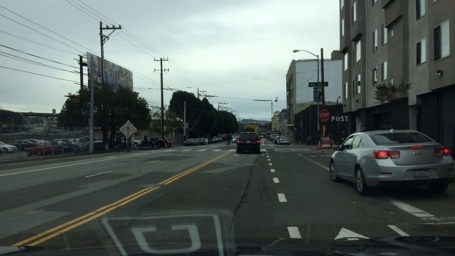}} & \raisebox{-0.5\height}{\includegraphics[width=\linewidth]{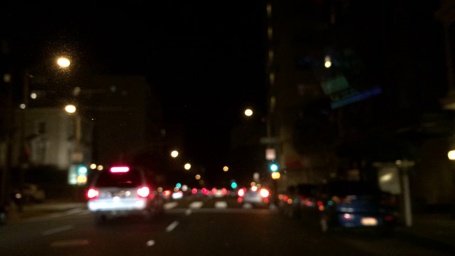}} & \raisebox{-0.5\height}{\includegraphics[width=\linewidth]{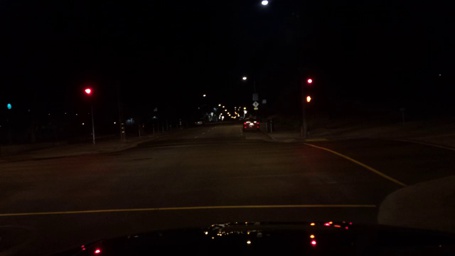}}
\\
& \multicolumn{1}{c}{(a)GT:\includegraphics[scale=1.2]{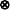}, Pred: \includegraphics[scale=1.2]{figures/nolight.pdf}} & \multicolumn{1}{c}{(b)GT: \includegraphics[scale=1.2]{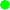}, Pred: \includegraphics[scale=1.2]{figures/green.pdf}} & \multicolumn{1}{c}{(c)GT: \includegraphics[scale=1.2]{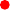}, Pred: \includegraphics[scale=1.2]{figures/red.pdf}}\\
\\
\rotatebox[origin=c]{90}{Input Image} & \raisebox{-0.5\height}{\includegraphics[width=\linewidth]{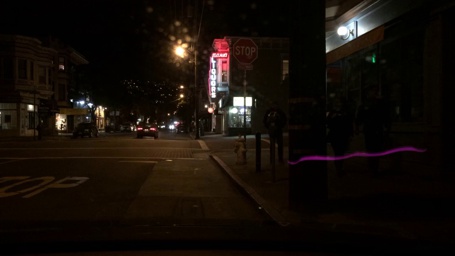}} & \raisebox{-0.5\height}{\includegraphics[width=\linewidth]{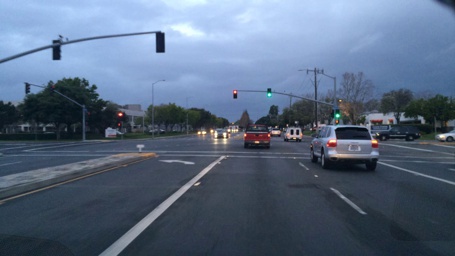}} & \raisebox{-0.5\height}{\includegraphics[width=\linewidth]{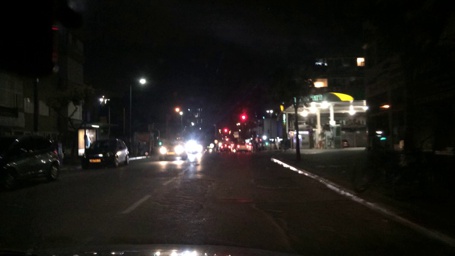}}\\ 
& \multicolumn{1}{c}{(d)GT: \includegraphics[scale=1.2]{figures/nolight.pdf}, Pred: \includegraphics[scale=1.2]{figures/green.pdf}} & \multicolumn{1}{c}{(e)GT: \includegraphics[scale=1.2]{figures/green.pdf}, Pred: \includegraphics[scale=1.2]{figures/red.pdf}} & \multicolumn{1}{c}{(f)GT: \includegraphics[scale=1.2]{figures/red.pdf}, Pred: \includegraphics[scale=1.2]{figures/green.pdf}} 
\end{tabular} 
\caption{Qualitative evaluation of AtteNet predictions on the Nexar dataset for traffic light recognition. The top row shows examples of correctly classified images, and the bottom illustrates those of incorrectly classified images. Below each image, we also illustrate the groundtruth label (GT) and the network prediction (Pred). In images (b) and (c), the network is able to attend to the significant part of the image containing the traffic light and thus producing the correct prediction despite the small size of the light and the illumination noise from other sources of light. For the misclassified images, the attention of the network was placed on an incorrect area of the image (e.g. in image (e)), resulting in an incorrect prediction.} 
\label{fig:nexarQE}
\end{figure*}

Furthermore, we perform qualitative analysis of the recognition accuracy of our proposed AtteNet on the Nexar dataset in~\figref{fig:nexarQE}. \figref{fig:nexarQE}(d-f) show incorrect classifications by our method, where in~\figref{fig:nexarQE}(d), green light reflected off a glass structure is misidentified for a green traffic light signal due to both the shape and position of the light matching the shape and potential placement of a traffic light. Similarly, in~\figref{fig:nexarQE}(f), the green sign of the shop is misidentified as the traffic light resulting in an incorrect classification. In~\figref{fig:nexarQE}(e), the lack of information identifying the driving direction of the car results in a misclassification as the network incorrectly identifies the left-most traffic light to be the relevant one. However, despite the significant motion blur and low lighting conditions, our proposed model is able to accurately predict the state of the traffic light as shown in~\figref{fig:nexarQE}(a-c). 

\begin{figure*}
\footnotesize 
\centering 
\setlength{\tabcolsep}{0.2cm} 
\begin{tabular}{p{0.1cm}p{4cm} p{4cm} p{4cm} p{4cm}}
\rotatebox[origin=c]{90}{Input Image} & \raisebox{-0.5\height}{\includegraphics[width=\linewidth]{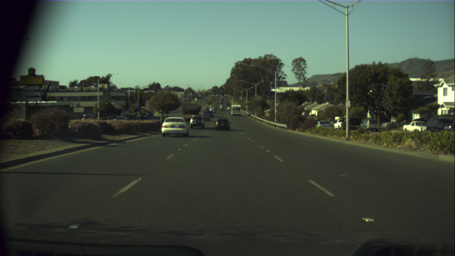}} & \raisebox{-0.5\height}{\includegraphics[width=\linewidth]{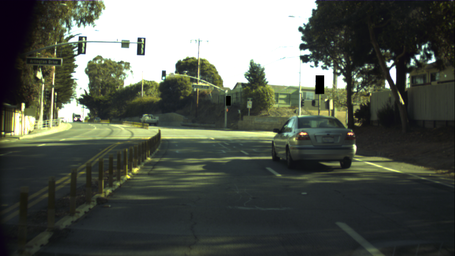}} & \raisebox{-0.5\height}{\includegraphics[width=\linewidth]{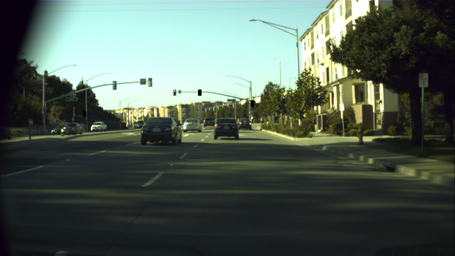}} & \raisebox{-0.5\height}{\includegraphics[width=\linewidth]{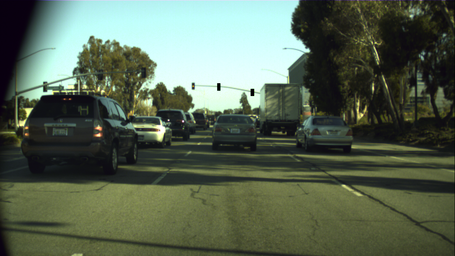}} 
\\
& \multicolumn{1}{c}{(a)GT: \includegraphics[scale=1.2]{figures/nolight.pdf}, Pred: \includegraphics[scale=1.2]{figures/nolight.pdf}} & \multicolumn{1}{c}{(b)GT: \includegraphics[scale=1.2]{figures/green.pdf}, Pred: \includegraphics[scale=1.2]{figures/green.pdf}} & \multicolumn{1}{c}{(c)GT: \includegraphics[scale=1.2]{figures/red.pdf}, Pred: \includegraphics[scale=1.2]{figures/red.pdf}} & \multicolumn{1}{c}{(d)GT: \includegraphics[scale=1.2]{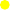}, Pred: \includegraphics[scale=1.2]{figures/yellow.pdf}}\\
\\
\rotatebox[origin=c]{90}{Input Image} & \raisebox{-0.5\height}{\includegraphics[width=\linewidth]{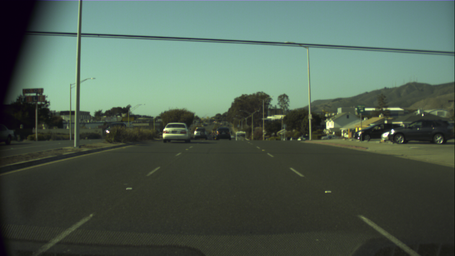}} & \raisebox{-0.5\height}{\includegraphics[width=\linewidth]{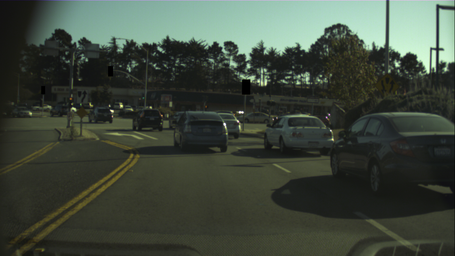}} & \raisebox{-0.5\height}{\includegraphics[width=\linewidth]{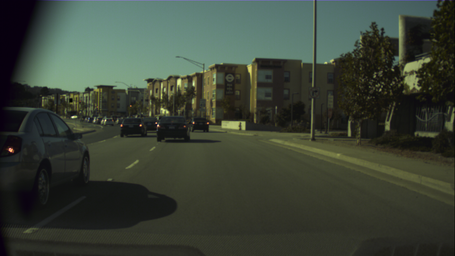}} & \raisebox{-0.5\height}{\includegraphics[width=\linewidth]{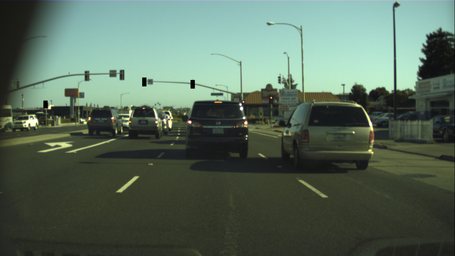}}\\ 
& \multicolumn{1}{c}{(e)GT: \includegraphics[scale=1.2]{figures/nolight.pdf}, Pred: \includegraphics[scale=1.2]{figures/red.pdf}} & \multicolumn{1}{c}{(f)GT: \includegraphics[scale=1.2]{figures/green.pdf}, Pred: \includegraphics[scale=1.2]{figures/nolight.pdf}} & \multicolumn{1}{c}{(g)GT: \includegraphics[scale=1.2]{figures/red.pdf}, Pred: \includegraphics[scale=1.2]{figures/nolight.pdf}} & \multicolumn{1}{c}{(h)GT: \includegraphics[scale=1.2]{figures/yellow.pdf}, Pred: \includegraphics[scale=1.2]{figures/red.pdf}}
\end{tabular} 
\caption{Qualitative evaluation of AtteNet on the Bosch Traffic Lights dataset. Images (a-d) illustrate correctly predicted examples, while images (e-h) illustrate misclassified predictions. Despite the small size of the traffic lights (fig. (b)), and the presence of multiple traffic lights in one image (fig. (c, d)), our proposed approach is able to accurately predict the state of the traffic light.} 
\label{fig:boschQE}
\end{figure*}

\begin{figure*}
\footnotesize 
\centering 
\setlength{\tabcolsep}{0.2cm} 
\begin{tabular}{p{0.1cm}p{4cm} p{4cm} p{4cm}}
\rotatebox[origin=c]{90}{Input Image} & \raisebox{-0.5\height}{\includegraphics[width=\linewidth]{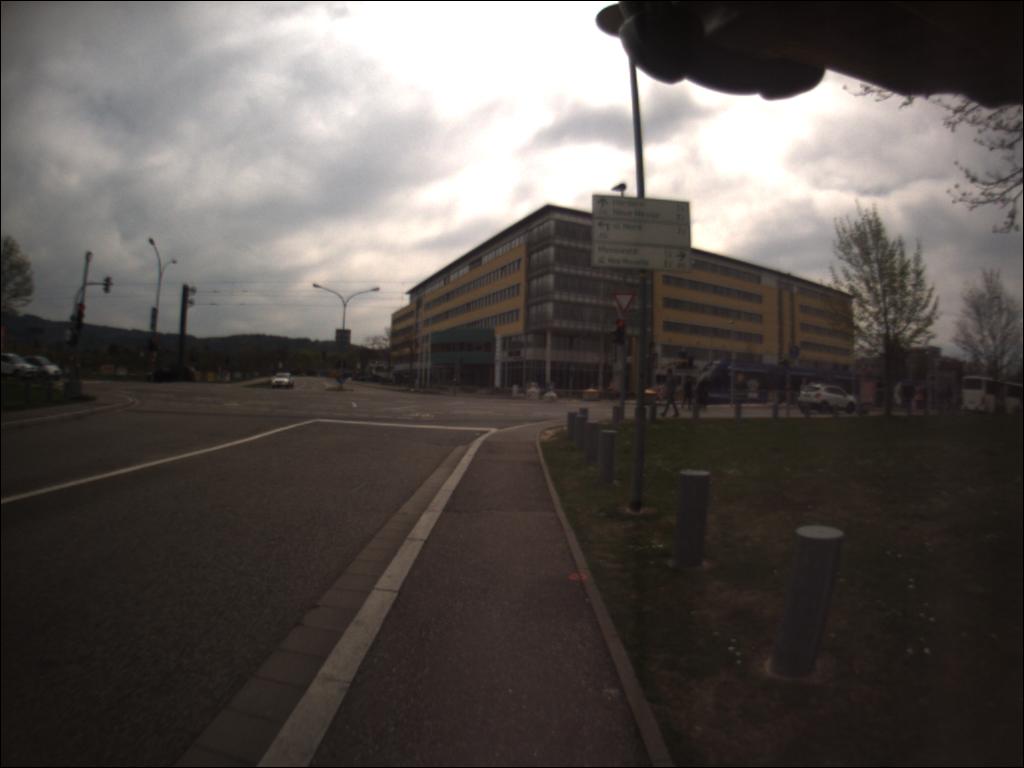}} & \raisebox{-0.5\height}{\includegraphics[width=\linewidth]{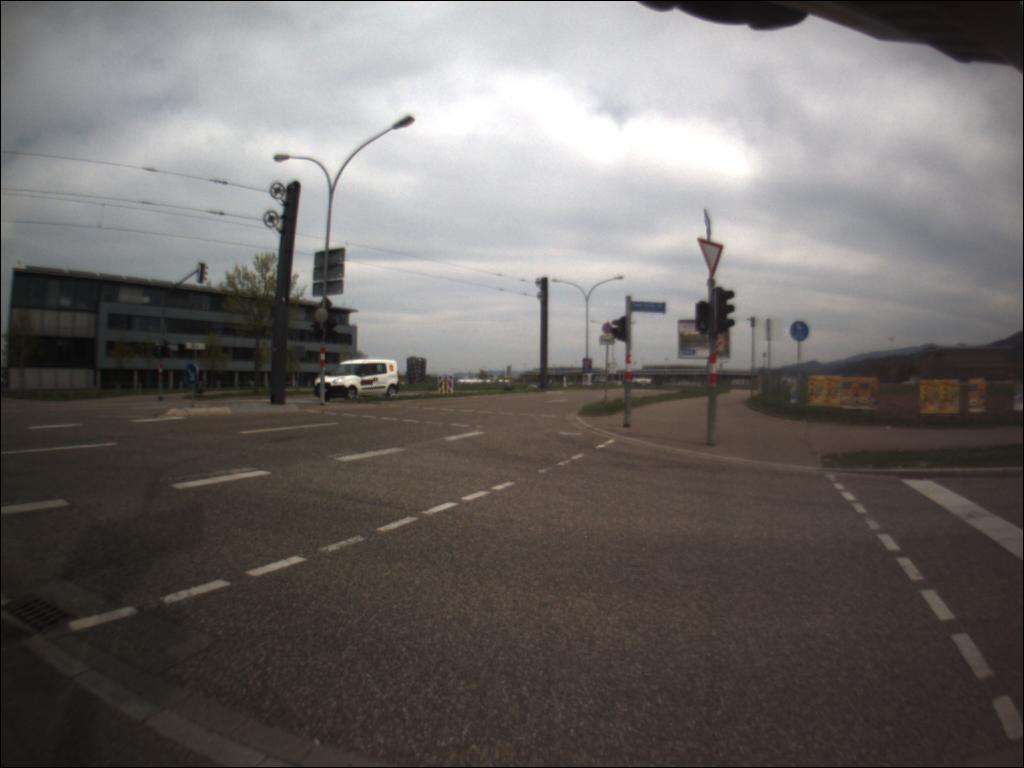}} & \raisebox{-0.5\height}{\includegraphics[width=\linewidth]{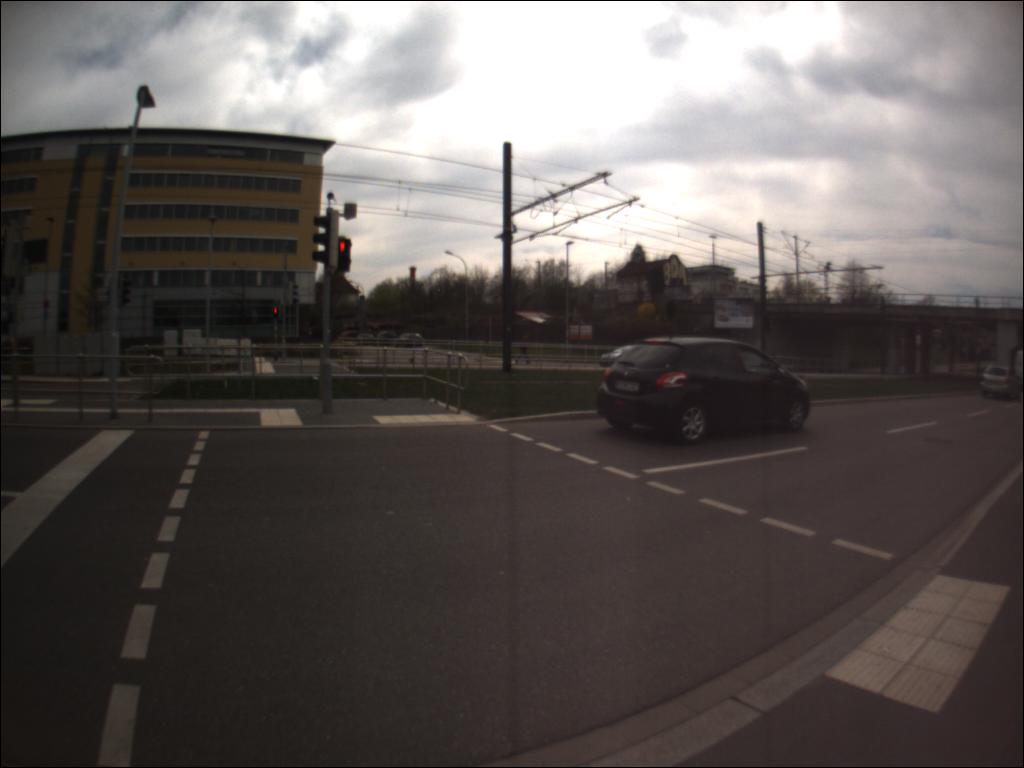}}
\\
& \multicolumn{1}{c}{(a)GT: \includegraphics[scale=1.2]{figures/nolight.pdf}, Pred: \includegraphics[scale=1.2]{figures/nolight.pdf}} & \multicolumn{1}{c}{(b)GT: \includegraphics[scale=1.2]{figures/green.pdf}, Pred: \includegraphics[scale=1.2]{figures/green.pdf}} & \multicolumn{1}{c}{(c)GT: \includegraphics[scale=1.2]{figures/red.pdf}, Pred: \includegraphics[scale=1.2]{figures/red.pdf}}\\
\\
\rotatebox[origin=c]{90}{Input Image} & \raisebox{-0.5\height}{\includegraphics[width=\linewidth]{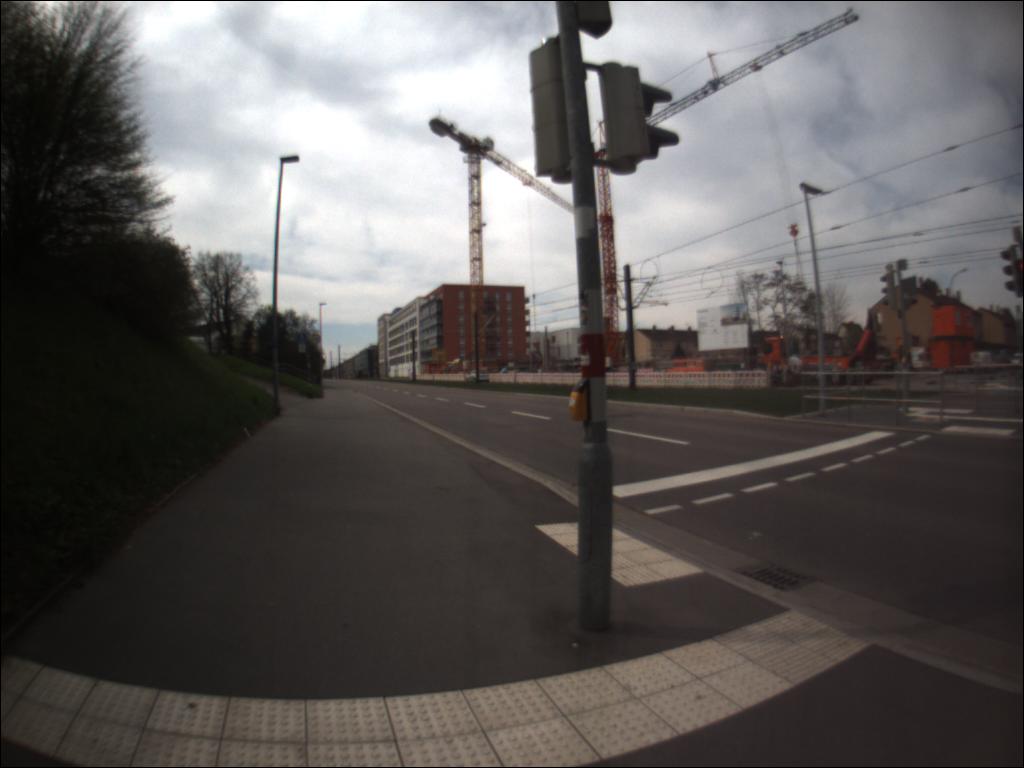}} & \raisebox{-0.5\height}{\includegraphics[width=\linewidth]{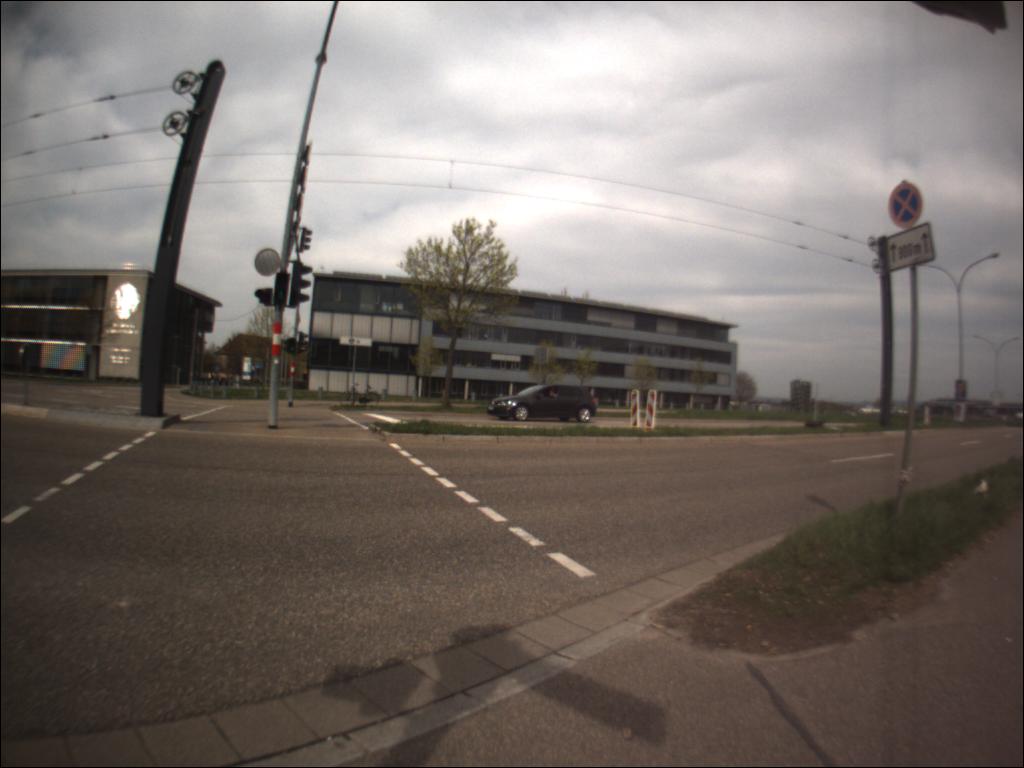}} & \raisebox{-0.5\height}{\includegraphics[width=\linewidth]{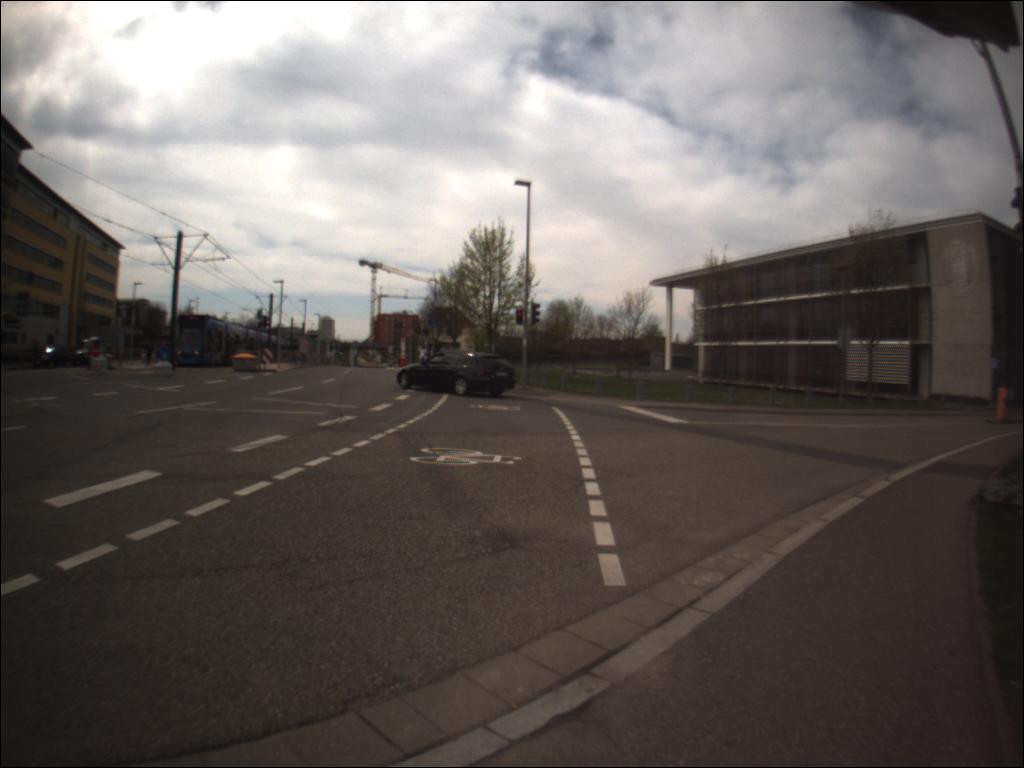}}\\ 
& \multicolumn{1}{c}{(d)GT: \includegraphics[scale=1.2]{figures/nolight.pdf}, Pred: \includegraphics[scale=1.2]{figures/red.pdf}} & \multicolumn{1}{c}{(e)GT: \includegraphics[scale=1.2]{figures/green.pdf}, Pred: \includegraphics[scale=1.2]{figures/nolight.pdf}} & \multicolumn{1}{c}{(f)GT: \includegraphics[scale=1.2]{figures/red.pdf}, Pred: \includegraphics[scale=1.2]{figures/nolight.pdf}} 
\end{tabular} 
\caption{Qualitative analysis of the proposed traffic light recognition method on the Freiburg Street Crossing dataset. Figures (a-c) illustrate correctly predicted images, while figures (d-f) illustrate mispredicted images. The small size of the traffic lights (fig. (b, f)), presence of noise sources (fig. (c, d)) and the varying illumination conditions (fig. (a, e)) under which the dataset was captured adds to the difficulty of the dataset for the given task. Despite these challenges, our approach is able to accurately recognize the state of the traffic light (fig. (a, b, c)).} 
\label{fig:fscQE}
\end{figure*}

In~\figref{fig:boschQE}, we show qualitative results on the Bosch traffic light dataset. \figref{fig:boschQE}(f,g) show misclassification examples where AtteNet incorrectly predicts no traffic light in the driving direction. In both cases this occurs due to the small size of the traffic light and the presence of partial occlusions such as a pole hiding part of the traffic light. In~\figref{fig:boschQE}(h) a yellow traffic light signal is incorrectly classified as red. We attribute the cause of the misprediction to the close similarity of the red and yellow colors particularly in this image which can be verified by comparing the color of the brake lights of the cars to that of the traffic light signal. \figref{fig:boschQE}(b) shows a correct classification of a green traffic light signal, where our proposed AtteNet is able to accurately recognize the traffic light signal despite the small size of the traffic light and the presence of partial occlusions. Similarly, in~\mbox{\figref{fig:boschQE}(c, d)}, our network is able to accurately recognize the red traffic light despite the presence of several surrounding traffic lights, the small size of the light and the presence of blur.

\begin{figure*}
\footnotesize 
\centering 
\setlength{\tabcolsep}{0.2cm} 
\begin{tabular}{p{3.5cm} p{3.5cm} p{3.5cm} p{3.5cm}}
\multicolumn{1}{c}{Input Image} & \multicolumn{1}{c}{Activation Mask} & \multicolumn{1}{c}{Input Image} & \multicolumn{1}{c}{Activation Mask} \\
\includegraphics[width=3.5cm, height=3.5cm]{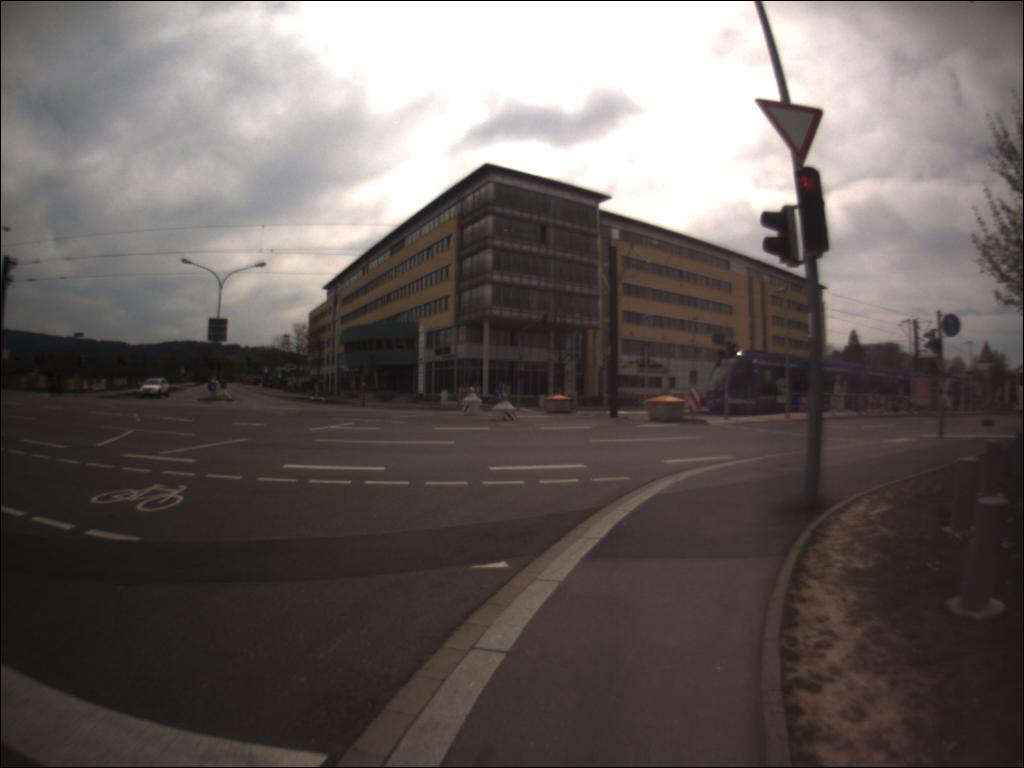} & \includegraphics[width=3.5cm, height=3.5cm]{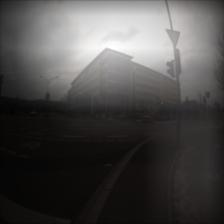} & \includegraphics[width=3.5cm, height=3.5cm]{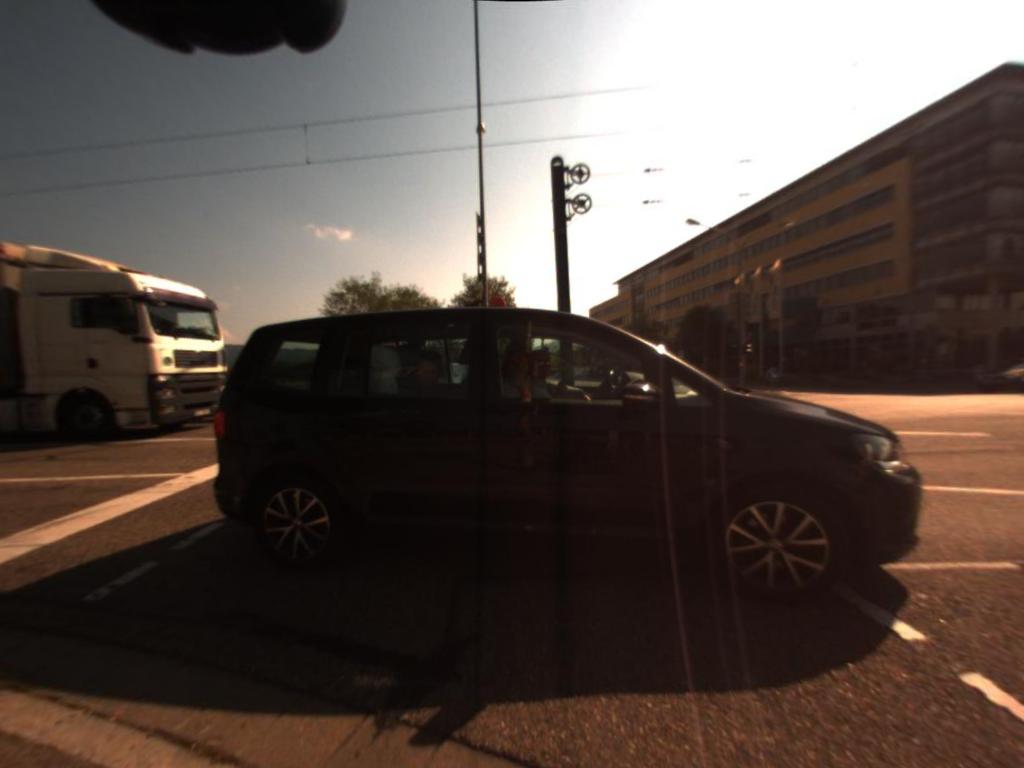} & \includegraphics[width=3.5cm, height=3.5cm]{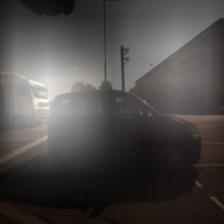}\\
\multicolumn{2}{c}{(a) GT: \includegraphics[scale=1.4]{figures/nolight.pdf}, Pred: \includegraphics[scale=1.4]{figures/nolight.pdf}} & \multicolumn{2}{c}{(b) GT: \includegraphics[scale=1.4]{figures/nolight.pdf}, Pred: \includegraphics[scale=1.4]{figures/red.pdf}}\\
\includegraphics[width=3.5cm, height=3.5cm]{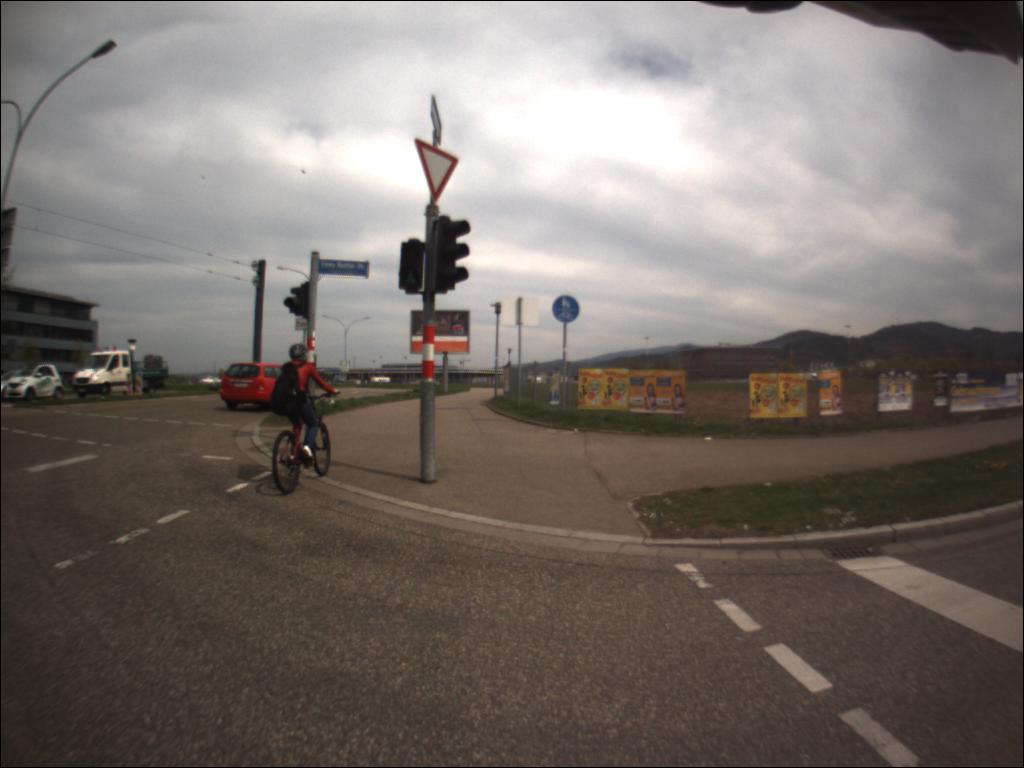} & \includegraphics[width=3.5cm, height=3.5cm]{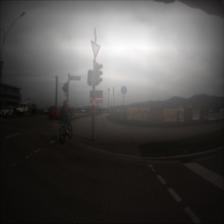} & \includegraphics[width=3.5cm, height=3.5cm]{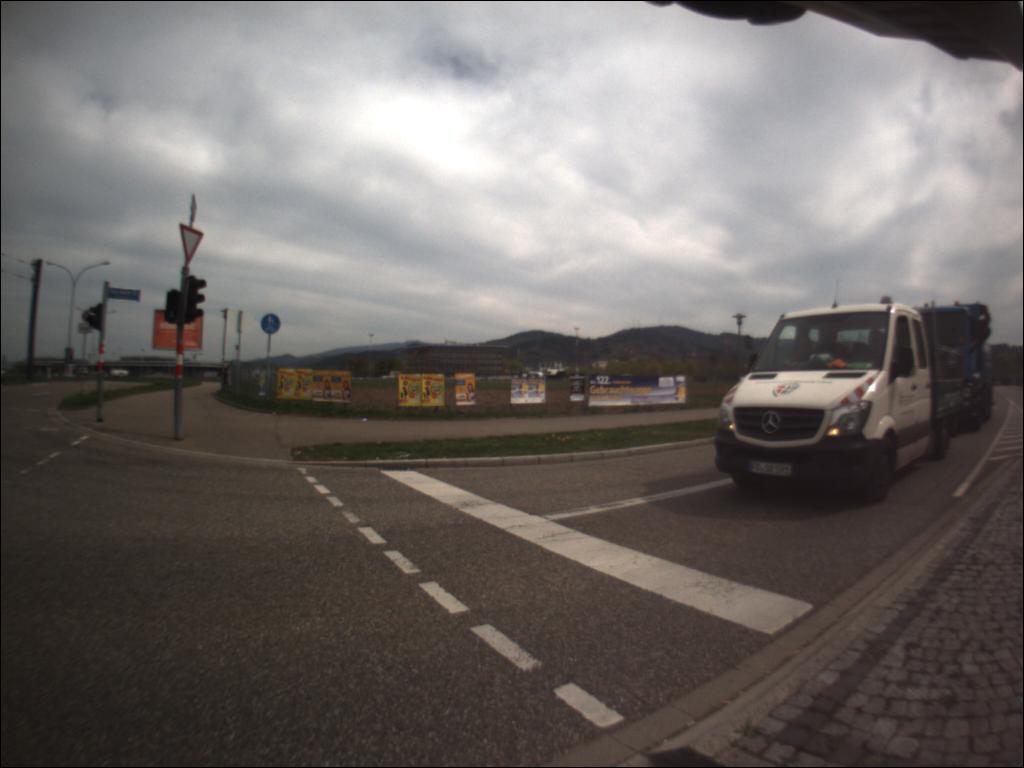} & \includegraphics[width=3.5cm, height=3.5cm]{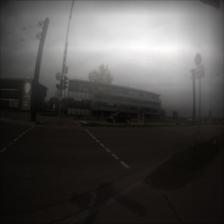} \\
\multicolumn{2}{c}{(c) GT: \includegraphics[scale=1.4]{figures/green.pdf}, Pred: \includegraphics[scale=1.4]{figures/green.pdf}} & \multicolumn{2}{c}{(d) GT: \includegraphics[scale=1.4]{figures/green.pdf}, Pred: \includegraphics[scale=1.4]{figures/nolight.pdf}}\\
\includegraphics[width=3.5cm, height=3.5cm]{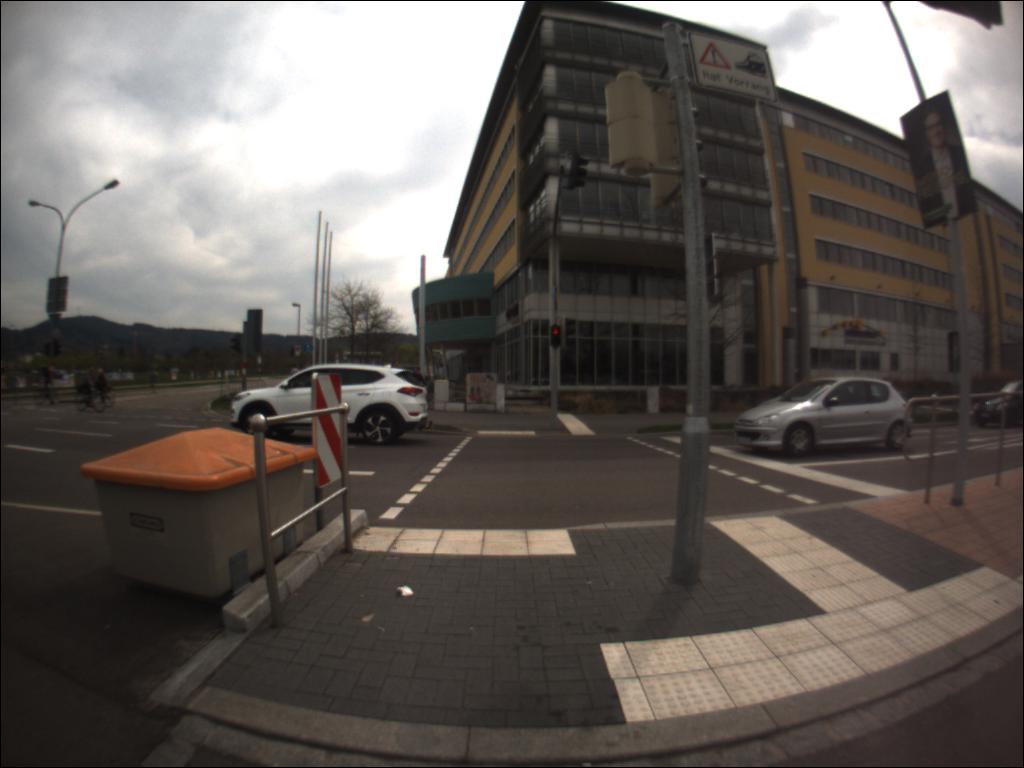} & \includegraphics[width=3.5cm, height=3.5cm]{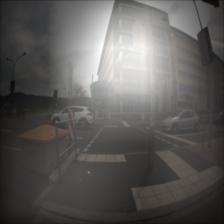} & \includegraphics[width=3.5cm, height=3.5cm]{figures/obelix_red_nolight.jpg} & \includegraphics[width=3.5cm, height=3.5cm]{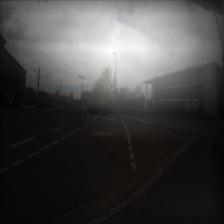}\\
\multicolumn{2}{c}{(e) GT: \includegraphics[scale=1.4]{figures/red.pdf}, Pred: \includegraphics[scale=1.4]{figures/red.pdf}} & \multicolumn{2}{c}{(f) GT: \includegraphics[scale=1.4]{figures/red.pdf}, Pred: \includegraphics[scale=1.4]{figures/nolight.pdf}}
\end{tabular} 
\caption{Visualization of the classification activation maps using Grad-CAM~\cite{selvaraju2016grad} of our AtteNet on the Freiburg Street Crossing (FSC) dataset. We overlay the activation masks on the image for ease of visualization. Below each image, we depict the ground-truth label (GT) and the network prediction (Pred).}
\label{fig:fscCAM}
\end{figure*}

We present a qualitative evaluation of the performance of AtteNet on the FSC dataset in~\figref{fig:fscQE}. In~\figref{fig:fscQE}(d), the red and white pattern on the traffic light pole cause our network to incorrectly predict the presence of a red traffic light signal despite the absence of a traffic light within the image. \figref{fig:fscQE}(a-c) show challenging scenarios in which AtteNet is able to accurately recognize the state of the traffic light despite the lighting conditions, presence of multiple light sources and the small size of the traffic light. Despite the achieved accuracy of the network, failing to recognize a traffic light as shown in~\mbox{\figref{fig:fscQE}(e, f)} will lead to unintended circumstances. While on the one hand, recognizing the traffic light in both images is quite challenging even for humans due to its small size in the image, one cannot rely solely on the traffic light recognition system to decide the safety of the intersection. Our proposed approach for autonomous street crossing prediction rather combines the information from both the traffic light recognition and motion prediction modules to accurately predict the intersection safety for crossing as discussed in the following section.

In order to provide further insight into the AtteNet architecture, in~\figref{fig:fscCAM} we utilize the Grad-CAM method by~\cite{selvaraju2016grad} to visualize the activation masks of AtteNet on the FSC dataset. Visualizing the output of the penultimate layer of our network using Grad-CAM produces a gradient-weighted class activation mask highlighting regions of the relevant regions in the image for predicting the output, thus providing us with a better understanding of the network predictions. For each image we show the activation mask, the ground-truth label and the network prediction. In~\figref{fig:fscCAM}(a, c, e), the attention of the network is placed on areas of the image that contain the traffic light hence leading to correct predictions. \figref{fig:fscCAM}(b) shows an example image where a car crossing the intersection is occluding the traffic light. The activation mask shows that the attention of the network is incorrectly placed on the brake lights of the car which in turn lead to the incorrect prediction of a red traffic light signal. In~\figref{fig:fscCAM}(d, f), the small size of the traffic light in the image increase the difficulty of locating and recognizing it as can be seen from the activation masks.

\subsection{Evaluation of the Crossing Decision}
\label{subsec:crossEv}
We evaluate the performance of our proposed Autonomous Road Crossing Predictor \textit{(ARCP)} by reporting the accuracy, precision and recall rates for the ``Safe" class prediction on the FSC dataset. We compare the performance of our approach ARCP(TLR+MP) with the following methods:

\begin{itemize}
    \item Random Forest classifier of~\cite{radwan17iros}.
    \item CV+TLR: Constant velocity motion prediction model coupled with our proposed traffic light recognition module.
    \item ARCP(MP): Variant of our proposed predictor that relies only on the motion prediction module.
    \item ARCP(TLR): Variant of our proposed predictor that relies only on the traffic light recognition module.
    \item NCP(TLR+MP): Naive Crossing Predictor (NCP) which is a binary classifier trained on a concatenation of the predictions from the TLR and MP modules. 
\end{itemize}

Comparing the performance of the full approach with both ARCP(MP) and ARCP(TLR) variants helps in evaluating the tolerance of the learned predictor to mispredictions and noise from the information source. Additionally, comparing with the NCP(TLR+MP) helps in understanding whether utilizing the proposed fusion strategy and incorporating the uncertainties in the predictions has an impact on the overall crossing strategy. We use the original 10 sequences from the FSC dataset for training, and test on the newly captured 8 sequences. \tabref{tab:prCross} demonstrates the precision, recall and accuracy for each of the aforementioned methods. We compute the aforementioned metrics with respect to the ``Safe" class. More precisely, a prediction is considered a true positive if both the groundtruth and the detection label are for the class ``Safe". We observe a drop in the accuracy of the Random Forest classifier in comparison to the reported results in~\cite{radwan17iros}, which we attribute to the inability of the classifier to generalize to unseen behavior as in the newly captured sequences. This occurs as the Random Forest classifier learns a discriminative model of the problem which leads to suboptimal behavior in new scenarios.

\begin{table}
\footnotesize 
\centering
\caption{Comparative Analysis of the learned crossing decision on the FSC dataset. The following metrics are computed for the ``Safe" class prediction.}
\label{tab:prCross}
\begin{tabular}{p{1.9cm}p{1.2cm}p{1.2cm}p{1.2cm}}
\hline\noalign{\smallskip}
Method & Precision & Recall & Accuracy\\
\noalign{\smallskip}\hline\hline\noalign{\smallskip}
Random Forest & $78.8\%$ & $60.7\%$ & $75.4\%$ \\
ARCP(TLR) & $53.1\%$ & $61.3\%$ & $54.6\%$ \\
ARCP(MP) & $87.8\%$ & $\mathbf{93.5\%}$ & $85.4\%$ \\
CV+TLR & $64.1\%$ & $73.7\%$ & $69.6\%$ \\
NCP(TLR+MP) & $69.6\%$ & $58.1\%$ & $63.81\%$\\
\noalign{\smallskip}\hline\noalign{\smallskip}
ARCP(TLR+MP) & $\mathbf{91.9\%}$ & $82.3\%$ & $\mathbf{86.2\%}$ \\
\noalign{\smallskip}\hline\noalign{\smallskip}
\end{tabular}
\end{table}

\begin{figure*}
\footnotesize 
\centering 
\setlength{\tabcolsep}{0.1cm} 
\begin{tabular}{p{4cm} p{4cm} p{4cm}}
{\includegraphics[width=1\linewidth]{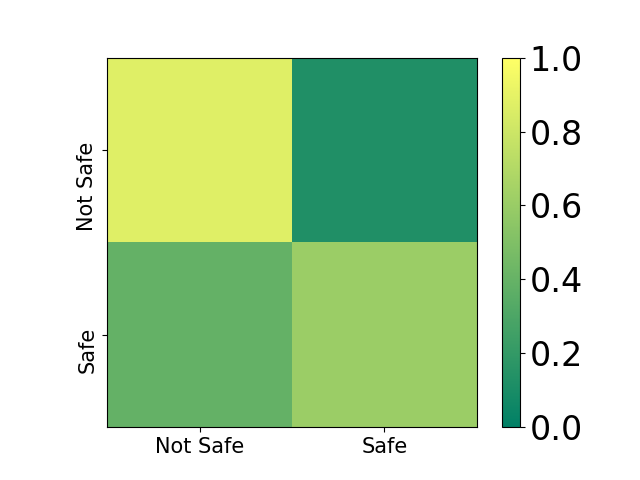}} &
{\includegraphics[width=1\linewidth]{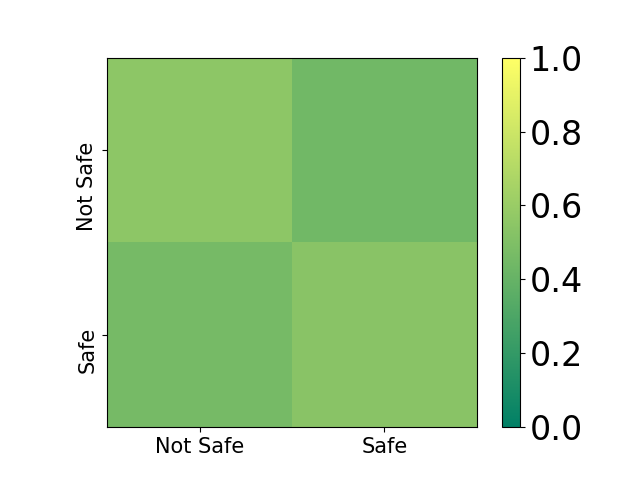}} &
{\includegraphics[width=1\linewidth]{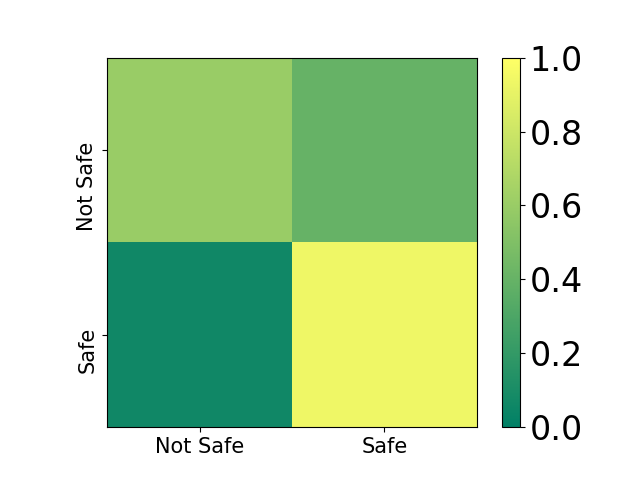}} \\
\multicolumn{1}{c}{(a) Random Forest} & \multicolumn{1}{c}{(b) ARCP(TLR)} & 
\multicolumn{1}{c}{(c) ARCP(MP)} \\
{\includegraphics[width=1\linewidth]{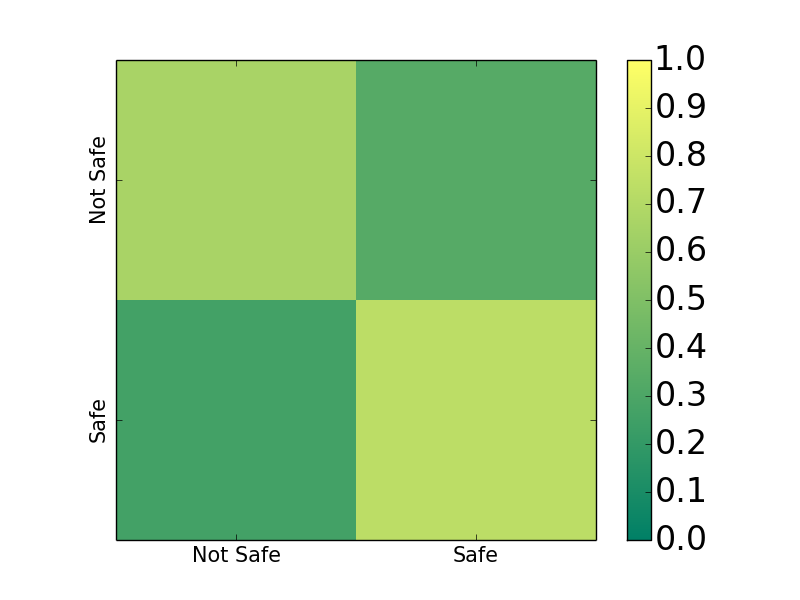}} &
{\includegraphics[width=1\linewidth]{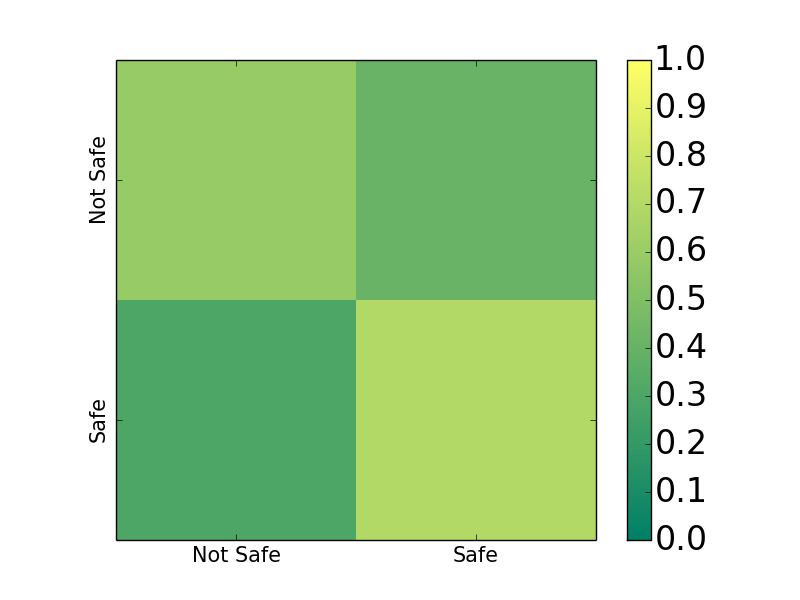}} &
{\includegraphics[width=1\linewidth]{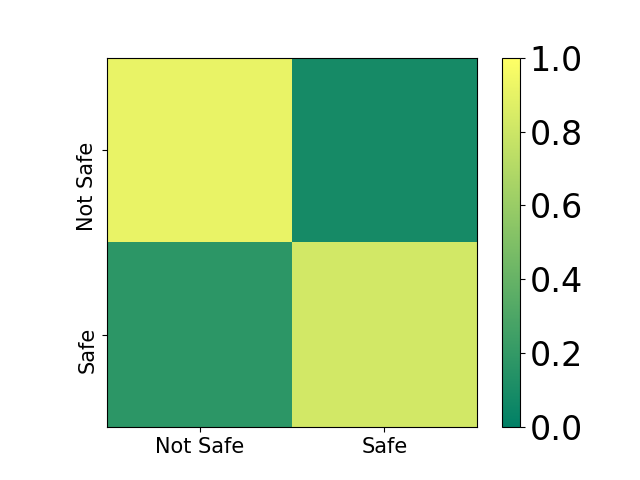}} \\ 
\multicolumn{1}{c}{(d) CV + TLR} & \multicolumn{1}{c}{(e) NCP(TLR+MP)}  & \multicolumn{1}{c}{(f) ARCP(TLR+MP)} \\
\end{tabular} 
\caption{Confusion matrix of the various Autonomous Crossing Predictors (ARCP) on the FSC dataset in comparison to the baseline Random Forest Classifier , a constant velocity motion prediction module coupled with our proposed traffic light prediction module (CV + TLR) and a Naive Crossing Predictor NCP(TLR+MP) fusing both the predictions from the traffic light and the motion prediction modules. The best performance is achieved by utilizing both information from the Traffic Light Recognition (TLR) module and the Motion Prediction (MP) module to learn the crossing safety. The groundtruth labels are depicted on the x-axis, while the predictions on the y-axis.}
\label{fig:confMatsCross}
\end{figure*}

Furthermore, despite the high accuracy of the proposed AtteNet for traffic light recognition, we observe that utilizing only information from this module, as in ARCP(TLR), results in minor improvement in the accuracy over random guessing. We attribute this to the difficulty of accurately predicting the intersection safety for crossing in the absence of a traffic light or in cases where the classifier fails to detect the presence of one, which is further demonstrated in~\figref{fig:confMatsCross}(b), where the confusion matrix does not show strong distinction between the various classes. 
%employing only information from the traffic light recognition module as in ACP(TLR) results in an accuracy slightly better than random guessing. Despite the high accuracy of the proposed AtteNet for traffic light recognition, in the absence of a traffic light in the image or failure to detect it, predicting the safety of the intersection is a challenging task.

\begin{figure*}
\footnotesize 
\centering 
\setlength{\tabcolsep}{0.1cm} 
\begin{tabular}{p{0.1cm} p{4cm} p{4cm} p{4cm} p{4cm}}
\rotatebox[origin=c]{90}{Seq-1} & \raisebox{-0.5\height}{\includegraphics[width=1\linewidth]{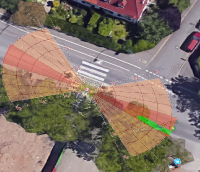}} &
\raisebox{-0.5\height}{\includegraphics[width=1\linewidth]{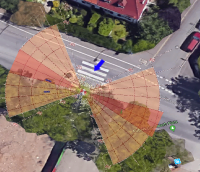}} &
\raisebox{-0.5\height}{\includegraphics[width=1\linewidth]{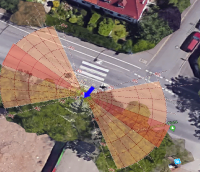}} &
\raisebox{-0.5\height}{\includegraphics[width=0.75\linewidth]{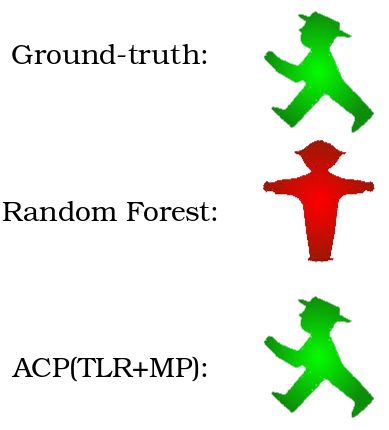}}
\\
& \multicolumn{1}{c}{(a) time=$0.0\second$} & \multicolumn{1}{c}{(b) time=$2.5\second$} & \multicolumn{1}{c}{(c) time=$5.0\second$} & \\
\\
\rotatebox[origin=c]{90}{Seq-2} & \raisebox{-0.5\height}{\includegraphics[width=1\linewidth]{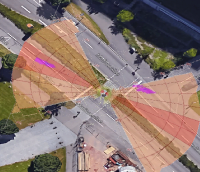}} &
\raisebox{-0.5\height}{\includegraphics[width=1\linewidth]{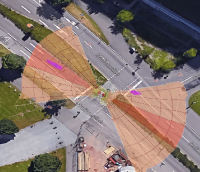}} &
\raisebox{-0.5\height}{\includegraphics[width=1\linewidth]{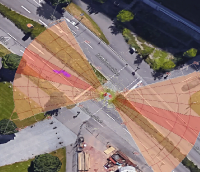}} &
\raisebox{-0.5\height}{\includegraphics[width=0.75\linewidth]{figures/crossingExamples/log_6-1_info_ampelmann.png}}
\\
& \multicolumn{1}{c}{(a) time=$0.0\second$} & \multicolumn{1}{c}{(b) time=$2.5\second$} & \multicolumn{1}{c}{(c) time=$5.0\second$} & \\
\\
\rotatebox[origin=c]{90}{Seq-3} & \raisebox{-0.5\height}{\includegraphics[width=1\linewidth]{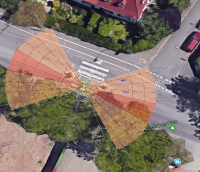}} &
\raisebox{-0.5\height}{\includegraphics[width=1\linewidth]{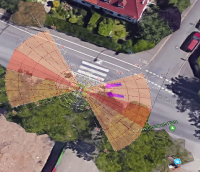}} &
\raisebox{-0.5\height}{\includegraphics[width=1\linewidth]{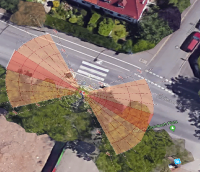}} &
\raisebox{-0.5\height}{\includegraphics[width=0.75\linewidth]{figures/crossingExamples/log_6-1_info_ampelmann.png}}
\\
& \multicolumn{1}{c}{(a) time=$0.0\second$} & \multicolumn{1}{c}{(b) time=$2.5\second$} & \multicolumn{1}{c}{(c) time=$5.0\second$} & \\
\\
\rotatebox[origin=c]{90}{Seq-4} & \raisebox{-0.5\height}{\includegraphics[width=1\linewidth]{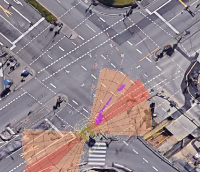}} &
\raisebox{-0.5\height}{\includegraphics[width=1\linewidth]{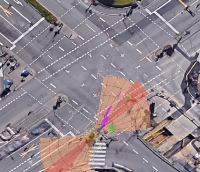}} &
\raisebox{-0.5\height}{\includegraphics[width=1\linewidth]{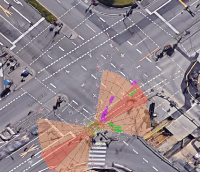}} &
\raisebox{-0.5\height}{\includegraphics[width=0.75\linewidth]{figures/crossingExamples/log_6-1_info_ampelmann.png}}
\\
& \multicolumn{1}{c}{(a) time=$0.0\second$} & \multicolumn{1}{c}{(b) time=$2.5\second$} & \multicolumn{1}{c}{(c) time=$5.0\second$} & \\
\\
\rotatebox[origin=c]{90}{Seq-5} & \raisebox{-0.5\height}{\includegraphics[width=1\linewidth]{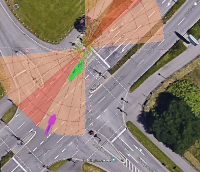}} &
\raisebox{-0.5\height}{\includegraphics[width=1\linewidth]{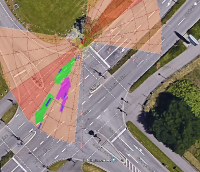}} &
\raisebox{-0.5\height}{\includegraphics[width=1\linewidth]{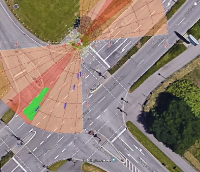}} &
\raisebox{-0.5\height}{\includegraphics[width=0.75\linewidth]{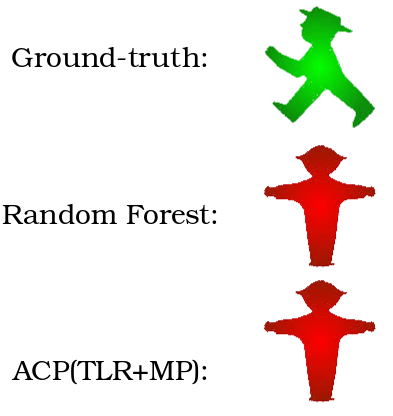}}
\\
& \multicolumn{1}{c}{(a) time=$0.0\second$} & \multicolumn{1}{c}{(b) time=$2.5\second$} & \multicolumn{1}{c}{(c) time=$5.0\second$} & \\
\\
\multicolumn{5}{c}{\includegraphics[width=0.35\linewidth]{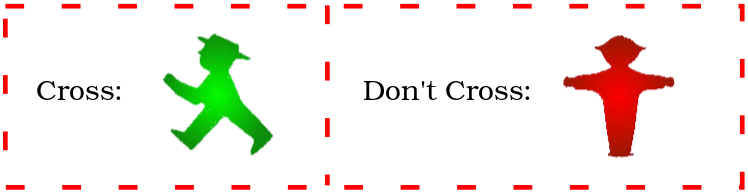}} \\
\end{tabular} 
\caption{Depiction of various crossing scenarios from the FSC dataset. For each sequence, we depict three timesteps corresponding to the beginning, middle and end of the interval. We overlay the sensor visualization on birds-eye-view images from the corresponding intersections. The robot location can be identified as the intersection of the two radar cones. Detected dynamic objects are represented by arrows, where magenta arrows signify objects moving towards the robot, and green arrows signify objects moving away from the robot. Furthermore, the size of the arrow grows proportionally with the detected velocity. With the exception of zebra crossings, traffic lights are located at both ends of the sidewalk, as well as at the middle island. For each sequence, on the right-most column, we depict the ground-truth label versus the predictions of the Random Forest classifier as a baseline and our proposed ARCP(TLR+MP) classifier. The safe to cross signal is depicted by a green walking symbol, while the unsafe is depicted by a red standing symbol. Please refer to the text for a more detailed description of each sequence.}
\label{fig:crossingExamples}
\end{figure*}

On the other hand, by employing information from only the motion prediction module, the overall accuracy of the crossing decision as well as the precision and recall are improved by $10.0\%, 9.0\%$ and $32.8\%$ respectively. Unlike ARCP(TLR), the confusion matrix shown in~\figref{fig:confMatsCross}(c) shows that the learned classifier is able to better differentiate between safe and unsafe crossing intervals. However, comparing the top row of the confusion matrix of ARCP(MP) with that of the Random Forest baseline shows that the ARCP(MP) classifier is more likely to label safe intervals as unsafe which could potentially lead to deadlock-type situations where the robot is stuck at the side of the road unable to cross. 

Inspecting the results of employing a constant velocity motion model coupled with the traffic light recognition module in~\figref{fig:confMatsCross}(d) shows that similar to the ARCP(MP) model, this model provides an improvement of $11.0\%, 12.4\%$ and $15.0\%$ to the precision, recall and accuracy, respectively,  over solely relying on the traffic light classifier. However, unlike the ARCP(MP) classifier, we see more confusion between the safe and unsafe predictions. We hypothesize that this is due to the constant velocity assumption which can result in suboptimal performance in cases where a car speeds up or slows down at the end of an interval. The performance of this model is, nonetheless, superior to the NCP(TLR+MP) as shown in~\figref{fig:confMatsCross}(e). Simple fusion of the predictions of both modules into a binary classifier proves the least optimal strategy, albeit performing slightly better than relying solely on the traffic light information. We believe this to be a direct consequence of discarding the confidence information from both modules and the manner with which the features were fused resulting in the TLR features overpowering the MP features. 

Utilizing the ARCP(TLR+MP) model shown in~\figref{fig:confMatsCross}(f) reduces both the numbers of false positives and false negatives with respect to the intersection safety. By utilizing a structured approach for combining information from both the traffic light recognition and the motion prediction modules, the learned classifier is able to make accurate crossing predictions while being robust to the type of intersection encountered. Furthermore, by incorporating feature maps from the last downsampling stage in AtteNet and the Gaussian distribution parameters of IA-TCNN, the learned classifier can better generalize to unseen environments as shown by the improvement in the accuracy, precision and recall rates over the Random Forest baseline classifier in~\tabref{tab:prCross}.

In~\figref{fig:crossingExamples}, we perform qualitative analysis of the learned decision of our proposed ARCP(TLR+MP) classifier in comparison to the Random Forest classifier as a baseline on the FSC dataset. Each sequence is represented by three images corresponding to the beginning, middle and end of the interval. Furthermore, to provide a complete image of the scene, we overlay the sensor detections on birds-eye-view images of the intersections. Seq-1 depicts a situation where the robot is located at the side of a zebra crossing that is clear, with the exception of a cyclist (represented by the blue arrow) that is moving towards the robot. Despite the intersection being safe for crossing, the Random Forest classifier incorrectly labels the interval as unsafe for crossing. On the other hand, the ARCP(TLR+MP) classifier correctly identifies the intersection state by utilizing the information from the motion prediction module to infer the driving direction of the cyclist.

In Seq-2, the robot is located in the middle island at a signalized intersection, with traffic coming from the left-hand side. As the pedestrian traffic light is green, the approaching vehicle slows down throughout the observed interval rendering the intersection safe for crossing. By utilizing information from the traffic light recognition module to detect the state of the traffic light, in combination with the motion prediction module to identify that the approaching vehicle is reducing its velocity, our ARCP(TLR+MP) classifier is able to correctly label the interval as safe for crossing. %On the other hand, the Random Forest classifier mispredicts the interval as unsafe. 

Similar to Seq-1, in Seq-3 the robot is located at the side of a zebra crossing. This sequence demonstrates a situation where a false positive detection by the tracker causes an incorrect classification of the intersection as unsafe by the Random Forest classifier. Our proposed classifier is however able to correctly predict the safety of the intersection for crossing as it is able to identify the spurious detection as a false positive or a ghost detection by the tracker. Another scenario is depicted in Seq-4, where the robot is located at a grid-type signalized intersection, with vehicles approaching from the upper right corner and heading towards the street parallel to the robot. By utilizing only information from the tracker, the Random Forest classifier labels the crossing unsafe as it appears that the cars are approaching perpendicularly to the robot. However, by predicting the behavior of the vehicles for the remainder of the interval, our proposed ARCP(TLR+MP) classifier is able to correctly classify the safety of the interval for crossing. 

Finally, Seq-5 depicts an interval for which both classifiers incorrectly label the intersection as unsafe. The robot is placed at a signalized intersection with a green pedestrian light and a vehicle approaching from the lower left corner of the image. As the intersection contains a middle island, and since there is no traffic approaching from the significant direction (top left corner of the image), the crossing is labeled as safe. However, as neither classifier has a representation of the structure of the intersection showing the presence of the middle island, the interval is in turn misclassified as unsafe for crossing. By incorporating semantic knowledge of the scene or learning an obstacle map of the environment, the aforementioned problem can be rectified as the classifier can learn about the various road topologies and their effect on the crossing decision. 

\subsection{Generalization across Different Intersections}
\label{sec:genExps}
In the following, we perform an extended evaluation of our proposed pipeline for predicting the safety of the intersection for crossing by analyzing the performance of each module as well as the entire system in a real-world scenario. We place our robotic platform shown in~\figref{fig:obelix}, with the proposed framework, at a busy street intersection which contains high speed traffic, a pedestrian crossing, and a tram line. Note that this intersection is not included in either the training or test sequences of the FSC dataset, rather the goal of this experiment is to evaluate the generalization capabilities of our framework to new environments.

\begin{figure*}
\footnotesize 
\centering 
\setlength{\tabcolsep}{0.1cm} 
\begin{tabular}{p{0.1cm} p{3.5cm} p{3.5cm} p{3.5cm} p{3.5cm}}
\rotatebox[origin=c]{90}{Seq-1} & \raisebox{-0.5\height}{\includegraphics[width=1\linewidth]{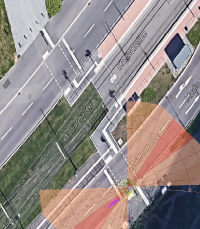}} &
\raisebox{-0.5\height}{\includegraphics[width=1\linewidth]{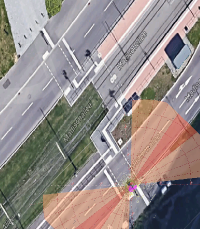}} &
\raisebox{-0.5\height}{\includegraphics[width=1\linewidth]{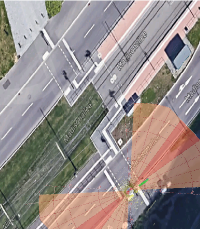}} &
\raisebox{-0.5\height}{\includegraphics[width=1\linewidth]{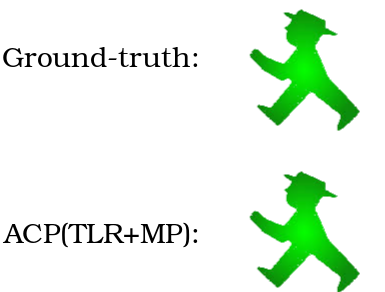}}
\\
& \multicolumn{1}{c}{(a) time=$0.0\second$} & \multicolumn{1}{c}{(b) time=$2.5\second$} & \multicolumn{1}{c}{(c) time=$5.0\second$} & \\
\rotatebox[origin=c]{90}{Seq-2} & \raisebox{-0.5\height}{\includegraphics[width=1\linewidth]{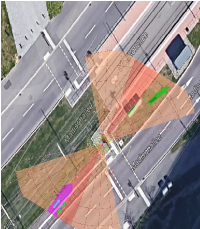}} &
\raisebox{-0.5\height}{\includegraphics[width=1\linewidth]{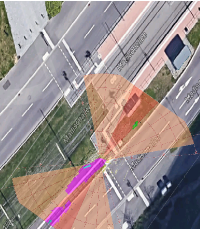}} &
\raisebox{-0.5\height}{\includegraphics[width=1\linewidth]{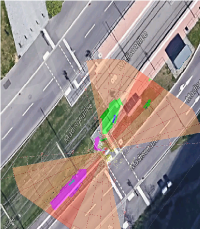}} &
\raisebox{-0.5\height}{\includegraphics[width=1\linewidth]{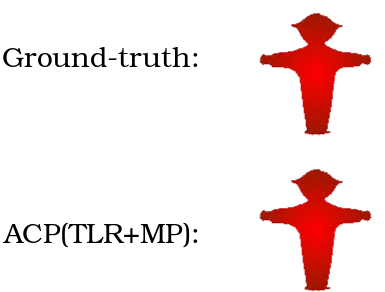}}
\\
& \multicolumn{1}{c}{(a) time=$0.0\second$} & \multicolumn{1}{c}{(b) time=$2.5\second$} & \multicolumn{1}{c}{(c) time=$5.0\second$} & \\
\rotatebox[origin=c]{90}{Seq-3} & \raisebox{-0.5\height}{\includegraphics[width=1\linewidth]{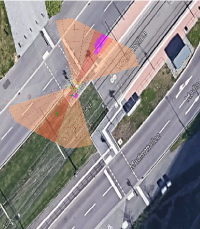}} &
\raisebox{-0.5\height}{\includegraphics[width=1\linewidth]{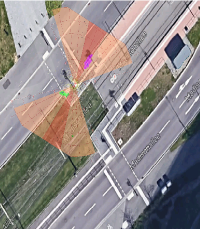}} &
\raisebox{-0.5\height}{\includegraphics[width=1\linewidth]{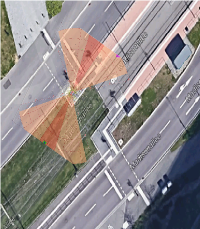}} &
\raisebox{-0.5\height}{\includegraphics[width=1\linewidth]{figures/crossingExamples/deeploccross_ex-1_info.png}}
\\
& \multicolumn{1}{c}{(a) time=$0.0\second$} & \multicolumn{1}{c}{(b) time=$2.5\second$} & \multicolumn{1}{c}{(c) time=$5.0\second$} & \\
\\
\multicolumn{5}{c}{\includegraphics[width=0.45\linewidth]{figures/crossingExamples/label_legend_ampelmann.png}}
\end{tabular} 
\caption{Depiction of the crossing scenarios from our generalization experiment. For each sequence, we depict three timesteps corresponding to the beginning, middle and end of the interval. We overlay the sensor visualization on birds-eye-view images from the corresponding intersections. The robot's location can be identified as the intersection of the two radar cones. All intersections visualized here are signal regulated with traffic lights at both sides of the road as well as the middle island. For each sequence, on the right-most column, we depict the ground-truth label versus the predictions of our proposed ARCP(TLR+MP) classifier. The legend is shown enclosed in a red rectangle. Please refer to the text for a detailed description of each sequence.}
\label{fig:deeploccrossExamples}
\end{figure*}

Employing our IA-TCNN  on the dataset to predict the future trajectories of all observable traffic participants, we achieve an average displacement error of $0.14\meter$, $11.41\degree$, $0.13\si{\meter\per\second}$ in terms of translation, rotation and velocity respectively. Furthermore, using AtteNet we achieve an accuracy of $78.5\%$ for predicting the state of the traffic light. Overall, predicting the safety of the intersection for crossing, our ACP classifier achieves a precision of $85.7\%$ and a recall of $78.2\%$. The low prediction errors achieved by the overall network as well as the individual modules demonstrate the generalization capabilities and efficacy of our proposed framework.

Additionally, we perform qualitative analysis of the crossing decision predicted by our proposed ACP classifier in~\figref{fig:deeploccrossExamples}. We depict three sequences from the intersection, where each sequence is represented by three images from the beginning, middle and end of the prediction interval. As in~\figref{fig:crossingExamples}, we overlay the sensor detections on birds-eye-view images of the intersection to provide a more comprehensive image of the scene. In each of these sequences, the robot's location can be identified as the intersection center of the sensor cones. Each of the the three sequences is from a signal regulated intersection with traffic lights at either end as well as the middle island. Seq-1 depicts a scenario wherein a cyclist is approaching the robot on the sidewalk from the direction of oncoming traffic and continues to cycle past the robot. Utilizing and predicting the orientation information for the observed traffic participants, our IA-TCNN network is able to predict an accurate trajectory for the cyclist continuing on the sidewalk and hence passing behind the robot as opposed to in front of it. This information is in turn used by our ACP classifier to correctly predict the safety of the intersection for crossing.

Seq-2 depicts a situation with heavy oncoming traffic, in which our ACP classifier accurately predicts the intersection at the observed interval to be not safe for crossing due to the heavy oncoming traffic. In Seq-3, in the first half of the interval a car is approaching the intersection, however, at the remaining half it slows down as the traffic light signal changes. By utilizing both the traffic light predictions from AtteNet showing the pedestrian traffic light to be green, and the trajectory information from IA-TCNN predicting the continued deceleration of the car until it comes to a halt at the end of the interval, our ACP classifier is able to accurately predict the safety of the intersection at the given interval for crossing.

\section{Conclusion}
\label{sec:conclusion}

In this paper, we proposed a system for autonomous street crossing using multimodal data. Our system consists of two main network streams; a traffic light recognition stream and an interaction-aware motion prediction stream. Information from both streams is fused as input to a convolutional neural network to predict the safety of the intersection for crossing. We proposed AtteNet, a convolutional neural network architecture for traffic light recognition that utilizes the global information in the images to selectively emphasize informative features suppressing irrelevant features, while being robust to noisy data. We performed extensive experimental evaluations on various traffic light recognition benchmarks and show that the proposed architecture outperforms the compared methods. Furthermore, we proposed an interaction-aware temporal convolutional neural network architecture that utilizes causal convolutions to accurately predict the trajectories of dynamic objects. We demonstrated that our approach is scalable to complex urban environments while simultaneously being able to predict accurate trajectories of all the observable traffic participants in the scene. Experimental evaluations on several benchmark datasets demonstrate that our architecture achieves state-of-the-art performance on both indoor and outdoor datasets, while achieving faster inference times and requiring less storage space in comparison to recurrent approaches. 

In order to learn a classifier that is robust to the type of intersection, feature maps from the traffic light recognition network and the interaction-aware motion prediction network are fused to learn the final crossing decision. By incorporating the uncertainty information from the motion prediction stream and the learned representations from the traffic light recognition stream, the classifier is robust to incorrect predictions by either task-specific subnetwork. Moreover, we extended the previously introduced Freiburg Street Crossing dataset by additional sequences captured at various intersections including signalized and zebra intersections, as well as various road curvatures and topologies which affect the crossing procedure. We deployed our proposed framework on a robotic platform and conducted real-world experiments that demonstrate the accuracy, robustness and generalization capabilities of the proposed system to new environments. We conducted comprehensive experimental evaluations that demonstrate the efficacy of the proposed system for determining the safety of the intersection for crossing. Furthermore, the results demonstrate the tolerance of the system to noise and inaccuracies in the data, while accurately generalizing to new unseen scenarios. 

For future work, we aim to additionally predict the obstacle map of the environment, as we believe that knowledge about the vicinity can improve the motion prediction accuracy by avoiding trajectories that may intersect with obstacles. Similarly, the crossing prediction accuracy would also benefit as it eliminates false negatives by leveraging the road structure. Moreover, learning to predict the traffic flow direction can aid in eliminating further sources of confusion for the crossing decision.

\bibliographystyle{SageH}
\bibliography{sections/references}

\appendix
\section{Appendix}

\subsection{Residual Units}
\label{app:resunits}
\begin{figure}
\footnotesize 
\centering 
\setlength{\tabcolsep}{0.2cm} 
\begin{tabular}{cc}
{\includegraphics[width=0.45\linewidth]{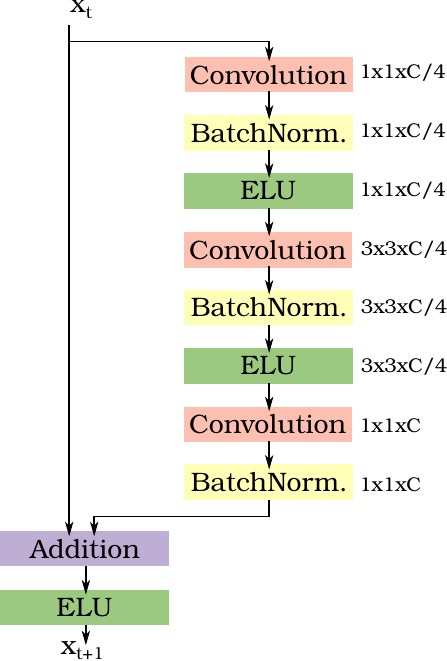}} &
{\includegraphics[width=0.45\linewidth]{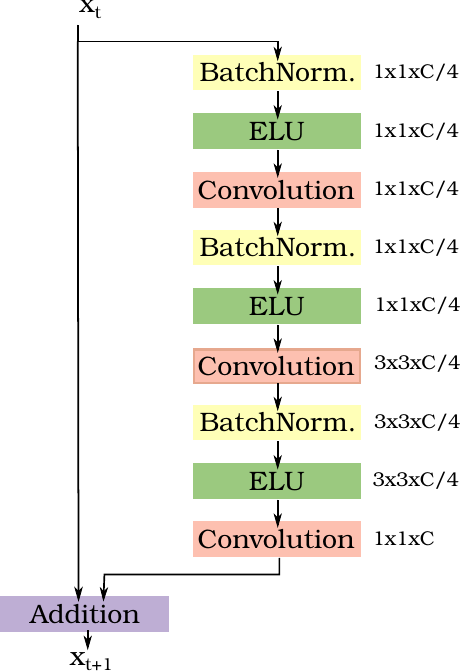}} \\ 
\\
\multicolumn{1}{c}{(a) Original Residual Unit} & \multicolumn{1}{c}{(b) Pre-activation Residual Unit}
\end{tabular} 
\caption{Schematic illustration of the original residual unit~\citep{he2016deep} and the pre-activation residual unit~\citep{he2016identity}.} 
\label{fig:resunit}
\end{figure}

We show the standard residual unit in~\figref{fig:resunit}(a) and the pre-activation residual unit in~\figref{fig:resunit}(b). Each residual unit, depicted in~\figref{fig:resunit}(b), consists of batch normalization and ELU preceding the convolutional layers. By moving the batch normalization layer to the beginning of the residual unit, the input of the layer is ensured to be renormalized after the addition operation from the previous layer, thereby improving the regularization of the network. Similarly, moving the ELU activations to the beginning of the unit as opposed to after the addition operation ensures that the original information is preserved throughout the entire network. 

\subsection{Squeeze-Excitation Blocks}
\label{app:seblock}

\begin{figure}
\footnotesize
\centering
\includegraphics[width=0.5\linewidth]{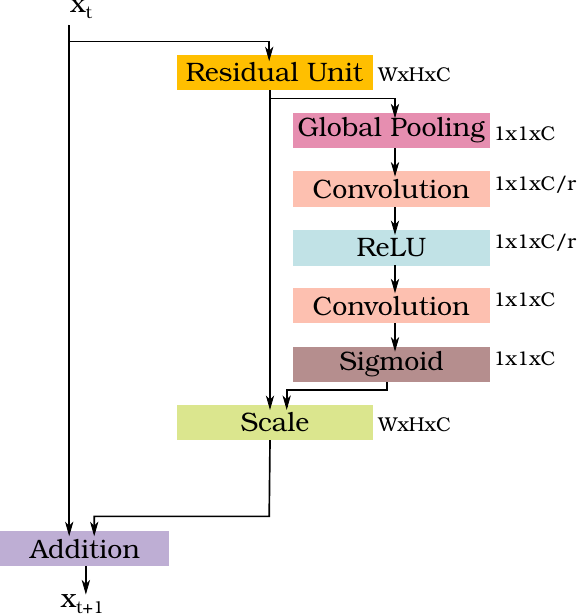}
\caption{Illustration of a Squeeze-Excitation (SE) residual block. The output of the residual block is first passed through a squeeze operation to aggregate the feature maps across the spatial dimensions, then through an excitation operation to emphasize the informative features and suppress the irrelevant features.}
\label{fig:se}
\end{figure}

We depict a Squeeze-Excitation (SE) block in~\figref{fig:se}. Each block is comprised of two operations; squeeze and excite. During the squeeze operation feature maps from the previous layer are aggregated across the spatial dimension. Thus embedding the global distribution of the features to be leveraged by upcoming layers in the network. The excitation operation which follows emphasizes the informative features and suppresses the irrelevant ones thus aiding in learning sample-specific activations for each channel. Thus instead of equally weighting all channels while creating the output feature maps, each Squeeze-Excitation (SE) block employs a content aware mechanism which learns an adaptive weighting for each channel with a minimum computational cost.

\subsection{Datasets}
\label{app:datasets}

In this section, we expand on the details of the public datasets that we have used for evaluating our method.

\subsubsection{Motion Prediction Datasets}
\label{app:mopred_data}\hfill\\

\textbf{L-CAS} dataset is a recently proposed benchmark for pedestrian motion prediction~\citep{yan2017online}. The data is captured using a 3D LiDAR scanner mounted on top of a Pioneer robot placed inside a university building. It consists of over 900 pedestrian tracks, each with an average length of $13.5$ seconds and is divided into a training and testing split. Each pedestrian is identified by a unique ID, a time frame at which they are detected, the spatial coordinates, and their orientation angle. Some of the challenges faced while benchmarking on this dataset are people pushing trolleys, children running and groups dispersing. \figref{fig:mpDatasets}(c) shows an example scan from the dataset, where pedestrians are marked by bounding boxes with arrows showing their trajectories for a sample interval. We use the same training and test splits provided by the authors for this dataset to facilitate comparison with other approaches. %On this dataset, we predict both the spatial coordinates and orientation for each pedestrian.

\textbf{ETH} crowd set dataset consists of two scenes: Univ and Hotel, containing a total of approximately 750 pedestrians exhibiting complex interactions~\citep{pellegrini2009you}. For each scene, the dataset contains an obstacle map file providing the static map information of the surroundings, and an annotations file which provides the trajectory information for each pedestrian. Each tracked pedestrian is identified by a pedestrian ID, the frame number at which they were observed, the spatial coordinates and velocity with which they were traveling. The dataset additionally provides a groups file that provides information on pedestrians forming a group and a destinations file reporting the assumed destinations of all subjects in the scene. The dataset is one of the widely used benchmarks for pedestrian tracking and motion prediction as it represents real world crowded scenarios with multiple non linear trajectories, covering a wide range of group behavior such as crossings, dispersing and forming of groups. We show an example image from the Hotel sequence in~\figref{fig:mpDatasets}(a), where arrows represent the trajectories of the pedestrians for a sample interval. The sequence is recorded near a public transport stop. It captures the complex behavior of pedestrians as they enter/exit the vehicle as well as surrounding pedestrians navigating the scene. For this dataset, we utilize only the information from the annotations file, keeping track of the spatial coordinates of each pedestrian at each time frame. Furthermore, we assume no knowledge of the destination of each pedestrian, nor do we utilize any information regarding group behavior or the structure of the environment.

\textbf{UCY} dataset consists of three scenes: Zara01, Zara02 and Uni, with a total of approximately 780 pedestrians~\citep{lerner2007crowds}. For each scene, the dataset provides an annotations file comprised of a series of splines each describing the trajectory of a pedestrian using the spatial coordinates, frame number and the viewing direction of the pedestrian. This dataset in addition to the ETH dataset are widely used in conjunction as benchmarks for motion prediction and pedestrian tracking due to the wide range of non linear trajectories and pedestrian interactions exhibited including group behavior and pedestrians idling nearby shop fronts. \figref{fig:mpDatasets}(b) shows a sample image from the Uni sequence, where pedestrian trajectories are represented by arrows. This particular sequence is the most challenging among the three sequences forming this dataset due to the large number of pedestrians observed concurrently, in addition to the complex crowd behavior demonstrated. We combine both this dataset with the ETH dataset similar to previous works~\citep{alahi2016social, vemula2017social} and apply a leave-one-out procedure during training by randomly selecting trajectories from all scenes except the scenes used for testing. Furthermore, in order to facilitate the combination of the datasets, we predict only the 2D spatial coordinates for each pedestrian.

\subsubsection{Traffic Light Datasets}
\label{app:tlr_data}\hfill\\

\textbf{Nexar Traffic Lights} dataset consists of over 18000 RGB images captured in varying weather and lighting conditions. The dataset was released as part of a challenge to recognize the traffic light state in images taken by drivers using the Nexar app~\citep{nexar}. Each image is labeled with the state of the traffic light in the driving direction, where $S = \left\lbrace \textit{Red, Green, Off} \right\rbrace$. Several factors make benchmarking on this dataset extremely challenging such as the varying light conditions, the presence of substantial motion blur and the presence of multiple traffic lights in the image. The top row in~\figref{fig:tlrDatasets} shows sample images from the dataset. In addition to the aforementioned challenges, the evaluation criteria for this dataset was selected to be the classification accuracy and model size, with a minimum success criteria of $95.0\%$ in terms of accuracy for submission acceptance. In order to train our method, we split the data into a training and a validation set using a split ratio of $4:1$ and perform augmentations on the training split in the form of random applications of brightness and contrast modulations.

\textbf{Bosch Small Traffic Lights} dataset contains RGB images at a resolution of $1280\times720$ pixels captured in the San Fransisco Bay Area and Palo Alto, California~\citep{BehrendtNovak2017ICRA}. The training set consists of over 5000 images which are annotated at a 2 second interval, while the test set consists of over 8000 images annotated at a frame rate of 15 fps. Each image contains multiple labeled traffic lights amounting to a total of over 10000 annotated traffic lights in the training set and 13000 in the test set, with a median traffic light width of 8.6 pixels. For each image, the label file includes the bounding box coordinates of the traffic light, the status of the light $S = \left\lbrace \textit{Red, Green, Yellow, Off} \right\rbrace$, and whether the light is occluded by some object. This dataset is among the challenging benchmarks for detecting and recognizing traffic lights due to the small size of the lights in the image as well as the varying lighting conditions, presence of shadows and occlusions. We show example images from this dataset in ~\figref{fig:tlrDatasets}(c, d). We use the same training and test split provided by the authors and apply augmentations on the training set in the form of random brightness and contrast modulations. As our approach only predicts the status of the traffic light and not its position, we preprocess each image by masking out all but one traffic light using the bounding box information from the label file. To learn identifying when no traffic light is present in the image, we additionally mask out all the traffic lights, thereby creating from each image $N+1$ images where $N$ is the number of non-occluded traffic lights.

\end{document}